\documentclass{bmvc2k}
\usepackage{float}

\title{Beyond False Stability: High-Noise Drift Gating for Test-Time Adversarial Defenses in Vision-Language Models
} 

\addauthor{Hashmat Shadab Malik}{hashmat.malik@mbzuai.ac.ae}{1}
\addauthor{Muzammal Naseer}{muhammadmuzammal.naseer@ku.ac.ae}{2,3}
\addauthor{Salman Khan}{salman.khan@mbzuai.ac.ae}{1,4}

\addinstitution{
 Mohamed Bin Zayed University of AI,\\
 UAE
}
\addinstitution{
 Khalifa University, UAE
}
\addinstitution{
 University of Western Australia, \\
 Australia
}
\addinstitution{
 Australian National University, \\
 Australia
}
\runninghead{Malik et.al}{Beyond False Stability}

\usepackage{graphicx}
\usepackage{amssymb}
\usepackage{amssymb}
\usepackage{wrapfig}
\usepackage{booktabs}
\usepackage[table]{xcolor}
\usepackage{multirow}
\usepackage{booktabs}
\usepackage{graphicx}
\usepackage[table]{xcolor}
\usepackage{algorithm}
\usepackage{algpseudocode}
\usepackage[rightcaption]{sidecap} 
\usepackage{booktabs}
\usepackage[table]{xcolor}
\usepackage{adjustbox} 

\begin{document}

\maketitle

\begin{abstract}
Vision--language models (VLMs) such as CLIP demonstrate strong zero-shot generalization but remain highly vulnerable to adversarial attacks. While adversarial training can improve robustness, it is often computationally expensive, motivating defenses that operate at test time. Recent test-time approaches attempt to improve robustness by exploiting how CLIP's visual representations respond to stochastic perturbations. Some aggregate predictions across multiple noisy views, while others construct Gaussian noise-averaged anchors and interpolate vision features toward these anchors, or apply counter-perturbations to move adversarial inputs away from their visual feature representations. Although these strategies can improve robustness, they often degrade clean accuracy, leading to an unfavorable clean--robust trade-off.
In this work, we revisit stochastic test-time defenses and identify an underexplored \emph{noise-regime transition} in CLIP's visual representation space. While recent work has explored stochastic 
perturbations primarily in the weak-noise 
regime, where adversarial examples can 
appear unusually stable (\emph{false 
stability}), our analysis shows that this 
characterization reverses as perturbation 
strength increases: beyond the weak-noise 
regime, adversarial representations become 
\emph{markedly more unstable} than clean 
ones, yielding a substantially clearer 
separation signal. This transition is consistent across uniform and Gaussian noise, photometric/geometric transformations, datasets, and 
diverse attack objectives and perturbation 
budgets. Moreover, the separation largely disappears in adversarially trained models, supporting the view that the phenomenon is tied to the fragile local-basin geometry of adversarial representations in non-robust CLIP models.
Motivated by this finding, we propose a training-free, plug-in \emph{drift-gated} mechanism that uses high-noise feature drift as a lightweight gating signal to selectively trigger existing test-time defenses only when adversarial-like instability is detected. Evaluated across 13 downstream datasets, 
our approach consistently improves the 
clean--robust trade-off of existing 
test-time defenses. Across eight fine-grained datasets, the 
average of clean and adversarial accuracy 
improves from 65.7\% to 71.4\% for 
counterattack-style defenses and from 68.4\% 
to 73.2\% for noise-anchoring approaches; 
on ImageNet and four distribution-shifted 
variants, the corresponding gains are from 
56.1\% to 66.2\% and from 62.1\% to 67.6\%.
Our code will be made publicly available.

\end{abstract}

\section{Introduction}
\label{sec:introduction}

Large-scale vision--language models (VLMs)~\cite{yu2022coca,saharia2022photorealistic,ramesh2021zero,radford2021learning,jia2021scaling,zhang2024vision} have emerged as core multimodal foundation models by learning aligned image--text representations from web-scale corpora. A prominent example is CLIP~\cite{radford2021learning}, which enables zero-shot classification by scoring the similarity between an input image and a set of candidate text prompts and selecting the most compatible label in a shared embedding space. This simple interface, together with strong transferability, has made CLIP a common backbone for tasks beyond classification~\cite{wang2023improving,wang2024hard,zhou2022conditional,zhou2022learning}, including retrieval, segmentation, detection, and clustering~\cite{shin2022reco,zhou2022extract,zhang2023simple,zhao2022exploiting,cai2023semantic,li2023image}. Although CLIP demonstrates strong zero-shot generalization, it remains vulnerable to adversarial perturbations---a weakness widely observed across deep neural networks~\cite{szegedy2013intriguing,papernot2016limitations,moosavi2016deepfool,carlini2017towards,athalye2018obfuscated,croce2020reliable}. Even small, visually imperceptible perturbations can cause substantial shifts in predictions. This vulnerability is particularly concerning given the broad deployment of CLIP in real-world applications.

To mitigate CLIP's adversarial vulnerability, prior work has predominantly relied on adversarial training to enhance robustness~\cite{madry2017towards,li2024language,mao2022understanding,schlarmann2024robust,wang2024pre,zhang2024adversarial,zhou2024few}. Within this paradigm, methods typically either (i) perform adversarial finetuning, where adversarial examples are generated during training and the visual encoder is finetuned via a min--max objective---often using a limited auxiliary dataset---to obtain robustness that transfers in a zero-shot setting~\cite{mao2022understanding,schlarmann2024robust,wang2024pre,li2024language}, or (ii) adopt adversarial prompt tuning, which keeps the visual encoder frozen while optimizing textual prompts under an adversarial objective to improve alignment on perturbed inputs~\cite{zhang2024adversarial,zhou2024few}. Despite improved adversarial performance, these approaches are often time-consuming---adversarial finetuning in particular requires repeatedly generating adversarial examples during training---and are prone to overfitting to the finetuning distribution, which can hurt generalization to clean and adversarial data from other domains. Moreover, the adversarial robustness gain comes with a substantial drop in clean accuracy, underscoring a challenging clean--robustness trade-off~\cite{zhang2019theoretically}.

These limitations motivate test-time defenses that preserve pretrained weights and improve robustness without additional adversarial training or its associated computational cost. Recent works~\cite{xing2025clip,tong2025zero,sheng2025r} treat adversarial vulnerability as a test-time problem and leverage a shared behavioral insight: non-robust CLIP models exhibit systematic differences in how clean and adversarial inputs respond to stochastic perturbations. Specifically, \cite{xing2025clip} injects 
small uniform noise and measures 
latent-space drift, using unusually low 
drift (\emph{false stability}) as a gating 
signal to identify adversarial inputs and 
selectively apply a PGD-based~\cite{madry2017towards} 
counterattack that pushes representations 
away from their adversarial embedding 
region. In contrast, \cite{tong2025zero} is motivated by the observation that Gaussian noise injection can suppress adversarial perturbations: it constructs a Gaussian noise-averaged feature anchor from multiple noisy views and interpolates the original representation toward this anchor. Despite their promise, these approaches expose a recurring trade-off between clean and adversarial performance. In \cite{xing2025clip}, the drift-based gating signal is not perfectly reliable; consequently, some clean inputs are also flagged as adversarial and counterattacked, which leads to reduced clean accuracy. Likewise, the Gaussian noise injection and feature interpolation in \cite{tong2025zero}, while beneficial for adversarial samples, can distort clean representations---especially as noise and interpolation strength increase---leading to degraded clean performance.

In this work, we argue that this trade-off is partly caused by analyzing stochastic perturbations in a restricted weak-noise regime. Prior work often evaluates random perturbations at small strengths and concludes that they provide limited robustness, or uses weak-noise drift as a trigger for adversarial detection. However, weak stochastic perturbations only reveal one side of CLIP’s feature-space behavior. Under weak noise, adversarial examples can appear locally stable and exhibit smaller feature drift than clean inputs. As perturbation strength increases, this behavior reverses: adversarial representations become substantially more unstable than clean ones, producing a clearer separation signal in the high-noise regime. This suggests that conclusions drawn from weak-noise evaluations do not fully characterize the behavior of stochastic test-time defenses.

Motivated by this observation, we revisit stochastic test-time 
robustness from a \emph{noise-regime} perspective. We systematically 
analyze CLIP visual features across increasing perturbation strengths 
and show that the transition from weak-noise \emph{false stability} 
to high-noise instability appears across uniform noise, Gaussian 
noise, and broader photometric and geometric transformations. This 
finding also clarifies that the robustness gains of noise-based 
defenses are not tied to a specific corruption distribution: 
sufficiently strong stochastic perturbations can destabilize 
adversarial representations across multiple transformation families. 
Empirically, we observe that this separation is prominent for 
non-robust CLIP models but largely disappears in adversarially 
trained CLIP variants~\cite{schlarmann2024robust,wang2025double}, 
supporting the interpretation that high-noise drift exposes fragile adversarial feature regions that are reduced by adversarial training but are more prominent in non-robust CLIP models.

We use this high-noise drift signal to build a simple training-free 
\emph{drift-gated} test-time defense. Instead of applying test-time defensive 
interventions to every input, we first probe the input with a strong 
stochastic perturbation and measure feature drift in CLIP's visual 
embedding space. If the input exhibits high drift, we treat it as 
adversarial-like and activate a test-time intervention; otherwise, 
we preserve standard CLIP inference. This selective mechanism 
reduces unnecessary corruption of clean representations while 
retaining robustness gains on adversarial examples. Our approach combines with existing stochastic defenses without any
additional training. We summarize our main contributions as:

\begin{figure}[t]
    \centering
    \includegraphics[width=0.98\linewidth]{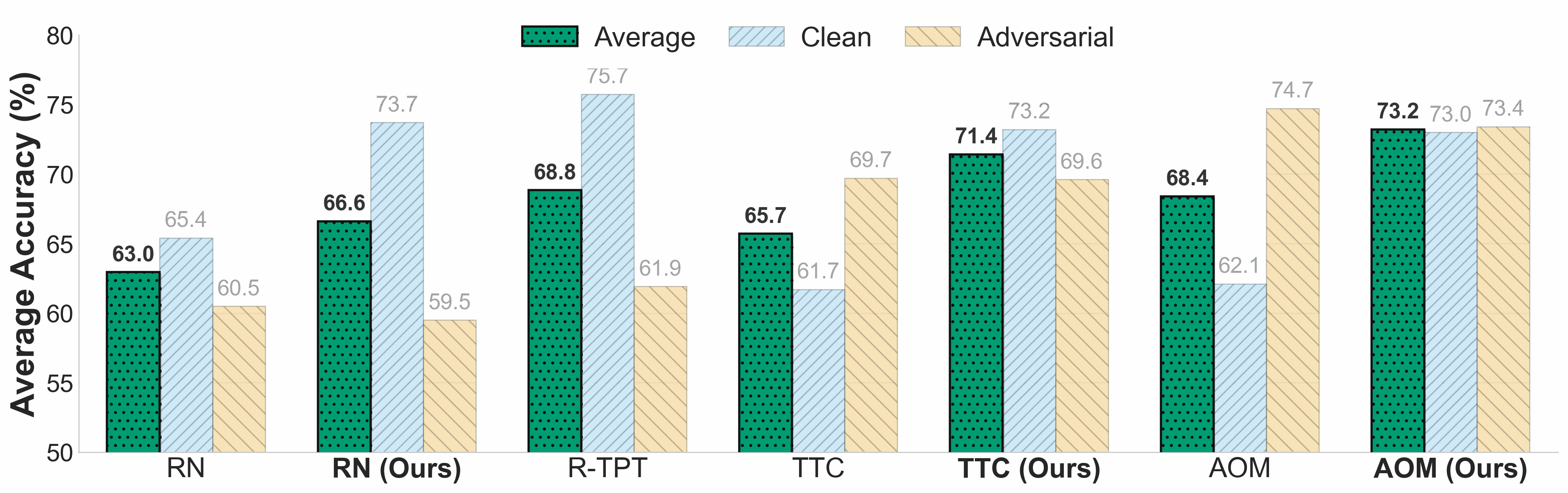}
    \vspace{-1em}
\caption{Clean \emph{(blue)}, Adversarial 
\emph{(yellow; PGD-100, 
$\epsilon{=}4/255$)}, and Average accuracy 
\emph{(green; 
$(\text{clean}+\text{adversarial})/2$)} for 
test-time defense methods, averaged across 
eight fine-grained datasets. Current methods~\cite{xing2025clip,
tong2025zero} employ a 
low-strength random-noise (RN) baseline 
at noise strength similar to the attack 
\emph{($\epsilon{=}4/255$)}, which 
provides no effective robustness.
However, applying high-strength random noise 
\emph{(RN, $\epsilon{=}48/255$)} to all 
inputs already recovers substantial 
adversarial accuracy. Using our high-noise drift as a gating 
signal to selectively apply test-time 
interventions only when adversarial-like 
instability is detected improves the 
clean--robust trade-off, increasing average 
performance from 
$63.0\%{\rightarrow}66.6\%$ for RN,
$65.7\%{\rightarrow}71.4\%$ for TTC~\cite{xing2025clip} 
and $68.4\%{\rightarrow}73.2\%$ for AOM~\cite{tong2025zero}.}
    \label{fig:main_intro_figure}
    \vspace{-1.5em}
\end{figure}

\begin{itemize}

    \item \textbf{Noise-regime transition in CLIP representations.}
    We identify and characterize a consistent \emph{transition} in 
    CLIP's visual feature-space response to stochastic perturbations 
    that has been overlooked by prior work. Under weak perturbations, 
    adversarial examples exhibit \emph{false stability}---smaller 
    latent drift than clean inputs, as noted by~\cite{xing2025clip}. 
    As perturbation strength increases, this ordering \emph{reverses}: 
    adversarial inputs become markedly more unstable than clean ones, 
    yielding a substantially clearer separation signal. This transition 
    is consistent across noise types (uniform, Gaussian, photometric, 
    geometric), multiple attack budgets 
    ($\epsilon \in \{1/255, 4/255, 8/255\}$), and diverse datasets. 

    \item \textbf{Beyond Gaussian-specific suppression.}
    While \cite{tong2025zero} constructs a Gaussian noise-averaged anchor and interpolates features toward it to suppress adversarial perturbations, we show that similar gains can be achieved using sufficiently strong \emph{uniform} noise. Analogous trends also hold for photometric and geometric transformations, indicating that perturbation \emph{strength}, rather than the specific corruption distribution, is the key factor.

    \item \textbf{Drift-gated selective defense.}
    As a direct application of the above diagnosis, we propose a 
    training-free, plug-in mechanism that uses high-noise latent drift 
    as a lightweight \emph{adversarial detector} to selectively trigger 
    existing test-time defenses (TTC~\cite{xing2025clip}, 
    AOM~\cite{tong2025zero}, R-TPT~\cite{sheng2025r}) only when 
    adversarial-like instability is detected. This design avoids 
    unnecessary corruption of clean representations and consistently 
    improves the clean--robust trade-off with no additional training 
    cost. Across eight fine-grained datasets (see Fig.~\ref{fig:main_intro_figure}), our approach improves 
    the clean/adversarial average from 65.7\% to 71.4\% for 
    TTC~\cite{xing2025clip}, from 68.4\% to 73.2\% for 
    AOM~\cite{tong2025zero}, and from 68.8\% to 73.2\% when combined 
    with R-TPT~\cite{sheng2025r}. On ImageNet and four 
    distribution-shifted variants (see Fig.~\ref{fig:main_imagenet_results}), the corresponding gains are from 
    56.1\% to 66.2\% for TTC and from 62.1\% to 67.6\% for 
    AOM.

\end{itemize}

\section{Related Work}

\textbf{Zero-shot Adversarial Robustness.}
Zero-shot adversarial robustness in CLIP remains challenging. Small, imperceptible perturbations can induce large changes in predicted labels, reflecting the adversarial brittleness of the underlying vision encoder and raising reliability concerns in safety-critical deployments \cite{szegedy2013intriguing,papernot2016limitations,carlini2017towards,croce2020reliable}. This issue is particularly relevant for CLIP \cite{radford2021learning}, which is widely used as a \emph{frozen} model for zero-shot classification and is also transferred to downstream tasks such as retrieval, detection, and segmentation \cite{shin2022reco,zhou2022extract,zhang2023simple,zhao2022exploiting,cai2023semantic}. A dominant line of work targets \emph{zero-shot adversarial robustness} by adversarially training or finetuning CLIP’s visual encoder, where adversarial examples are generated during training to optimize a min–max objective \cite{madry2017towards,mao2022understanding,schlarmann2024robust,wang2024pre,li2024language}. Other approaches keep the visual backbone fixed and instead optimize textual prompts under adversarial objectives to improve alignment on perturbed inputs \cite{zhang2024adversarial,zhou2024few}. While these strategies can improve adversarial robustness, they require additional training and repeated adversarial example generation, which can be computationally expensive and may introduce a clean–robust trade-off \cite{zhang2019theoretically}. These limitations motivate defenses that operate directly at test time, aiming to improve \emph{zero-shot} robustness without modifying pretrained models or incurring additional training.

\noindent \textbf{Test-Time Defenses for Zero-Shot Robustness.} 
Test-time defenses aim to improve the adversarial robustness of pretrained CLIP models without modifying model weights. A related line of work on test-time adaptation refines predictions using the test instance and its augmented views, for example by updating textual prompts via entropy minimization across augmentations \cite{shu2022test}, with extensions that diversify views \cite{feng2023diverse} or align test features to natural-image statistics \cite{abdul2023align}. However, these methods primarily target distribution shift rather than adversarial perturbations. Adversarial purification offers another direction by projecting adversarial inputs back toward the clean data manifold using generative models \cite{nie2022diffusion,wang2022guided}, but diffusion-based variants  are computationally expensive at inference. We therefore focus on efficient test-time defenses.

More recent methods improve zero-shot 
adversarial robustness in CLIP by exploiting 
how clean and adversarial inputs respond 
differently to stochastic perturbations. 
Test-Time Counter Attack (TTC) probes inputs 
with weak noise and measures latent-space 
drift, using unusually small representation 
shifts (\emph{false stability}) to identify 
adversarial inputs and trigger a 
counterattack~\cite{xing2025clip}. 
Anchor-guided One-step linear Movement (AOM) 
constructs a gaussian noise-averaged feature anchor 
and interpolates the representation toward 
this anchor to suppress adversarial 
distortions~\cite{tong2025zero}. Robust 
Test-Time Prompt Tuning (R-TPT) combines 
inference-time prompt updates with stochastic 
view aggregation~\cite{sheng2025r}. While 
effective, these methods often introduce a 
clean--robust trade-off, as interventions 
that improve adversarial robustness can also 
distort clean representations. This motivates 
a closer examination of how CLIP's visual 
representations behave across noise strengths 
and how this understanding can guide more 
selective test-time interventions.

\section{Method}
\subsection{Preliminaries}
\label{sec:prelim}

\paragraph{CLIP zero-shot classification.}
CLIP performs zero-shot recognition by matching 
images and text in a shared embedding space. 
Given an input image $x$ and class names 
$\{c_k\}_{k=1}^{K}$, each class is converted 
into a natural-language prompt using a template 
$T(\cdot)$, producing prompts $P_k = T(c_k)$. 
The image is encoded by the visual encoder 
$\mathcal{F}_v(\cdot)$ and the prompts are 
encoded by the text encoder $\mathcal{F}_t(\cdot)$. 
The resulting features are $\ell_2$-normalized as
\[
f_v = \frac{\mathcal{F}_v(x)}{\|\mathcal{F}_v(x)\|}, 
\qquad
f_t^{k} = 
\frac{\mathcal{F}_t(P_k)}{\|\mathcal{F}_t(P_k)\|}.
\]
Zero-shot prediction is obtained by computing 
cosine similarity between the visual feature 
and the textual features, followed by a softmax 
with temperature $t$:
\begin{equation}
p(y{=}k \mid x)=
\frac{\exp\big(\cos(f_v,f_t^{k})/t\big)}
{\sum_{j=1}^{K}\exp\big(\cos(f_v,f_t^{j})/t\big)}.
\label{eq:clip_zs}
\end{equation}
The predicted label corresponds to the class 
with the highest probability.

\paragraph{Adversarial attacks.}
Given a clean image $x$ with label $y$, an 
adversarial example is constructed as 
$x' = x+\delta$, where $\delta$ is constrained 
to be small (e.g., $\|\delta\|_{p}\le \epsilon$) 
but optimized to induce misclassification. 
Typically, $\delta$ is obtained by maximizing 
a loss defined on CLIP's zero-shot logits:
\begin{equation}
\delta^{*}=\arg\max_{\|\delta\|_{p}\le \epsilon}
\;\mathcal{L}(x+\delta, y),
\label{eq:adv_attack}
\end{equation}
where $\mathcal{L}$ is typically cross-entropy, 
solved iteratively using 
projected gradient descent 
(PGD)~\cite{madry2017towards}. Beyond logit-based attacks, we also 
consider \emph{vision-only} feature attacks 
to evaluate the generality of our approach: 
these perturb only the visual branch by 
maximizing the discrepancy between clean and 
perturbed visual 
representations~\cite{hu2024firm}. Let 
$\mathcal{F}_v^{(\ell)}(\cdot)$ denote the 
feature at layer $\ell$ of the vision encoder, 
and let $\mathcal{S}$ be a set of probed layers. 
We define:
\begin{equation}
\delta^{*}=\arg\max_{\|\delta\|_{p}\le \epsilon}
\;\sum_{\ell \in \mathcal{S}} w_\ell\,
D\!\Big(\mathcal{F}_v^{(\ell)}(x),\; 
\mathcal{F}_v^{(\ell)}(x+\delta)\Big),
\label{eq:feature_attack}
\end{equation}
where $D(\cdot,\cdot)$ is a feature-distance 
objective (e.g., cosine or $\ell_2$), and 
$w_\ell$ are weights (uniform when unspecified) 
applied for each layer.
\vspace{-1em}

\begin{figure}[t]
    \centering
    \includegraphics[width=\linewidth]{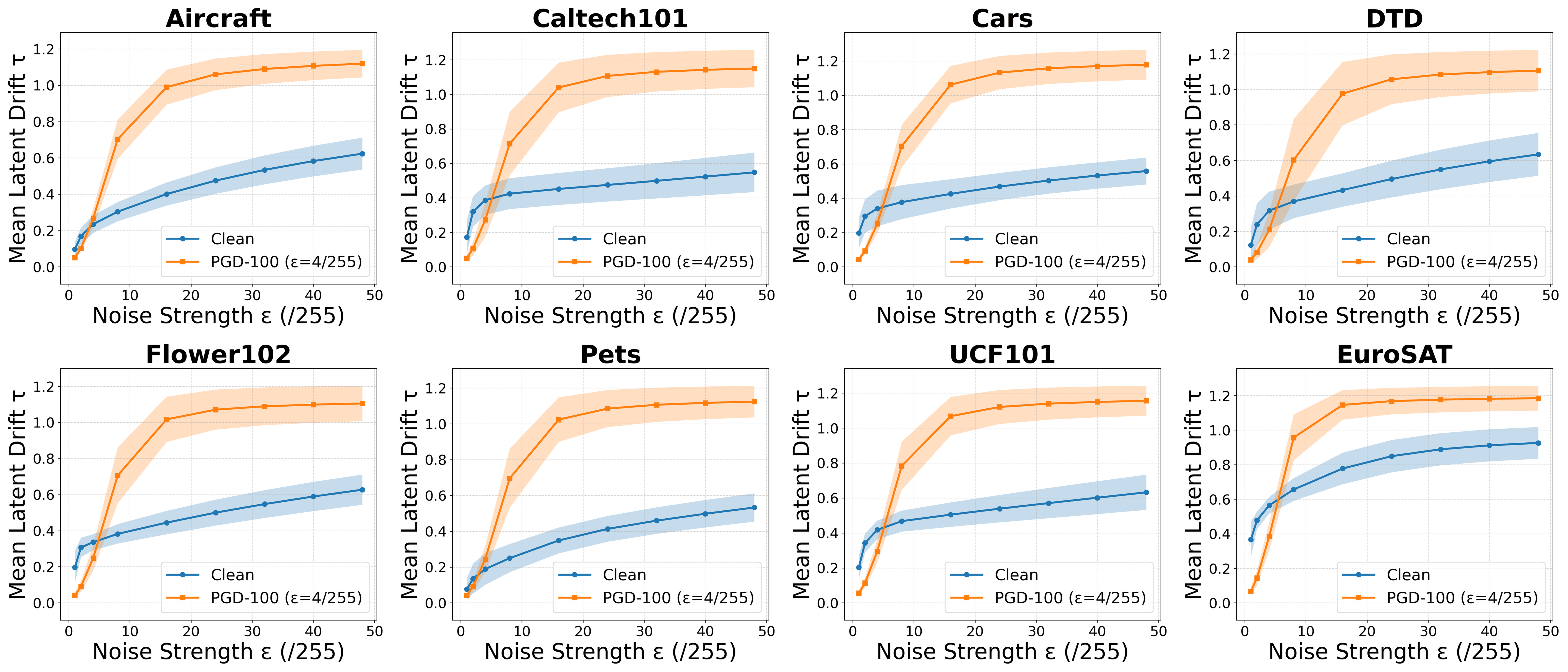}
        \vspace{-1em}
    \caption{Mean latent drift ($\tau$) versus uniform noise strength ($\epsilon$) for clean and adversarial samples across eight fine-grained datasets. Under weak noise, adversarial samples exhibit slightly lower drift than clean samples (\emph{false stability}). As noise strength increases, the curves cross and adversarial drift becomes substantially larger than clean drift, yielding a reliable high-noise separation signal.}
    \label{fig:uniform_noise_tau_anlalysis_eps4}
    \vspace{-1em}

\end{figure}

\begin{figure}[t]
\centering

\includegraphics[width=0.48\linewidth, height=0.22\textheight,keepaspectratio]{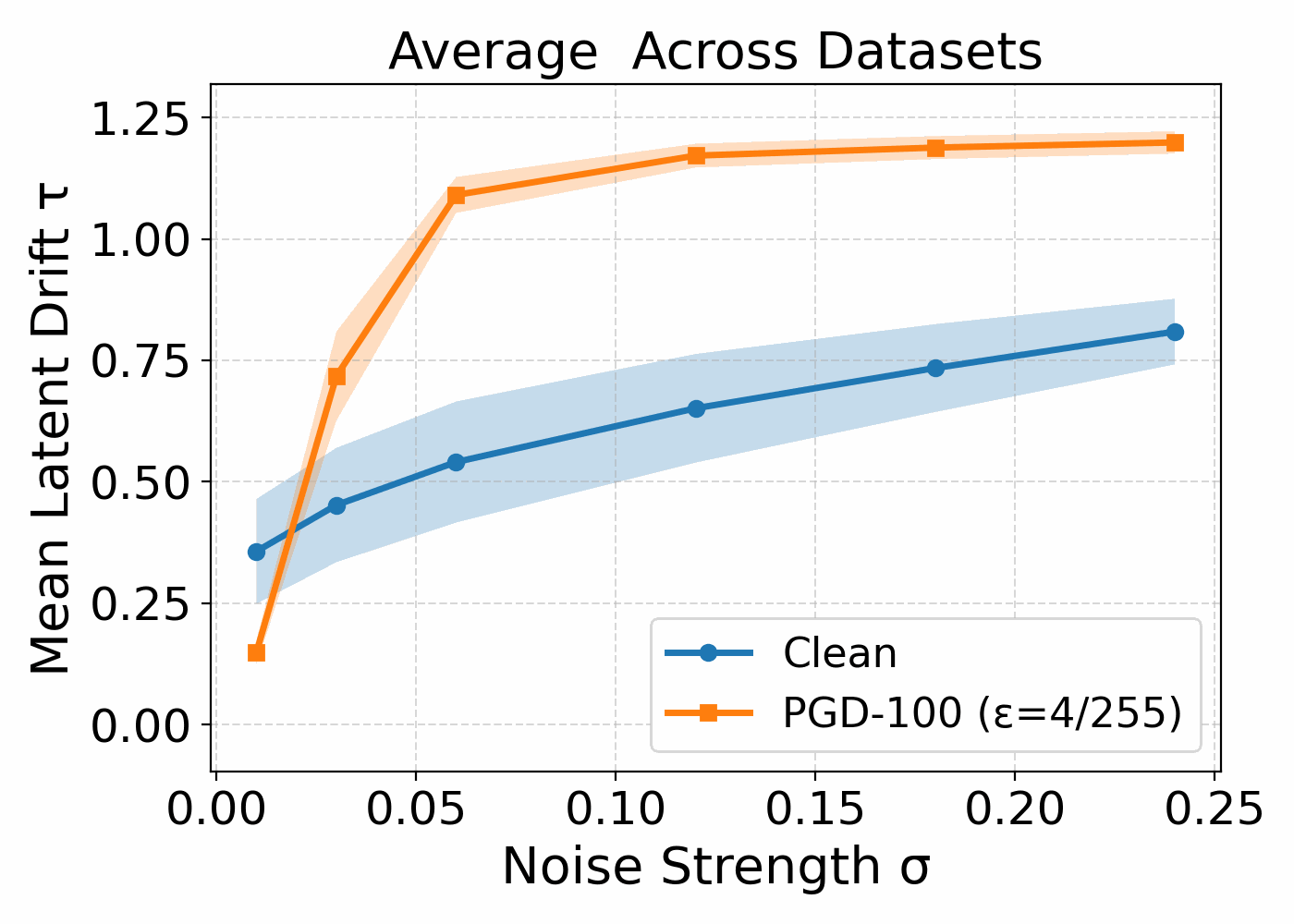}
\hfill
\includegraphics[width=0.48\linewidth, height=0.22\textheight,keepaspectratio]{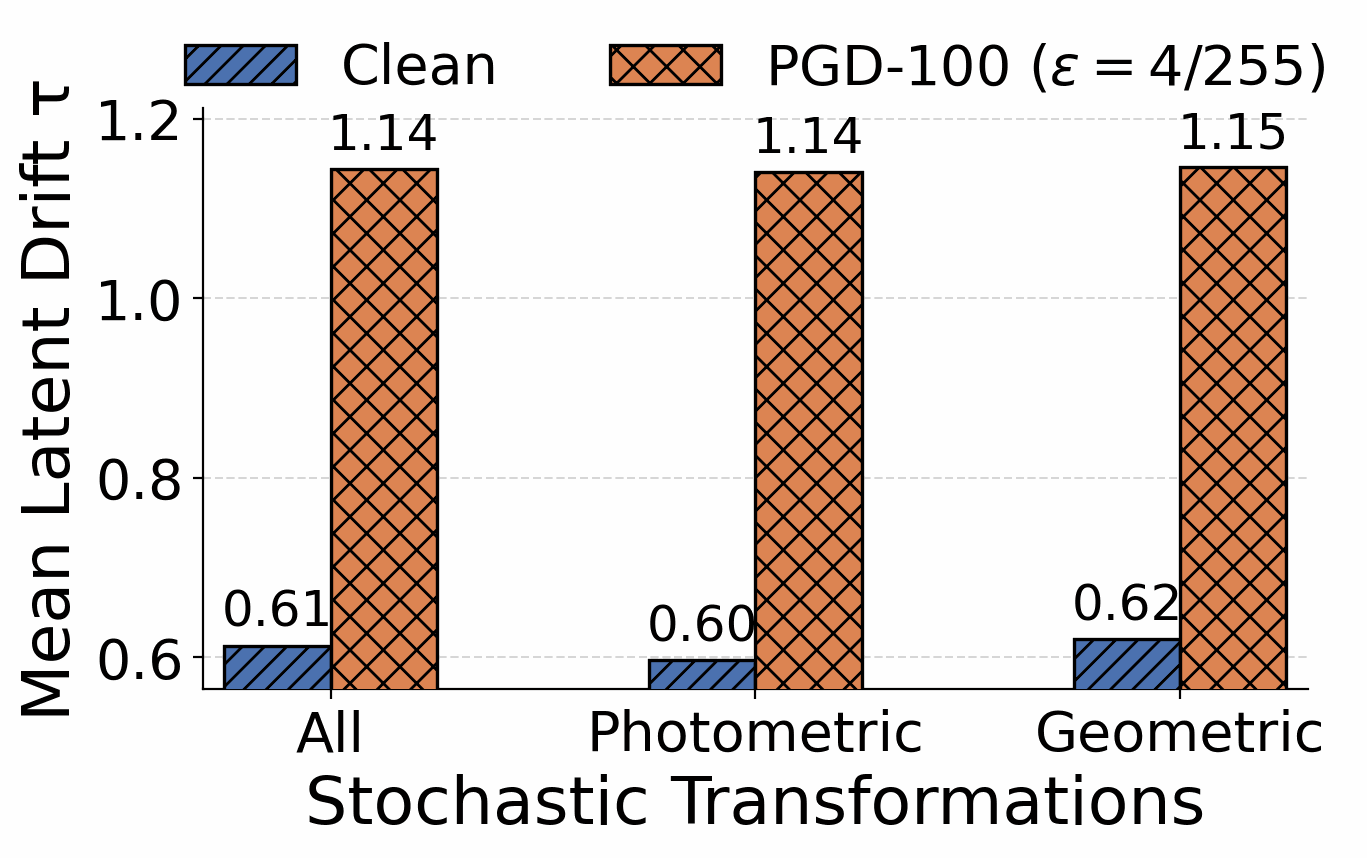}
\vspace{-1em}

\caption{Mean latent drift ($\tau$) under Gaussian noise\emph{(left)} and diverse stochastic transformations \emph{(right)},  averaged across eight fine-grained datasets. The left panel shows the same regime transition as uniform noise, moving from low-noise \emph{false stability} to high-noise separability. The right panel confirms that this separation is not specific to additive noise: it also appears under high-strength photometric, geometric, and combined transformations.}
\label{fig:tau_anlaysis_gaussian_and_tpt}
\vspace{-1.5em}
\end{figure}

\begin{figure}[t]
\centering
\includegraphics[width=\linewidth]{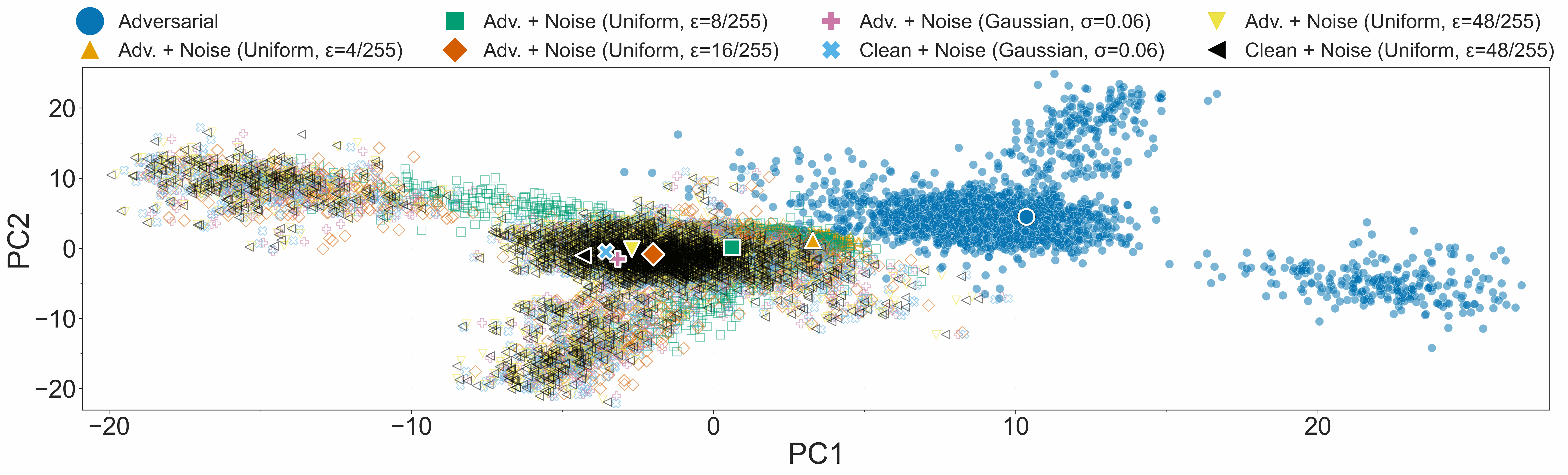}


\includegraphics[width=0.65\linewidth]{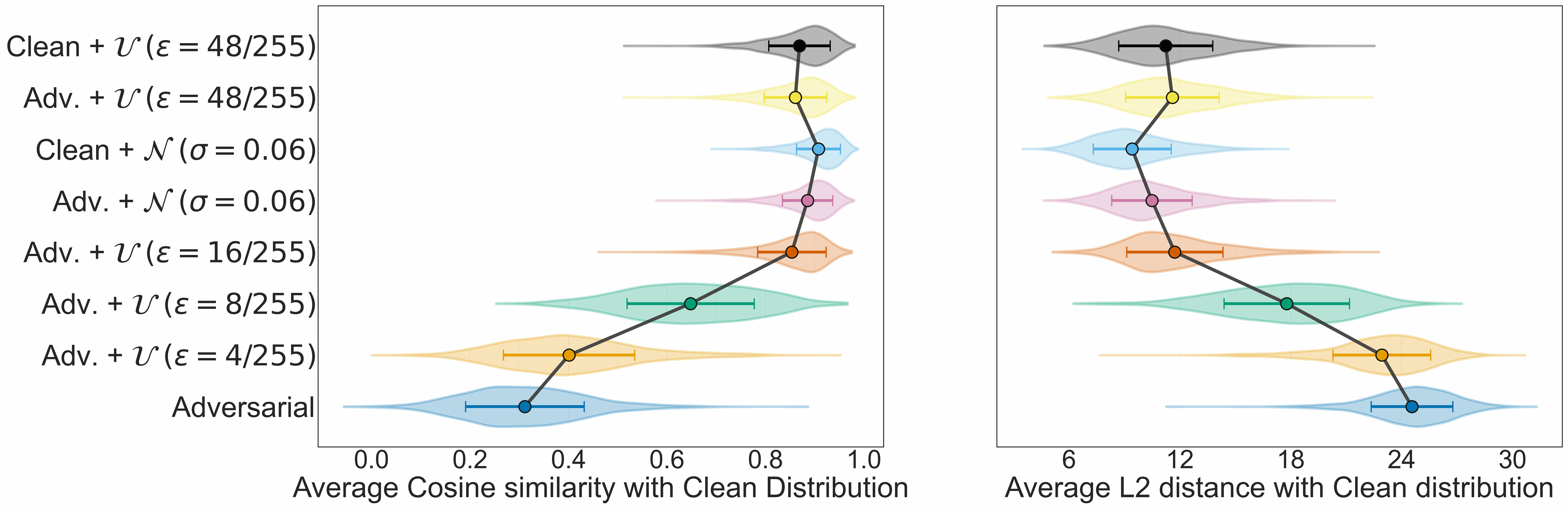}
\hfill
\includegraphics[width=0.33\linewidth]{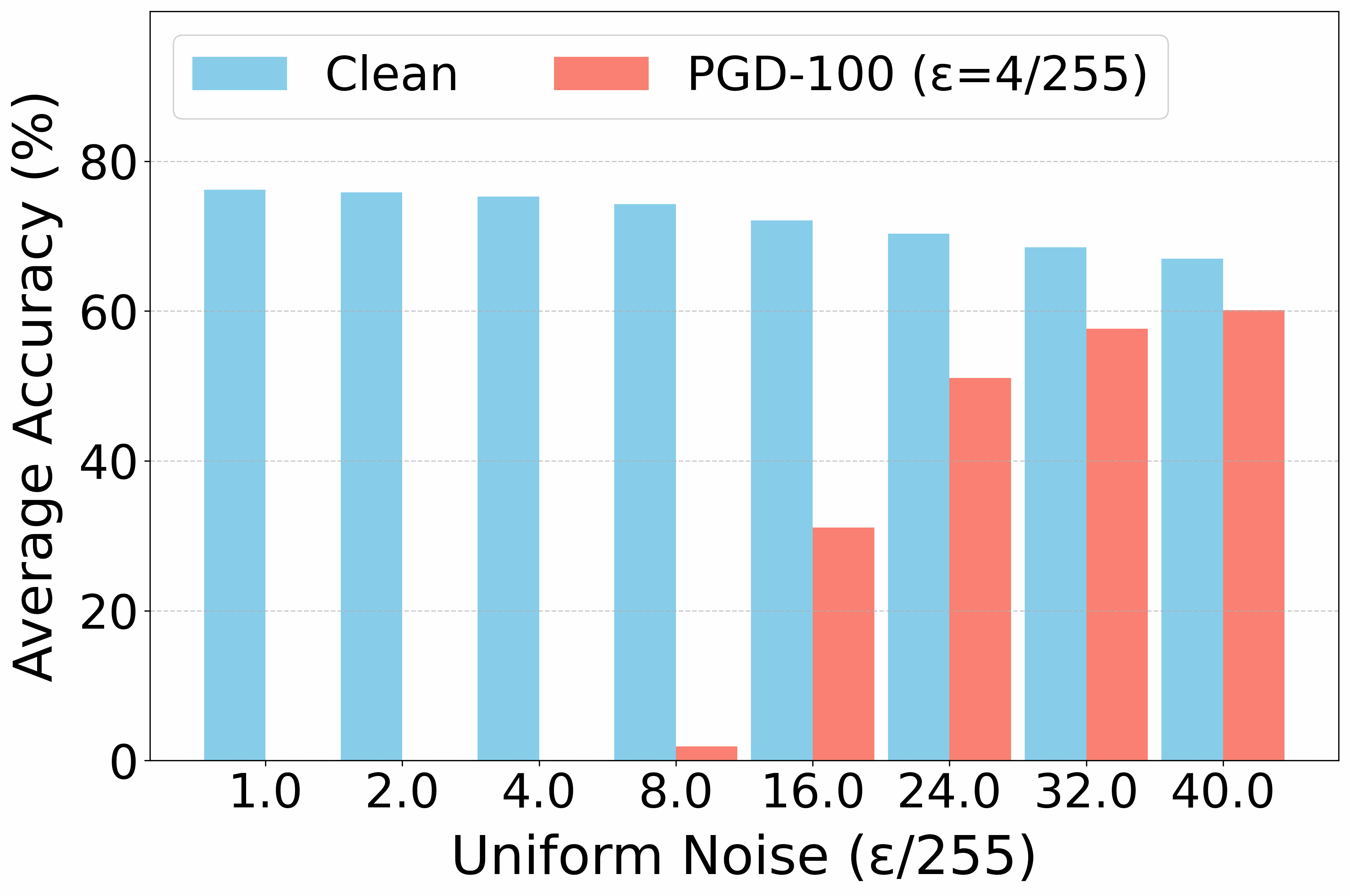}

\caption{\emph{Top:} PCA projection of CLIP 
visual features. Adversarial samples (blue dots) 
are displaced far from the origin into isolated 
local basins. As uniform noise strength increases 
($\epsilon \in \{4, 8, 16, 48\}/255$), mean positions 
of adversarial-plus-noise features migrate progressively toward the 
clean-plus-noise cluster, tracing the escape 
trajectory back toward the clean semantic manifold. 
\emph{Bottom-left and center:} 
Distributions of average cosine similarity and 
$\ell_2$ distance between each condition and the 
clean feature distribution. Adversarial features 
are strongly misaligned; 
both metrics converge monotonically toward the 
clean-plus-noise reference with increasing noise, 
confirming that strong noise injections 
\emph{realign} adversarial features with the 
clean manifold. \emph{Bottom-right:} Average 
clean and adversarial accuracy under increasing 
uniform noise strength, evaluated across eight fine-grained datasets.}
\label{fig:noise_accuracy_and_feature_dynamics}
\vspace{-2em}
\end{figure}

\subsection{Noise-Regime Analysis of Stochastic 
Perturbations}
\label{sec:noise_regime_analysis}

Recent test-time defenses exploit a shared 
behavioral observation: clean and adversarial 
inputs respond differently to stochastic 
perturbations. However, existing 
work~\cite{xing2025clip} has 
only explored this behavior under uniform 
noise of limited strength, leaving it unclear 
how the clean--adversarial distinction evolves 
across noise strengths, noise 
distributions, and transformation types. To 
study this systematically, we define 
\emph{latent drift} in CLIP's visual 
representation space as
\begin{equation}
\tau(x) = \big\|\mathcal{F}_v(x) - 
\mathcal{F}_v(\mathcal{T}(x))\big\|_2,
\label{eq:latent_drift}
\end{equation}
where $\mathcal{T}(\cdot)$ denotes a stochastic 
perturbation or transformation applied to $x$, 
and $\tau(x)$ measures the magnitude of the 
resulting feature displacement.

Fig.~\ref{fig:uniform_noise_tau_anlalysis_eps4} 
plots the mean latent drift under increasing 
uniform noise strength across eight 
fine-grained datasets. 
Under small perturbations, adversarial samples 
exhibit slightly lower drift than clean samples, 
reproducing the \emph{false stability} effect 
reported in TTC~\cite{xing2025clip}. As 
perturbation strength increases, the curves 
cross: adversarial drift rises and eventually 
becomes substantially larger than clean drift, 
revealing a previously underexplored 
\emph{high-noise regime} with stronger 
clean--adversarial separability. 
Fig.~\ref{fig:tau_anlaysis_gaussian_and_tpt} 
confirms this transition is not specific to 
uniform noise: the same crossover appears 
under Gaussian noise and under photometric and 
geometric stochastic view augmentations 
commonly used in test-time adaptation 
methods~\cite{shu2022test,abdul2023align,
sheng2025r}. Across all these settings, 
adversarial samples consistently exhibit higher 
drift once transformation strength is 
sufficiently large. A more detailed analysis across  datasets, attack budgets ($\epsilon \in \{1/255, 4/255, 8/255\}$), attack types, and model variants is provided in Appendix~\ref{sec:app_mean_latent_drift_analysis}.

\subsection{Geometric Interpretation of the 
Noise Regime}
\label{sec:geometric_interpretation}

The drift patterns in 
Figs.~\ref{fig:uniform_noise_tau_anlalysis_eps4} 
and~\ref{fig:tau_anlaysis_gaussian_and_tpt} can 
be interpreted through the geometry of CLIP's 
visual representation space. Clean images are 
mapped to a broad semantic manifold learned from 
large-scale 
pretraining~\cite{radford2021learning,
liu2024clips}, and moderate stochastic 
perturbations of clean inputs tend to produce 
local movements around the same semantic region, 
preserving the model's 
prediction~\cite{cohen2019certified,
lecuyer2018certified}. Adversarial examples, by 
contrast, are optimized to induce 
misclassification while remaining visually close 
to the clean 
image~\cite{szegedy2013intriguing}, moving 
representations into fragile, locally 
optimized regions that lie off the clean 
semantic 
manifold~\cite{stutz2019disentangling,
madry2017towards}.

This view explains the weak-noise 
\emph{false-stability} 
behavior~\cite{xing2025clip}. When the injected 
perturbation is small, adversarial representations 
remain trapped within their local adversarial 
region, producing limited feature displacement 
and therefore smaller drift than clean samples. 
As perturbation strength increases, however, the 
adversarial region becomes unstable: stronger 
perturbations push adversarial representations 
away from this local region, causing 
substantially larger feature displacement. Clean 
representations, by contrast, continue to move 
locally around the semantic manifold, producing 
the observed drift crossover.

Fig.~\ref{fig:noise_accuracy_and_feature_dynamics}~\emph{(top)}
provides empirical support for this interpretation. 
The PCA plot visualizes how features \emph{move} 
rather than where they are: each point represents 
a drift vector capturing how much and in 
what direction a representation shifted, either 
due to the adversarial attack or the subsequent 
stochastic noise injection. All displacement vectors 
are jointly projected to 2D so that the axes 
capture the principal directions of movement, and 
the relative positions of point clouds reflect 
whether different types of displacement point in 
similar or opposing directions. The adversarial drift vectors cluster far to the 
right, reflecting a large, coherent displacement 
away from the clean manifold. Clean inputs 
perturbed with weak  noise produce small 
drift vectors near the center, consistent with 
local movement within the semantic manifold. 
Under strong uniform noise, however, both clean 
and adversarial inputs produce large drift vectors 
on the opposing left side of the plot --- and 
strikingly, their point clouds \emph{overlap}. 
This convergence confirms that sufficiently strong 
stochastic perturbations return adversarial 
representations to the same region of drift space 
occupied by clean representations under equivalent 
noise. Under weak noise, adversarial recovery 
drifts remain near the center and progressively 
migrate leftward as noise strength increases, 
tracing the gradual escape trajectory back to the 
manifold.

The cosine-similarity and $\ell_2$-distance 
plot in Fig.~\ref{fig:noise_accuracy_and_feature_dynamics}~\emph{(bottom)} quantify the same trend. Adversarial 
features exhibit the lowest cosine similarity 
and highest $\ell_2$ distance relative to the 
clean distribution. As noise strength increases, 
both metrics converge monotonically toward 
clean-plus-noise values, confirming progressive 
realignment with the clean feature region. Clean 
inputs under weak noise maintain consistently 
high similarity and low distance throughout, 
consistent with local manifold movement. This behaviour is further confirmed by 
Fig.~\ref{fig:noise_accuracy_and_feature_dynamics}~\emph{(bottom-right)}, 
which shows that averaged across the eight 
fine-grained datasets, adversarial accuracy 
improves sharply with noise strength while 
clean accuracy degrades only gradually, 
reflecting the selective destabilization of 
adversarial representations.
Together, these observations establish $\tau(x)$ 
as a reliable proxy for adversarial-like 
instability---capturing the cross-region 
displacement that distinguishes adversarial 
inputs from clean ones across noise distributions 
and attack types. This geometric account further 
predicts that adversarial training, which aligns 
clean and adversarial representations in feature 
space, should eliminate this drift 
separation---an observation we verify in 
Sec.~\ref{sec:extended_eval}.


\subsection{Drift-Gated Selective  
Defense}
\label{sec:drift_gated_defense}

Our analysis shows that in the high-noise 
regime, adversarial inputs are consistently 
more unstable than clean inputs. We exploit 
this property as a lightweight 
\emph{gating signal} to selectively 
trigger test-time defenses only when needed.

\paragraph{Gating mechanism.}
Given an input image $x$, we compute latent 
drift $\tau(x)$ under a stochastic 
transformation $\mathcal{T}(\cdot)$ at high 
perturbation strength 
(Eq.~\ref{eq:latent_drift}) and apply a binary 
gating rule:
\begin{equation}
\text{gate}(x) = 
\begin{cases}
\text{activate defense } \mathcal{D} 
& \text{if } \tau(x) > \gamma \\
\text{standard CLIP inference}        
& \text{otherwise,}
\end{cases}
\label{eq:gate}
\end{equation}
where $\gamma$ is a drift threshold. 
Both the probe noise strength and $\gamma$ 
are selected by evaluating gated noise 
injection across a range of perturbation 
strengths and threshold values 
(Sec.~\ref{sec:noise_only_results}), 
choosing the configuration that maximizes 
the clean--adversarial average performance. These values 
are then kept fixed when integrating the 
gate with TTC, AOM, and R-TPT. We observe that the selected values remain near-optimal across different defensive intervention types and datasets (see Appendices~\ref{sec:app_mean_latent_drift_analysis} 
and~\ref{sec:app_eval_noise}).
The full procedure of drift-gated defense is formalized in 
Algorithm~\ref{alg:drift_gate}.


\begin{algorithm}[t]
\caption{Drift-Gated Defense}
\label{alg:drift_gate}
\begin{algorithmic}[1]

\Require Test image $x$; CLIP visual encoder 
         $\mathcal{F}_v$ and text encoder 
         $\mathcal{F}_t$ with class prompts 
         $\{P_k\}_{k=1}^K$; strong-noise transformation 
         $\mathcal{T}_{\epsilon_d}$; instability 
         threshold $\gamma$

\State $\tau(x) \leftarrow 
       \|\mathcal{F}_v(x) - 
       \mathcal{F}_v(\mathcal{T}_{\epsilon_d}(x))
       \|_2$
\Comment{One additional forward pass}

\State $\text{output} \leftarrow 
\begin{cases}
\mathcal{D}(x) & \tau(x) > \gamma \\
\text{Predict}(x; \mathcal{F}_v, \mathcal{F}_t)
               & \text{otherwise}
\end{cases}$

\State \Return $\text{output}$

\Statex \textit{$\mathcal{D}(x)$ denotes the prediction after
defensive intervention.}

\end{algorithmic}
\end{algorithm}


\section{Experiments}

\subsection{Setup}

\noindent\textbf{Datasets.}
We conduct our primary evaluation on eight 
fine-grained classification benchmarks covering 
diverse visual domains: Caltech101 (general 
objects)~\cite{fei2004learning}, Pets 
(animals)~\cite{parkhi2012cats}, Flower102 
(plants)~\cite{nilsback2008automated}, Cars and 
Aircraft 
(vehicles)~\cite{krause20133d,maji2013fine}, DTD 
(textures)~\cite{cimpoi2014describing}, EuroSAT 
(satellite imagery)~\cite{helber2019eurosat}, and 
UCF101 (actions)~\cite{soomro2012dataset}. In 
addition, we verify that the same trends hold on 
ImageNet~\cite{deng2009imagenet} and four 
out-of-distribution (OOD) variants: 
ImageNet-V2~\cite{recht2019imagenet}, 
ImageNet-Sketch~\cite{wang2019learning}, 
ImageNet-A~\cite{hendrycks2021natural}, and 
ImageNet-R~\cite{hendrycks2021many}.

\noindent\textbf{CLIP Models.}
Unless stated otherwise, our main results in Sec.~\ref{sec:main_eval} use 
CLIP ViT-L/14 pretrained on 
DataComp-1B~\cite{liu2024clips}, which offers 
stronger generalization than the original CLIP 
ViT-L/14~\cite{radford2021learning}. We 
additionally evaluate  two adversarially trained CLIP 
variants---FARE~\cite{schlarmann2024robust} 
(adversarially finetuned on ImageNet at 
$\epsilon{=}4/255$) and 
DeltaCLIP-L~\cite{wang2025double} (adversarially 
trained at scale on DataComp-1B)---and report 
those results in 
Sec.~\ref{sec:extended_eval}. For detailed analysis across each model, refer to the Appendix.

\noindent\textbf{Baselines.}
We compare against three recent test-time defenses 
for CLIP: TTC~\cite{xing2025clip}, 
AOM~\cite{tong2025zero}, and 
R-TPT~\cite{sheng2025r}. For TTC, we follow 
the original settings exactly: the counterattack 
perturbation strength is set to 
$\epsilon_{TTC}=4/255$, the number of 
counterattack steps is $N_{TTC}=5$, and 
the false-stability gating threshold 
$\gamma_{TTC}=0.2$ is computed at 
$\epsilon=2/255$, which yields the best overall 
performance (detailed ablations are provided 
in Appendix~\ref{sec:app_eval_ttc}). For 
R-TPT, we use the same configuration and 
hyperparameters reported in the original 
work. For AOM, we follow the same anchor-based 
interpolation pipeline, 
setting $\sigma_{AOM}=0.12$ as it yields the best 
average clean--robust performance, and 
constructing the noisy feature anchor by 
averaging $n_{AOM}=10$ gaussian noise injected samples (detailed ablations are provided 
in Appendix~\ref{sec:app_eval_aom}). 
We report results across multiple interpolation 
strengths $\alpha_{AOM}$.

\noindent\textbf{Adversarial Evaluation.}
Following prior 
work~\cite{schlarmann2024robust,xing2025clip,
sheng2025r,tong2025zero}, adversarial examples 
are generated using PGD-100~\cite{madry2017towards} 
with an $\ell_\infty$ perturbation budget of 
$\epsilon=4/255$ against the base CLIP model 
before applying any test-time defense. This 
reflects a realistic deployment scenario in 
which attackers rely on publicly available 
pretrained models and are unaware of the 
deployed defense---the same protocol used by 
all three baselines~\cite{xing2025clip,
sheng2025r,tong2025zero}. This assumption is 
appropriate here because our drift gate uses 
high-noise probing only as a detection signal; 
the underlying defenses operate on the original 
image without modification. We report clean 
accuracy, adversarial accuracy, and their 
average $\big((\text{clean}+\text{adversarial})/2\big)$. 
Main results in Sec.~\ref{sec:main_eval} are 
evaluated on the eight fine-grained datasets; 
further extended evaluations  are 
provided in Sec.~\ref{sec:extended_eval}.

\subsection{Main Results}
\label{sec:main_eval}

\subsubsection{Probe Strength and Threshold 
Selection.}
\label{sec:noise_only_results}

\begin{figure}[t]
    \small \centering
    \begin{minipage}{0.65\textwidth}
        \centering
        \includegraphics[width=\linewidth]{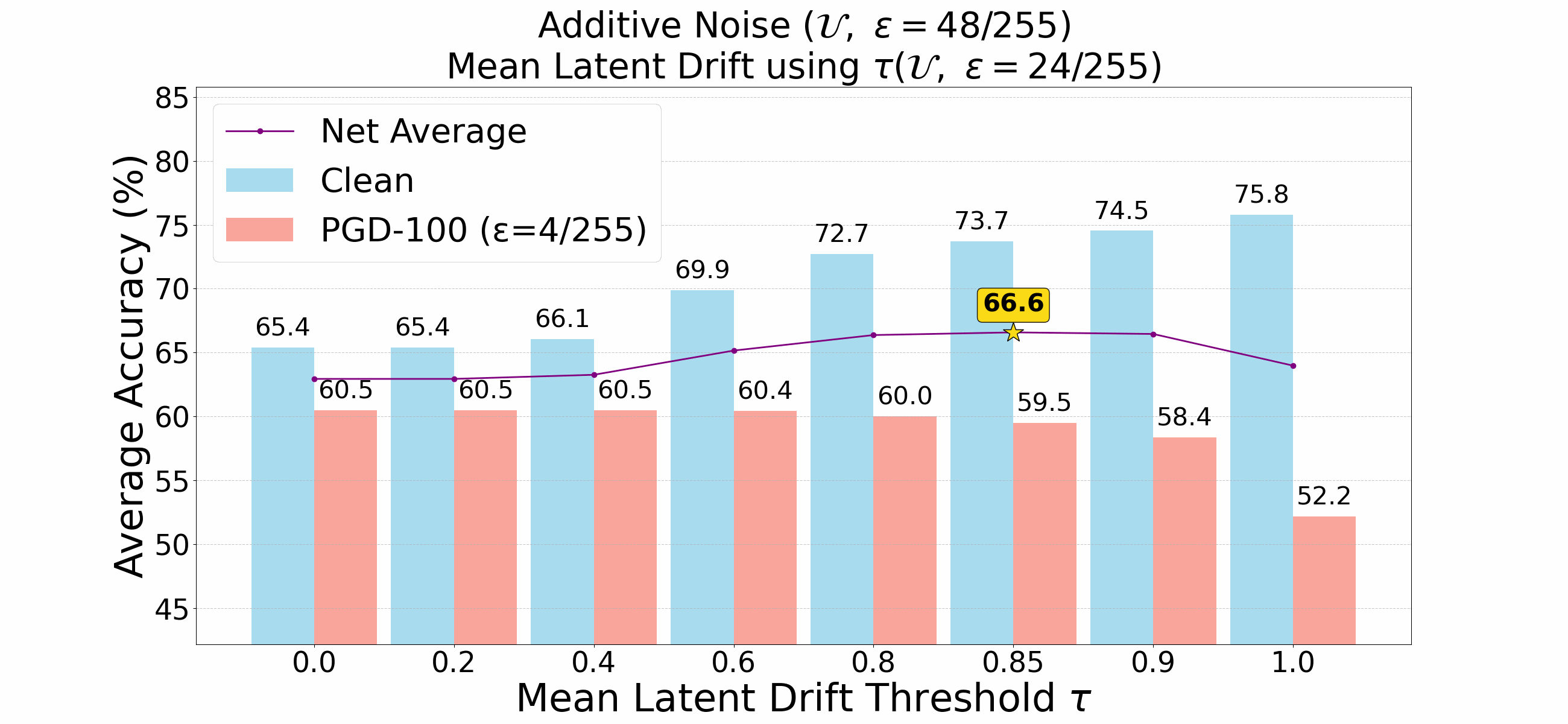}

        \vspace{0.5em}

        \includegraphics[width=\linewidth]{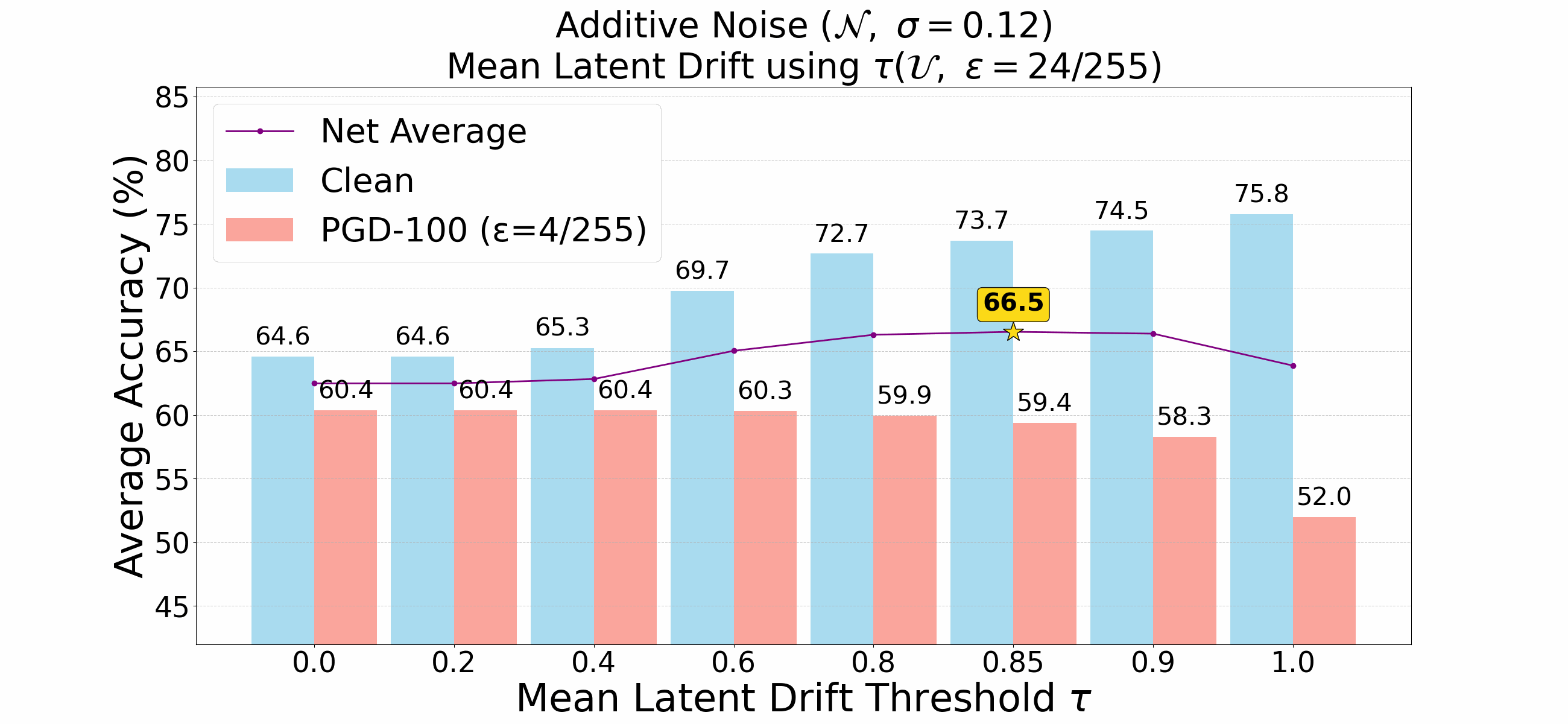}
    \end{minipage}
    \begin{minipage}{0.33\textwidth}
\caption{Threshold $\gamma$ selection via 
drift-gated noise injection, averaged across 
eight fine-grained datasets. Clean (blue), 
PGD-100 at $\epsilon=4/255$ (red), and their 
average (purple) are shown as a function of 
$\gamma$. \textbf{Top:} additive uniform noise 
($\epsilon=48/255$). \textbf{Bottom:} additive 
Gaussian noise ($\sigma=0.12$). $\tau$ is 
computed using a fixed probe noise (Uniform, 
$\epsilon=24/255$). Increasing $\gamma$ applies 
the intervention more selectively, restoring 
clean accuracy while largely preserving 
adversarial robustness; the best trade-off 
occurs near $\gamma \approx 0.85$ (highlighted), 
which is used as the default threshold in all 
subsequent experiments.}
        \label{fig:results_noise_only}
    \end{minipage}
    \vspace{-1.5em}
\end{figure}

This experiment serves two purposes: 
validating that the high-noise regime 
identified in 
Sec.~\ref{sec:noise_regime_analysis} 
translates into practical robustness gains, 
and selecting the gating threshold $\gamma$ 
used throughout all subsequent experiments. 
Results across different probe noise 
strengths for both uniform and Gaussian 
noise are reported in 
Appendix~\ref{sec:app_eval_noise}, where we 
identify the probe strength that yields the 
strongest clean--adversarial performance for 
each noise type; here we fix the probe noise 
and sweep across $\gamma$.

Fig.~\ref{fig:results_noise_only} shows 
clean, adversarial, and average accuracy as 
a function of $\gamma$, averaged across the 
eight fine-grained datasets. In the top plot, $\tau(x)$ is computed 
using a fixed uniform probe noise 
($\epsilon=24/255$) and the defensive 
intervention injects uniform noise 
($\epsilon=48/255$) onto the input. In the 
bottom plot, $\tau(x)$ is computed using a 
fixed Gaussian probe noise ($\sigma=0.12$) 
and the defensive intervention injects 
Gaussian noise ($\sigma=0.12$) onto the 
input. In both cases, we sweep across different 
values of $\gamma$ to find the threshold 
applied to $\tau(x)$ that best decides 
whether to inject noise onto the input.  Without 
any defense, zero-shot CLIP achieves 
$76.1\%$ clean accuracy, but adversarial accuracy drops to 
$0.01\%$. Applying the noise injection to all inputs 
without gating ($\gamma=0$) already recovers 
adversarial accuracy to $60.5\%$ (uniform) 
and $60.4\%$ (Gaussian), confirming the 
practical effectiveness of high-strength 
noise---in contrast to prior 
work~\cite{xing2025clip}, which 
applies random noise at low strength similar 
to the attack budget and observes limited 
robustness gains.
Increasing $\gamma$ applies the intervention 
more selectively, progressively recovering 
clean accuracy while largely preserving 
adversarial robustness. The best trade-off 
occurs at $\gamma\approx0.85$ for both 
settings, yielding peak averages of $66.6\%$ 
(uniform) and $66.5\%$ (Gaussian). Unless otherwise stated, we use 
$\gamma=0.85$ and compute latent drift $\tau(x)$ using 
uniform probe noise at $\epsilon=24/255$ 
across all subsequent experiments.

\subsubsection{Improving  Current Noise-based Test-Time Defenses.}
We next evaluate how our approach improves the clean–robust trade-off of existing test-time defenses by applying interventions only when an input exhibits adversarial-like instability. Detailed results for each method are 
provided in 
Appendices~\ref{sec:app_eval_ttc}, 
\ref{sec:app_eval_aom}, 
and~\ref{sec:app_eval_rtpt}.

\noindent\textbf{Test-Time Counterattack (TTC).}
TTC~\cite{xing2025clip} detects adversarial 
inputs using a weak-noise \emph{false stability} 
trigger and applies a PGD-based counterattack 
that maximizes the $\ell_2$ distance between 
the perturbed and original embeddings:
\begin{equation}
\delta^{*} = 
\arg\max_{\|\delta\|_p \leq \epsilon_{\text{TTC}}} 
\|\mathcal{F}_v(x + \delta) - 
\mathcal{F}_v(x)\|_2.
\label{eq:ttc}
\end{equation}
The weak-noise trigger is unreliable and 
incorrectly flags clean inputs as adversarial, 
reducing clean accuracy (Table~\ref{tab:ttc_results}, 
rows 1--2). We replace $\tau_{\text{TTC}}$ 
entirely with our high-noise drift gate: the 
counterattack fires only when $\tau(x) > \gamma$; 
otherwise standard CLIP inference is used. 
As shown in Table~\ref{tab:ttc_results}, 
this modification increases average accuracy 
from $65.7\%$ to $71.4\%$ when $\tau(x)$ 
is computed using uniform probe noise 
($\epsilon=24/255$) and to $70.2\%$ when 
computed using Gaussian probe noise 
($\sigma=0.12$), while preserving 
adversarial accuracy in both cases.

\noindent\textbf{Anchor Interpolation (AOM).}
AOM~\cite{tong2025zero} constructs a 
noise-averaged anchor using Gaussian 
corruptions, motivated by the observation that 
Gaussian noise suppresses adversarial 
perturbations, and moves the representation 
toward it via linear interpolation:
\begin{equation}
f_{\text{anchor}} = \frac{1}{n_{\text{AOM}}}
\sum_{i=1}^{n_{\text{AOM}}} 
\mathcal{F}_v(x + \delta_i),\;
\delta_i \sim \mathcal{N}(0,\sigma_{\text{AOM}}^2);
\quad
f_{\text{move}} = (1{-}\alpha_{\text{AOM}})
\mathcal{F}_v(x) + \alpha_{\text{AOM}} 
f_{\text{anchor}}.
\label{eq:aom}
\end{equation}
Applying this uniformly to all inputs distorts 
clean representations at larger 
$\alpha_{\text{AOM}}$: standard AOM at 
$\alpha_{\text{AOM}}=1.4$ reduces clean accuracy 
to $53.9\%$ (Table~\ref{tab:aom_results}). Our 
gate applies the anchor construction and 
interpolation only when $\tau(x) > \gamma$, 
preserving clean features exactly. Furthermore, 
consistent with our finding that adversarial 
suppression is not specific to Gaussian noise, 
we show that uniform noise anchors of suitable 
strength yield comparable gains. At 
$\alpha_{\text{AOM}}=1.4$, our method retains 
$72.9\%$ clean accuracy while achieving $72.6\%$ 
adversarial accuracy, improving average 
performance from $64.6\%$ to $72.8\%$ with 
uniform anchors and from $63.8\%$ to $72.7\%$ 
with Gaussian anchors. Detailed results across 
anchor types and strengths are provided in 
Appendix~\ref{sec:app_eval_aom}.

\noindent\textbf{Robust Test-Time Prompt Tuning.}
TPT-based approaches are known to improve 
clean performance over zero-shot CLIP through 
prompt adaptation at test time~\cite{shu2022test,sheng2025r}. As shown in 
Table~\ref{tab:rtpt_results}, vanilla 
ensembling of augmented views generated in the TPT framework already provides 
a substantial robustness gain, 
and R-TPT's reliability-based weighted 
ensembling further refines this by 
downweighting adversarial inputs during 
prediction ($55.09\% 
\rightarrow 61.55\%$ adversarial accuracy). However, the adversarial gains 
from the prompt tuning objective itself are 
minimal: comparing TPT ($61.09\%$) and R-TPT 
($61.92\%$) against weighted ensembling alone 
($61.55\%$) shows that the prompt tuning loss 
contributes negligibly to adversarial 
robustness. The robustness gains are thus 
driven almost entirely by ensembling 
predictions across augmented views, which has 
a natural ceiling---augmented views alone 
cannot sufficiently correct large adversarial 
feature displacements.

We address this limitation by augmenting 
TPT-based approaches with our drift-gated TTC 
intervention: when $\tau(x) > \gamma$, a 
TTC-style counterattack (Eq.~\ref{eq:ttc}) is 
applied to a sample $x$ before passing it 
 into the prompt tuning pipeline; 
otherwise $x$ is passed unchanged. This 
directly corrects adversarial feature 
displacements before prompt tuning, dramatically 
improving adversarial accuracy  
while clean accuracy is largely preserved. 

\begin{table}[!t]
	\centering\small
		\setlength{\tabcolsep}{14pt}
        \scalebox{0.85}[0.85]{
\begin{tabular}{lccc}
\toprule
Method & Clean(\%) & Adversarial(\%) & Average(\%) \\
\midrule
\rowcolor{gray!12}
TTC~\cite{xing2025clip} (w/o thresholding) & 21.3 & 70.9 & 46.1 \\
\rowcolor{gray!12}
TTC~\cite{xing2025clip} (w/ thresholding)  & 61.7 & 69.7 & 65.7 \\
\midrule
TTC (Ours; $\tau$ Uniform $\epsilon=24/255$) & 73.2 & 69.6 & \textbf{71.4$_{(+5.7)}$} \\
TTC (Ours; $\tau$ Gaussian $\sigma=0.12$)    & 70.5 & 69.8 & \textbf{70.2$_{(+4.5)}$} \\
\bottomrule
\end{tabular}
}

\caption{Replacing TTC's weak-noise 
\emph{false-stability} gating threshold with 
our high-noise drift gate substantially 
improves clean accuracy from $61.7\%$ to 
$73.2\%$ while preserving adversarial accuracy, 
averaged across eight fine-grained datasets.}
\label{tab:ttc_results}
\vspace{-1em}
\end{table}
\begin{table}[!t]
	\centering\small
		\setlength{\tabcolsep}{7pt}
		\scalebox{0.85}[0.85]{
\begin{tabular}{lccc|ccc}
\toprule
& \multicolumn{3}{c}{\textbf{Uniform Anchor ($\epsilon_{AOM}=48/255$)}} 
& \multicolumn{3}{c}{\textbf{Gaussian Anchor ($\sigma_{AOM}=0.12$)}} \\
\cmidrule(lr){2-4} \cmidrule(lr){5-7}
\textbf{$\alpha_{AOM}$} 
& Clean(\%) & Adversarial(\%) & Average(\%) 
& Clean(\%) & Adversarial(\%) & Average(\%) \\
\midrule

\multicolumn{7}{c}{\textbf{AOM}~\cite{tong2025zero}} \\
\midrule
\rowcolor{gray!12}
1.0 & 63.0 & 58.9 & 60.9 & 62.3 & 58.9 & 60.6 \\
\rowcolor{gray!12}
1.2 & 58.6 & 71.5 & 65.1 & 57.5 & 71.1 & 64.3 \\
\rowcolor{gray!12}
1.4 & 55.2 & 73.9 & 64.6 & 53.9 & 73.7 & 63.8 \\

\midrule

\multicolumn{7}{c}{\textbf{AOM (Ours; $\tau$ Uniform $\epsilon=24/255$)}} \\
\midrule
1.0 & 73.3 & 57.9 & 65.6$_{(+4.7)}$ & 73.3 & 57.9 & 65.6$_{(+5.0)}$ \\
1.2 & 72.9 & 70.2 & 71.6$_{(+6.5)}$ & 72.9 & 69.8 & 71.3$_{(+7.0)}$ \\
1.4 & 72.9 & 72.6 & \textbf{72.8$_{(+8.2)}$} 
& 72.9 & 72.4 & \textbf{72.7$_{(+8.9)}$} \\

\bottomrule
\end{tabular}
}

\caption{Standard AOM suffers a pronounced 
clean-accuracy drop as $\alpha_{\text{AOM}}$ 
increases, whereas drift-gated AOM preserves 
clean performance and enables stronger 
interpolation.}
\label{tab:aom_results}
\vspace{-1em}
\end{table}
\begin{table}[!t]
    \centering\small
    \setlength{\tabcolsep}{11pt}
    \scalebox{0.8}[0.8]{
    \begin{tabular}{lccc}
    \toprule
    \textbf{Method} & Clean (\%) & Adversarial (\%) & Average (\%) \\
    \midrule
    \multicolumn{4}{c}{\textbf{Ensembling}} \\
    \midrule
    \rowcolor{gray!12}
    Vanilla Ensembling                 & 74.37 & 55.09 & 64.73 \\
    \rowcolor{gray!12}
    Weighted Ensembling        & 75.40 & 61.55 & 68.48 \\
    \midrule
    \multicolumn{4}{c}{\textbf{Prompt Tuning + Weighted Ensembling}} \\
    \midrule
    \rowcolor{gray!12}
    TPT~\cite{shu2022test}   & 75.80 & 61.09 & 68.45 \\
    \rowcolor{gray!12}
    R-TPT~\cite{sheng2025r}                      & 75.74 & 61.92 & 68.83 \\
    \midrule
    \multicolumn{4}{c}{\textbf{Prompt Tuning + Weighted Ensembling + drift-gated TTC (Ours)}} \\
    \midrule
RTPT + TTC (Ours; $\tau$ Uniform $\epsilon=24/255$) 
& 74.08 & 72.23 & \textbf{73.16$_{(+4.33)}$} \\
TPT + TTC (Ours; $\tau$ Uniform $\epsilon=24/255$) 
& 74.08 & 72.32 & \textbf{73.20$_{(+4.75)}$} \\
RTPT + TTC (Ours; $\tau$ Gaussian $\sigma=0.12$) 
& 73.37 & 72.35 & \textbf{72.86$_{(+4.03)}$} \\
TPT + TTC (Ours; $\tau$ Gaussian $\sigma=0.12$) 
& 73.34 & 72.42 & \textbf{72.88$_{(+4.43)}$} \\
    \bottomrule
    \end{tabular}}
\caption{Prompt tuning contributes negligibly 
to adversarial robustness beyond weighted 
ensembling alone; our drift-gated TTC 
intervention bridges this gap, substantially 
improving adversarial accuracy while 
preserving clean performance across both TPT 
and R-TPT.}
\vspace{-1em}
    \label{tab:rtpt_results}
\end{table}

\subsection{Extended Evaluations}
\label{sec:extended_eval}
We further extend our evaluation across  (i) diverse attack types 
and budgets, (ii) adversarially trained CLIP 
variants, (iii) higher attack budgets, 
ImageNet and OOD variants, and 
(iv) computational overhead.

\begin{figure}[t]
\centering

        \includegraphics[width=0.4\linewidth, height=0.18\textheight,keepaspectratio]{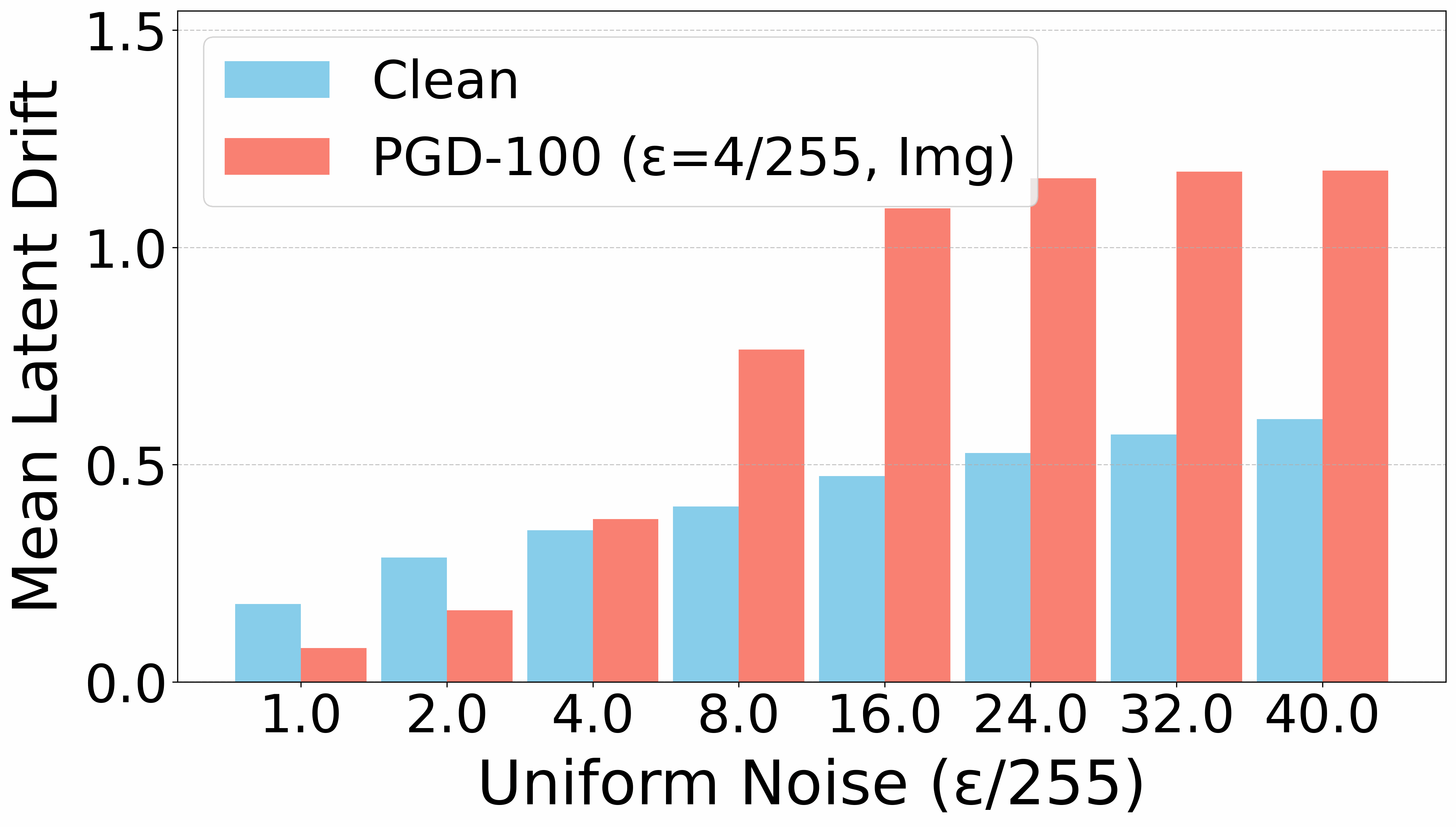}
        \hfill
      \includegraphics[width=0.28\linewidth, height=0.18\textheight,keepaspectratio]
        {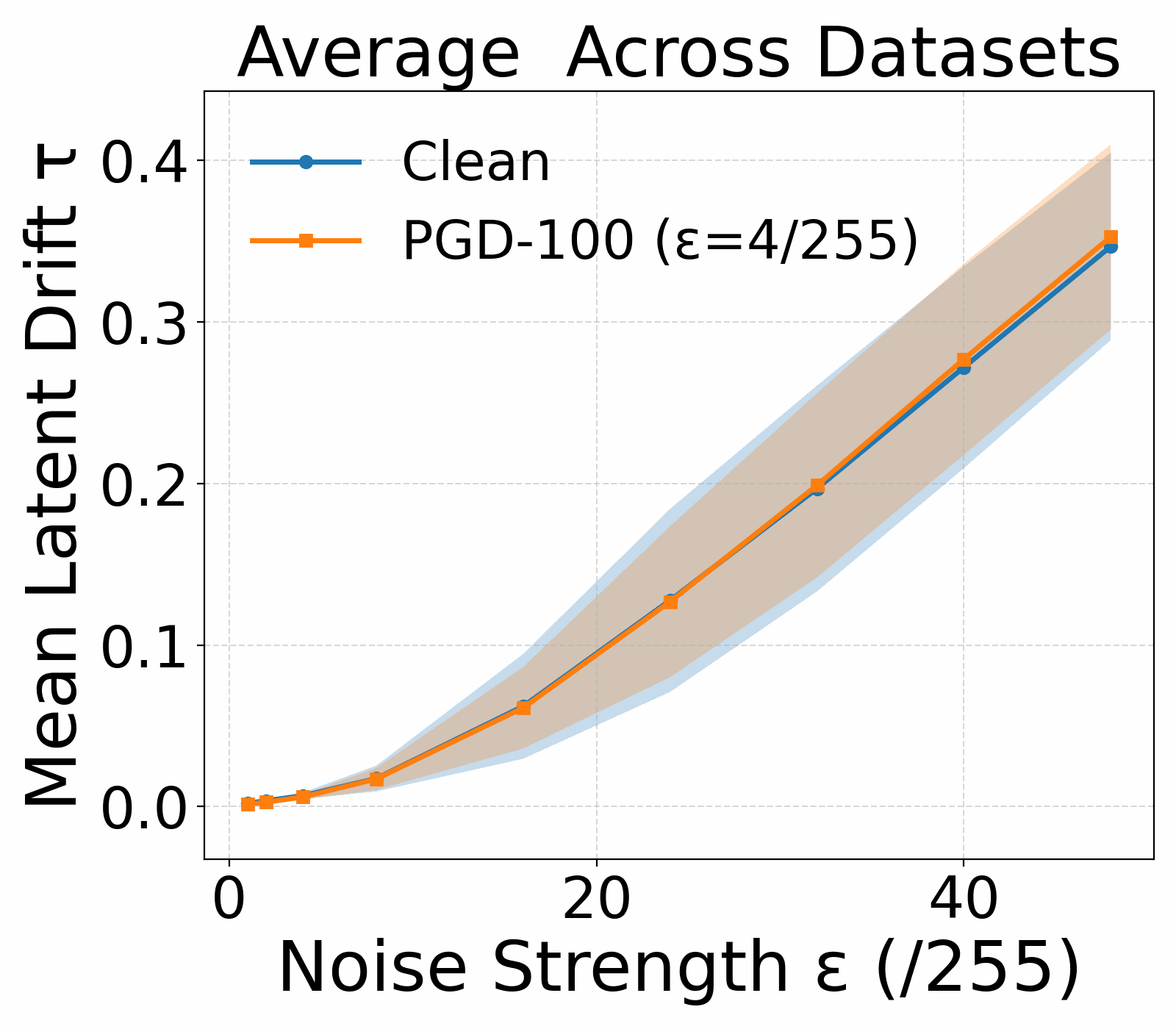}
        \hfill
      \includegraphics[width=0.28\linewidth, height=0.18\textheight,keepaspectratio]
        {figures/robust_models/fare4/a1_mean_tau_curve_eps_4.0_steps_100.png}
\vspace{-1em}
\caption{Mean latent drift under uniform 
noise for a non-robust CLIP model under 
image-only attacks~\cite{hu2024firm} 
\emph{(left)}, and for adversarially 
trained FARE~\cite{schlarmann2024robust} 
and DeltaCLIP-L~\cite{wang2025double} 
\emph{(center, right)}. Results are evaluated across eight fine-grained datasets.}

\label{fig:mean_latent_drift_ablation}
\vspace{-1em}

\end{figure}

\noindent\textbf{Across attack types.} We 
evaluate the best configuration of each 
drift-gated defense from 
Tables~\ref{tab:ttc_results}--\ref{tab:rtpt_results} 
against PGD, 
EOT-PGD~\cite{athalye2018obfuscated}, 
CW~\cite{carlini2017towards}, and 
MI-FGSM~\cite{dong2018boosting} at 
$\epsilon=4/255$, reporting average 
clean--adversarial accuracy in 
Table~\ref{tab:attack_diversity}. Performance 
remains strong across all attack types with 
only modest drops relative to PGD, confirming 
that the high-noise drift signal generalises 
across attack objectives. 
Fig.~\ref{fig:mean_latent_drift_ablation}~\emph{(left)} 
further shows that the drift separation 
persists under image-only feature 
attacks~\cite{hu2024firm}.

\noindent\textbf{Adversarially trained models.}
Fig.~\ref{fig:mean_latent_drift_ablation}~\emph{(center, right)} 
shows that FARE~\cite{schlarmann2024robust} 
and DeltaCLIP-L~\cite{wang2025double} exhibit 
no meaningful clean--adversarial drift 
separation, in stark contrast to non-robust 
CLIP models.  This is 
 consistent with the geometric interpretation in 
Sec.~\ref{sec:geometric_interpretation}: 
 adversarial training encourages clean and 
 adversarial inputs to occupy more similar 
 regions of feature 
 space~\cite{engstrom2019adversarial}. 
As a consequence, adversarial examples no 
longer reside in isolated local basins 
disconnected from the clean 
manifold---consistent with the observation 
that adversarially trained models exhibit 
interpretable gradients and smooth loss 
surfaces~\cite{shafahi2019adversarial}, 
properties that reflect the absence of the 
fragile basin structure that stochastic 
perturbations would otherwise 
destabilize.  Consequently, the drift gate 
is not applicable to these models; 
DeltaCLIP-L and FARE  achieve average
adversarial accuracy of $49.86\%$ and 
$25.86\%$ respectively on fine-grained datasets, and defensive 
interventions can be applied directly without 
gating, yielding modest improvements as 
reported in~\cite{xing2025clip,
sheng2025r}. 

\begin{figure}[htbp]
    \centering
    \begin{minipage}[b]{0.48\linewidth}
        \centering
        \includegraphics[width=\linewidth]
        {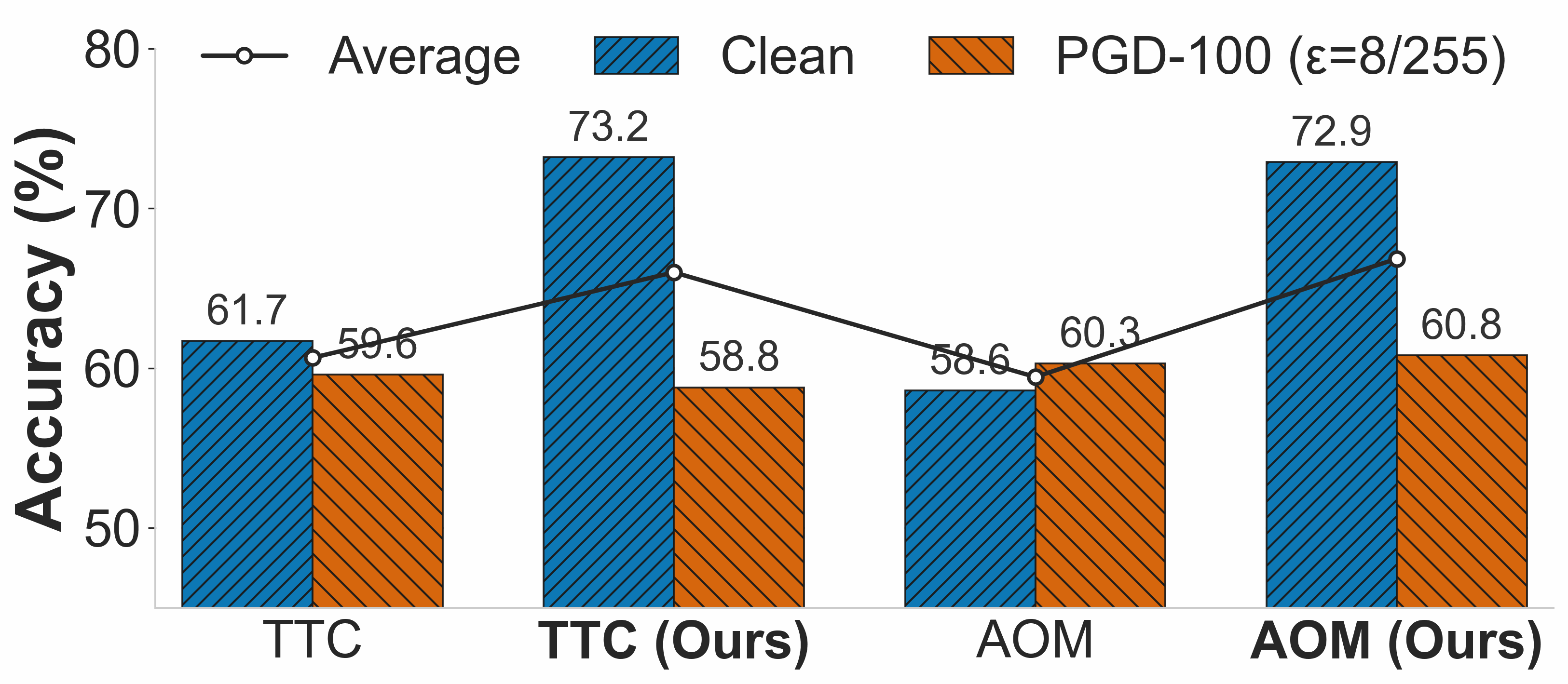}
    \end{minipage}
    \hfill
    \begin{minipage}[b]{0.48\linewidth}
        \centering
        \includegraphics[width=\linewidth]
        {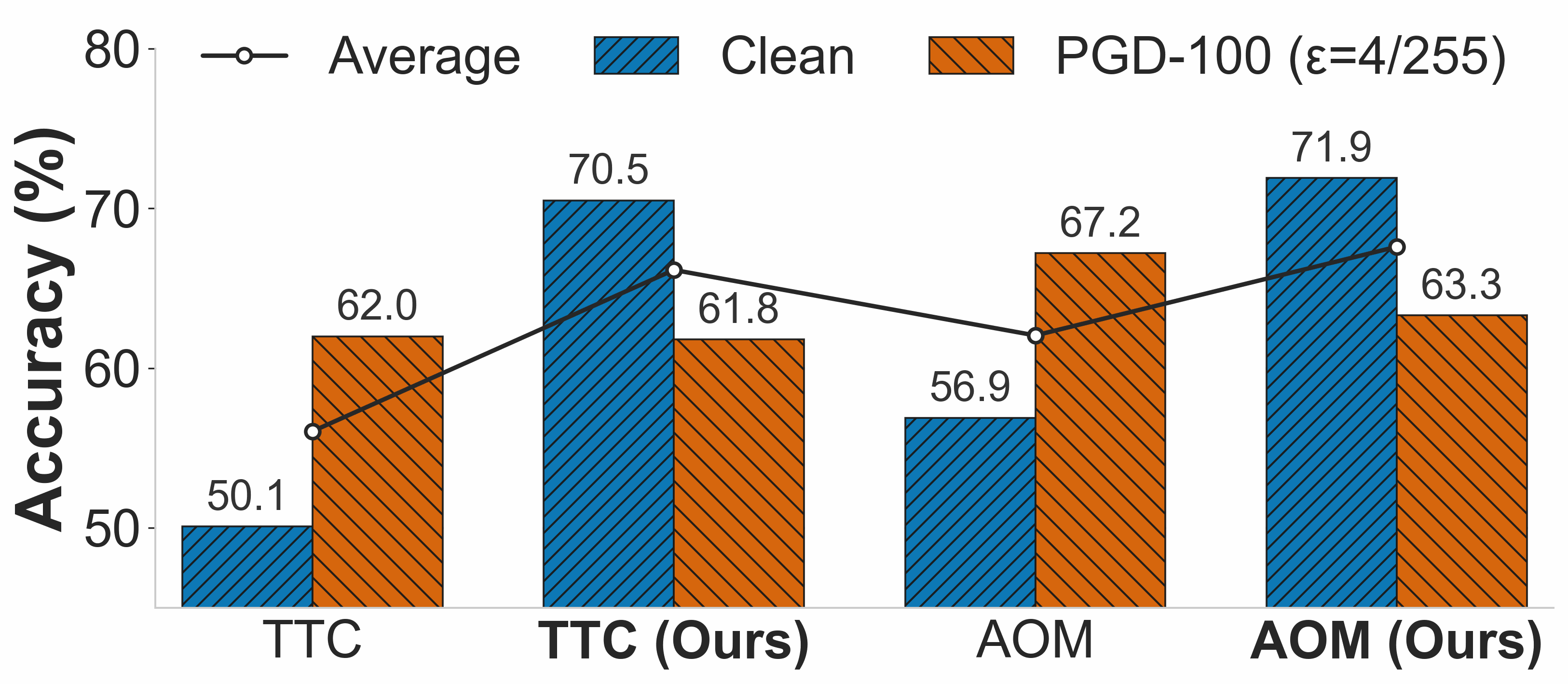}
    \end{minipage}
    \vspace{-1em}
    \caption{Average 
    performance across eight fine-grained 
    datasets under stronger PGD attacks~\emph{(left)}. Average 
    performance across ImageNet and its four 
    OOD variants~\emph{(right)}.}
    \label{fig:main_imagenet_results}
        \vspace{-1.5em}
\end{figure}

\noindent\textbf{Higher attack budget, 
ImageNet and distribution shifts.}
Fig.~\ref{fig:main_imagenet_results}~\emph{(left)} 
shows that our drift-gated strategy maintains 
consistent gains over TTC~\cite{xing2025clip} 
and AOM~\cite{tong2025zero} on the eight 
fine-grained datasets under a stronger attack 
budget ($\epsilon=8/255$), substantially 
recovering clean accuracy while preserving 
adversarial performance. 
Fig.~\ref{fig:main_imagenet_results}\emph{(right)} 
confirms the same trend on ImageNet and its 
four OOD variants ($\epsilon=4/255$), with 
our approach consistently improving the 
clean--robust trade-off across both baselines. 
Latent drift behaviour for these settings is 
detailed in 
Appendix~\ref{sec:app_mean_latent_drift_analysis}.

\noindent\textbf{Computational overhead.}
The computational cost of our approach depends 
solely on the defensive intervention applied, 
plus a single additional forward pass to 
compute the drift threshold. By using a highly 
separable high-noise drift signal, we 
effectively limit expensive defensive 
interventions to inputs exhibiting 
adversarial-like instability, avoiding 
unnecessary processing of clean samples. 
Averaged across the eight fine-grained 
datasets, our gate prevents the defensive 
intervention from being applied to $90.34\%$ 
of clean samples, substantially reducing 
average inference cost compared to applying 
the defense unconditionally.

\begin{table}[!t]
    \centering\small
    \setlength{\tabcolsep}{10pt}
    \scalebox{0.85}[0.85]{
    \begin{tabular}{lcccc}
    \toprule
    \textbf{Method} & \textbf{PGD} & 
    \textbf{EOT-PGD}~\cite{athalye2018obfuscated} & 
    \textbf{CW}~\cite{carlini2017towards} & 
    \textbf{MI-FGSM}~\cite{dong2018boosting} \\
    \midrule
    TTC (Ours)          & 71.4 & 70.4 & 70.8 & 70.2 \\
    AOM (Ours)          & 72.8 & 71.0 & 71.4 & 71.1 \\
    R-TPT + TTC (Ours)  & 73.2 & 70.6 & 71.9 & 70.8 \\
    \bottomrule
    \end{tabular}}
    \caption{Average clean--adversarial performance 
    (\%) across eight fine-grained datasets.
    }
    \label{tab:attack_diversity}
        \vspace{-1.5em}
\end{table}

\section{Discussion and Conclusion}
We identify a noise-regime transition in 
CLIP's visual representation space overlooked 
by prior work: adversarial inputs appear 
falsely stable under weak perturbations but 
become markedly more unstable than clean ones 
as perturbation strength increases. This 
transition is consistent across noise types, 
attack budgets, and datasets, and is explained 
geometrically by adversarial representations 
residing in fragile off-manifold local regions 
that strong perturbations destabilize. Its 
absence in adversarially trained 
variants~\cite{schlarmann2024robust,
wang2025double}---where such regions are 
eliminated---provides  support for this 
interpretation hypothesis.
Building on this finding, we propose a 
training-free drift-gated mechanism that 
selectively activates existing test-time 
defenses only when adversarial-like 
instability is detected, preventing 
unnecessary intervention on $90.34\%$ of 
clean samples. This plug-in strategy 
consistently improves the clean--robust 
trade-off of current 
baselines~\cite{xing2025clip,tong2025zero,
sheng2025r}. Overall, our observations  offer a principled and efficient direction for training-free robustness improvement.

\bibliography{egbib}

\newpage
\appendix

\section{Appendix}
\section*{Overview}

This appendix provides additional analyses, 
ablations, and extended experimental results 
supporting the main paper. We analyze how 
CLIP visual representations respond to 
stochastic perturbations across noise regimes, 
and evaluate how these perturbations influence 
classification performance and the behavior of 
existing test-time defenses.

\paragraph{Mean Latent Drift Analysis (Sec.~\ref{sec:app_mean_latent_drift_analysis}).}

We present a detailed study of mean latent 
drift in CLIP visual representations under 
stochastic perturbations. For ViT-L/14 
(DataComp-1B)~\cite{liu2024clips}, 
Figs.~\ref{fig:app_uniform_noise_tau_anlalysis_eps4_vit_l_14_datacomp}--\ref{fig:app_gaussian_noise_tau_anlalysis_eps8_vit_l_14_datacomp} 
report per-dataset drift curves for clean 
and PGD-100 adversarial inputs (at $\epsilon \in \{4/255, 8/255\}$) under 
increasing uniform and Gaussian noise 
strengths. 
Figs.~\ref{fig:app_uniform_noise_tau_anlalysis_eps_all_vit_l_14_datacomp} 
and~\ref{fig:app_gaussian_noise_tau_anlalysis_eps_all_vit_l_14_datacomp} 
summarize the same trends averaged across 
datasets for multiple attack budgets and 
objectives. Fig.~\ref{fig:app_tpt_aug_tau_anlalysis_eps4_vit_l_14_datacomp} 
extends the analysis to photometric, 
geometric, and combined stochastic 
transformations beyond additive noise.

We additionally report the same latent drift analysis on ImageNet and its out-of-distribution variants in Figures~\ref{fig:app_uniform_noise_tau_anlalysis_eps4_vit_l_14_datacomp_imagenet} and~\ref{fig:app_gaussian_noise_tau_anlalysis_eps4_vit_l_14_datacomp_imagenet}. To verify that the observed drift behavior is not specific to a single CLIP variant, we repeat the analysis for the original CLIP ViT-L/14 model~\cite{radford2021learning} (Figures~\ref{fig:app_uniform_noise_tau_anlalysis_eps4_vit_l_14}–\ref{fig:app_gaussian_noise_tau_anlalysis_eps_all_vit_l_14}). Finally, we extend the study to adversarially trained CLIP variants, including FARE~\cite{schlarmann2024robust}(Figures~\ref{fig:app_uniform_noise_tau_anlalysis_eps4_fare}–\ref{fig:app_gaussian_noise_tau_anlalysis_eps_all_fare}) and DeltaCLIP-L~\cite{wang2025double}(Figures~\ref{fig:app_uniform_noise_tau_anlalysis_eps4_deltaclip}–\ref{fig:app_gaussian_noise_tau_anlalysis_eps_all_deltaclip}), where we examine how latent drift evolves under stochastic perturbations for robust models.

\paragraph{Evaluating Performance with Random Stochastic Transformations (Sec.~\ref{sec:app_eval_noise}).}

This section evaluates how injecting 
stochastic perturbations directly onto input 
samples affects clean and adversarial 
classification accuracy across different CLIP 
model variants. For ViT-L/14 (DataComp-1B), 
Figs.~\ref{fig:app_uniform_noise_accuracy_anlalysis_eps_all_vit_l_14_datacomp} 
and~\ref{fig:app_gaussian_noise_accuracy_anlalysis_eps_all_vit_l_14_datacomp} 
report how clean and adversarial accuracy 
evolve as the strength of uniform and 
Gaussian noise injected onto the input 
samples increases, averaged across the eight 
fine-grained datasets under multiple 
adversarial budgets and attack objectives.

Motivated by the clean--adversarial drift 
separation observed in the high-noise regime, 
Figs.~\ref{fig:results_eval_uniform_noise_threshold_vit_l_14_datacomp} 
and~\ref{fig:results_eval_gaussian_noise_threshold_vit_l_14_datacomp} 
evaluate our drift-gated stochastic 
intervention, where mean latent drift is used 
as a gating signal to selectively inject 
noise onto inputs that exhibit 
adversarial-like instability. These 
experiments show how varying the drift 
threshold $\gamma$ influences the clean--robust 
trade-off under different noise strengths.
The same evaluation is repeated for the 
original CLIP ViT-L/14 
(Figs.~\ref{fig:app_uniform_noise_accuracy_anlalysis_eps_all_vit_l_14}--\ref{fig:app_gaussian_noise_accuracy_anlalysis_eps_all_vit_l_14}), 
and extended to adversarially trained 
variants FARE and DeltaCLIP-L 
(Figs.~\ref{fig:app_uniform_noise_accuracy_anlalysis_eps_all_fare4}--\ref{fig:app_gaussian_noise_accuracy_anlalysis_eps_all_deltaclip}), 
where we examine how clean and adversarial 
accuracy respond to increasing noise injection 
strengths.

\paragraph{Evaluating Performance with Test-Time Counter Attacks (Sec.~\ref{sec:app_eval_ttc}).}
We first evaluate the original TTC 
strategy~\cite{xing2025clip}, which uses 
weak uniform noise to compute a false-stability 
signal $\tau_{\text{TTC}}$ to differentiate 
between clean and adversarial samples and 
trigger the counterattack accordingly. 
Figure~\ref{fig:results_eval_uniform_noise_ttc_default_vit_l_14_datacomp} 
reports results across the eight fine-grained 
datasets, showing the clean--robust trade-off 
as the false-stability threshold varies under 
this weak-noise probing strategy.

We then evaluate how our high-noise drift-gating 
strategy improves upon this. 
Figures~\ref{fig:results_eval_uniform_noise_ttc_vit_l_14_datacomp}--\ref{fig:results_eval_gaussian_noise_ttc_vit_l_14_datacomp} 
report results averaged across the eight 
fine-grained datasets, while 
Figures~\ref{fig:results_eval_uniform_noise_ttc_vit_l_14_datacomp_imagenet} 
and~\ref{fig:results_eval_gaussian_noise_ttc_vit_l_14_datacomp_imagenet} 
report results averaged across ImageNet and 
its four out-of-distribution variants. 
Computing the gating signal in the high-noise 
regime yields a more reliable separation 
between clean and adversarial inputs, 
consistently improving the clean--robust 
trade-off over the original weak-noise 
strategy.

\paragraph{Evaluating Performance with AOM (Sec.~\ref{sec:app_eval_aom}).}
We further evaluate how our drift-gated strategy improves the Anchor-guided One-step linear Movement (AOM) defense~\cite{tong2025zero}. Figures~\ref{fig:results_eval_uniform_anchor_aom_vit_l_14_datacomp} and~\ref{fig:results_eval_gaussian_anchor_aom_vit_l_14_datacomp} report results averaged across the eight fine-grained datasets, while Figure~\ref{fig:results_eval_uniform_gaussian_anchor_aom_vit_l_14_datacomp_imagenet} reports results averaged across ImageNet and its four out-of-distribution variants. These experiments compare the original AOM formulation—which interpolates all samples toward a noisy anchor—with our drift-gated variant that selectively applies the interpolation based on mean latent drift. By computing the gating signal in the high-noise regime, our approach improves the clean–robust trade-off by avoiding unnecessary interventions on clean inputs.

\paragraph{Evaluating Performance with R-TPT (Sec.~\ref{sec:app_eval_rtpt}).}
Finally, we evaluate how our drift-gated strategy improves the Robust Test-Time Prompt Tuning (R-TPT) defense~\cite{sheng2025r}. Table~\ref{tab:rtpt_ttc_dataset_results} reports dataset-wise clean and adversarial accuracy across the eight fine-grained datasets.  We  augment the original R-TPT pipeline with our mean latent-drift gate to selectively trigger TTC-style counterattacks on inputs exhibiting adversarial-like instability before applying R-TPT. This hybrid strategy preserves the strong clean performance of R-TPT while significantly improving adversarial robustness across datasets.

\subsection{Mean Latent Drift Analysis}
\label{sec:app_mean_latent_drift_analysis}

\subsubsection{ViT-L/14 (DataComp-1B)}

\begin{figure}[t]
    \centering
    \includegraphics[width=\linewidth]{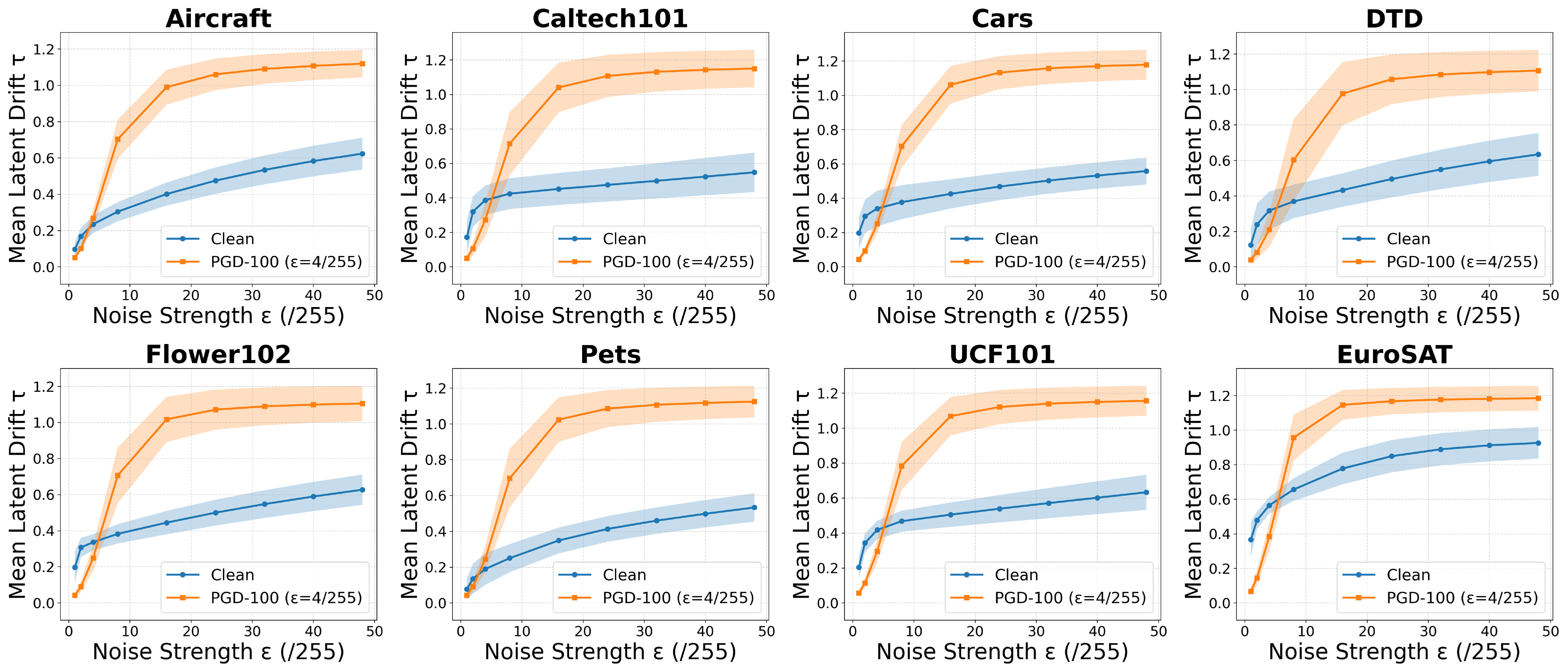}
    \caption{\textbf{ViT-L-14(DataComp-1B) at PGD-100 ($\epsilon=\frac{4}{100}$)}. Mean latent drift ($\tau$) versus uniform noise strength ($\epsilon$) for clean and adversarial samples across eight fine-grained datasets. Under weak noise, adversarial samples exhibit slightly lower drift than clean samples (\emph{false stability}). As noise strength increases, the curves cross and adversarial drift becomes substantially larger than clean drift, yielding a reliable high-noise separation signal.}
    \label{fig:app_uniform_noise_tau_anlalysis_eps4_vit_l_14_datacomp}
    \vspace{-1em}

\end{figure}

\begin{figure}[t]
    \centering
    \includegraphics[width=\linewidth]{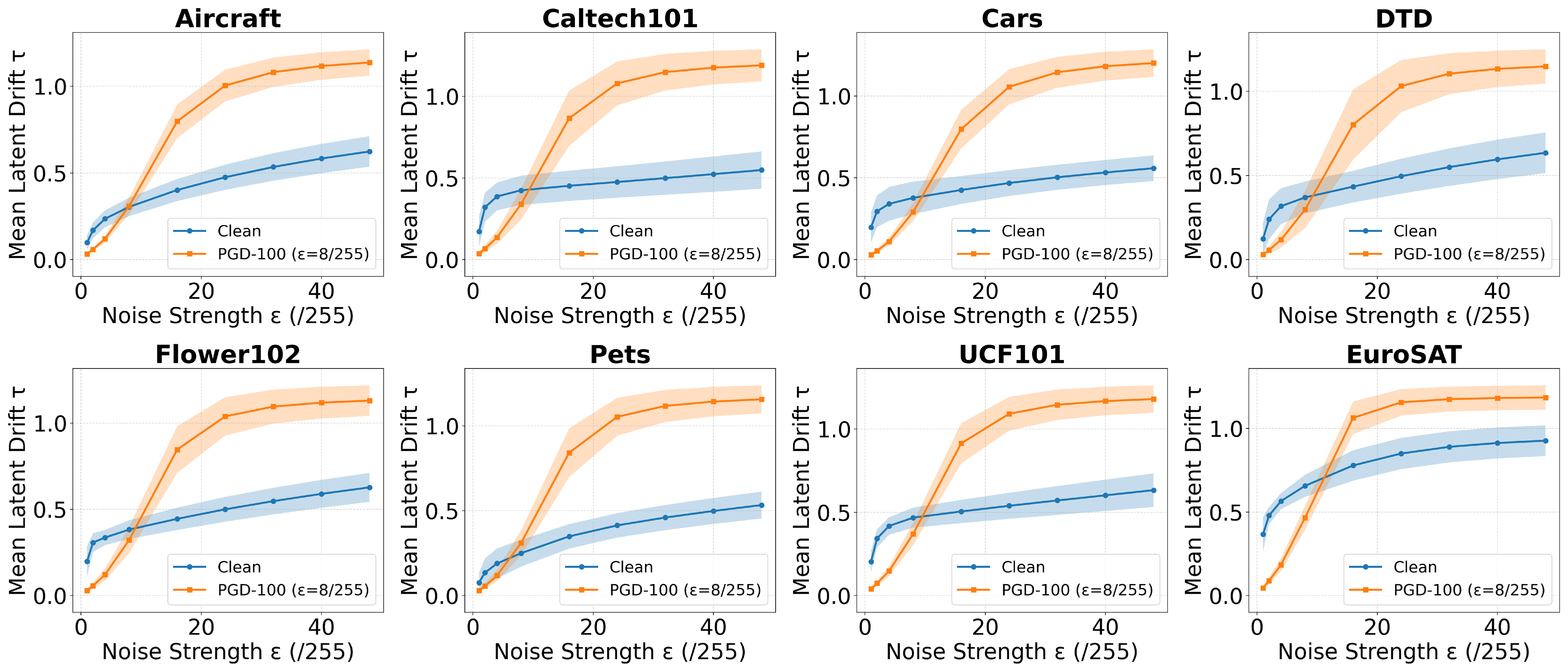}
    \caption{\textbf{ViT-L-14(DataComp-1B) at PGD-100 ($\epsilon=\frac{8}{100}$)}. Mean latent drift ($\tau$) versus uniform noise strength ($\epsilon$) for clean and adversarial samples across eight fine-grained datasets. Under weak noise, adversarial samples exhibit slightly lower drift than clean samples (\emph{false stability}). As noise strength increases, the curves cross and adversarial drift becomes substantially larger than clean drift, yielding a reliable high-noise separation signal.}
    \label{fig:app_uniform_noise_tau_anlalysis_eps8_vit_l_14_datacomp}
    \vspace{-1em}

\end{figure}

\begin{figure}[t]
    \centering
    \includegraphics[width=\linewidth]{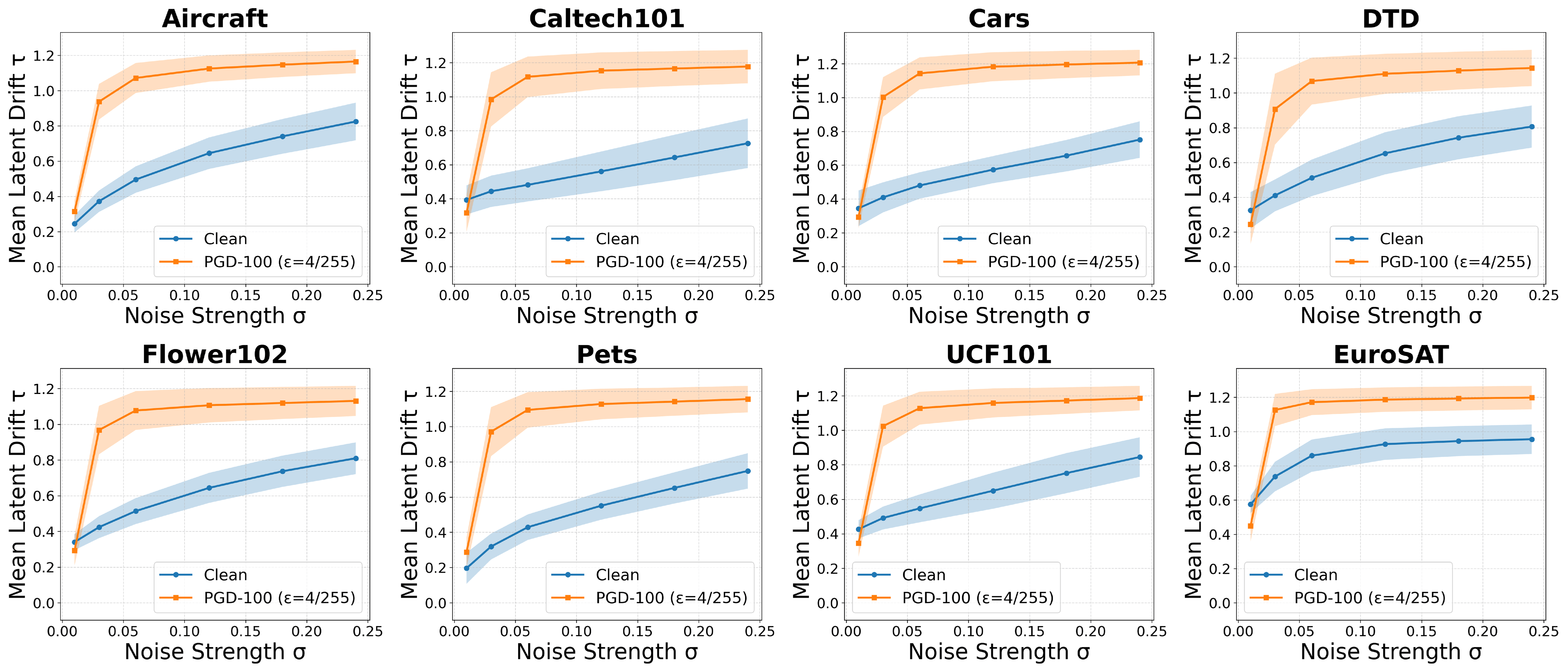}
    \caption{\textbf{ViT-L-14(DataComp-1B) at PGD-100 ($\epsilon=\frac{4}{100}$)}. Mean latent drift ($\tau$) versus gaussian noise strength ($\sigma$) for clean and adversarial samples across eight fine-grained datasets. Under weak noise, adversarial samples exhibit slightly lower drift than clean samples (\emph{false stability}). As noise strength increases, the curves cross and adversarial drift becomes substantially larger than clean drift, yielding a reliable high-noise separation signal.}
    \label{fig:app_gaussian_noise_tau_anlalysis_eps4_vit_l_14_datacomp}
    \vspace{-1em}

\end{figure}

\begin{figure}[t]
    \centering
    \includegraphics[width=\linewidth]{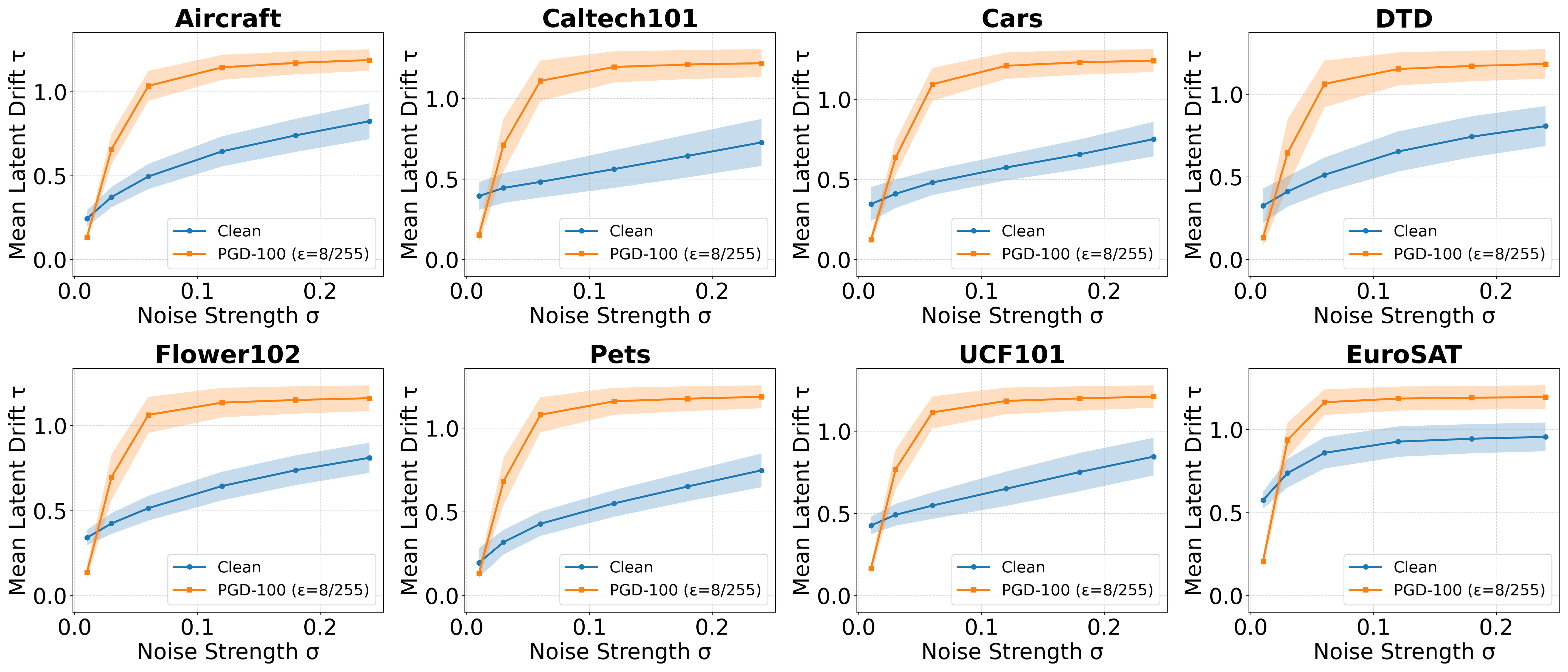}
    \caption{\textbf{ViT-L-14(DataComp-1B) at PGD-100 ($\epsilon=\frac{8}{100}$)}. Mean latent drift ($\tau$) versus gaussian noise strength ($\sigma$) for clean and adversarial samples across eight fine-grained datasets. Under weak noise, adversarial samples exhibit slightly lower drift than clean samples (\emph{false stability}). As noise strength increases, the curves cross and adversarial drift becomes substantially larger than clean drift, yielding a reliable high-noise separation signal.}
    \label{fig:app_gaussian_noise_tau_anlalysis_eps8_vit_l_14_datacomp}
    \vspace{-1em}

\end{figure}

\begin{figure}[t]
    \centering
    \includegraphics[width=\linewidth]{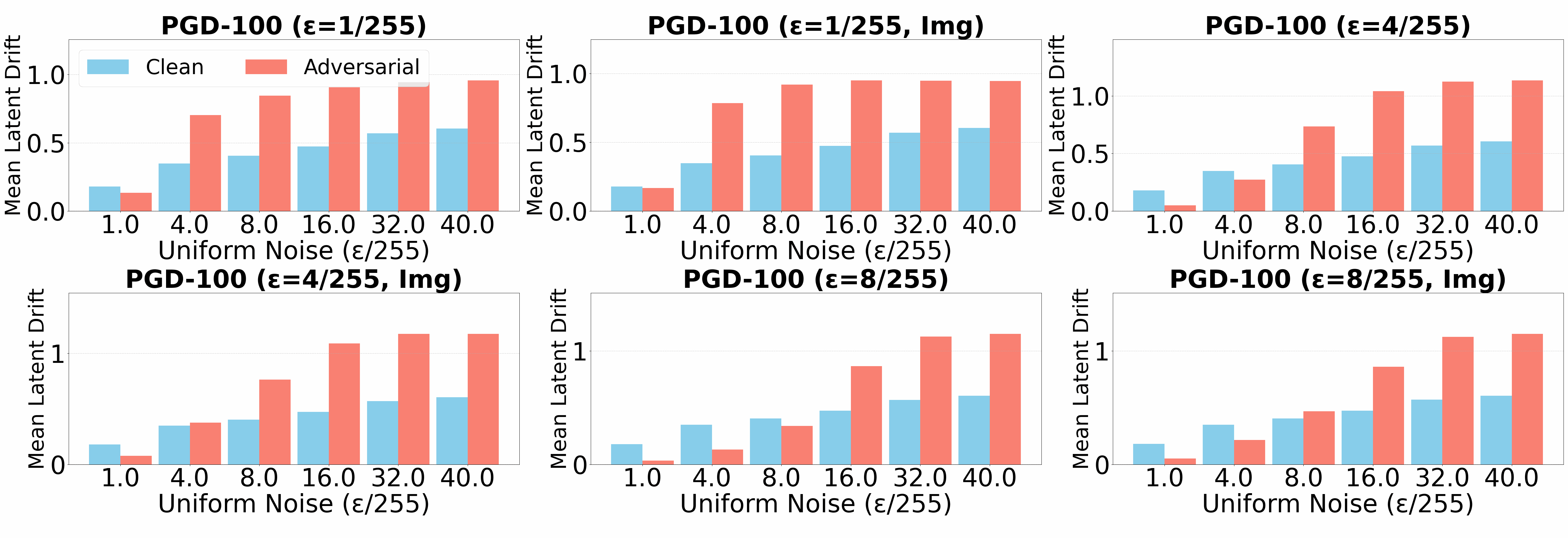}
\caption{\textbf{ViT-L-14(DataComp-1B) at PGD-100 ($\epsilon=\frac{X}{255}$)}. Mean latent drift ($\tau$) versus uniform noise strength for clean and adversarial samples averaged across eight fine-grained datasets. Adversarial examples are generated at different perturbation budgets and using two attack objectives: the standard objective that maximizes cross-entropy loss, and a vision-only objective (Img) that maximizes the discrepancy in visual features. Under weak noise, adversarial samples exhibit slightly lower drift than clean samples (\emph{false stability}). As noise strength increases, adversarial drift grows more rapidly and eventually exceeds clean drift, yielding a clear high-noise separation signal.}
    \label{fig:app_uniform_noise_tau_anlalysis_eps_all_vit_l_14_datacomp}
    \vspace{-1em}

\end{figure}

\begin{figure}[t]
    \centering
    \includegraphics[width=\linewidth]{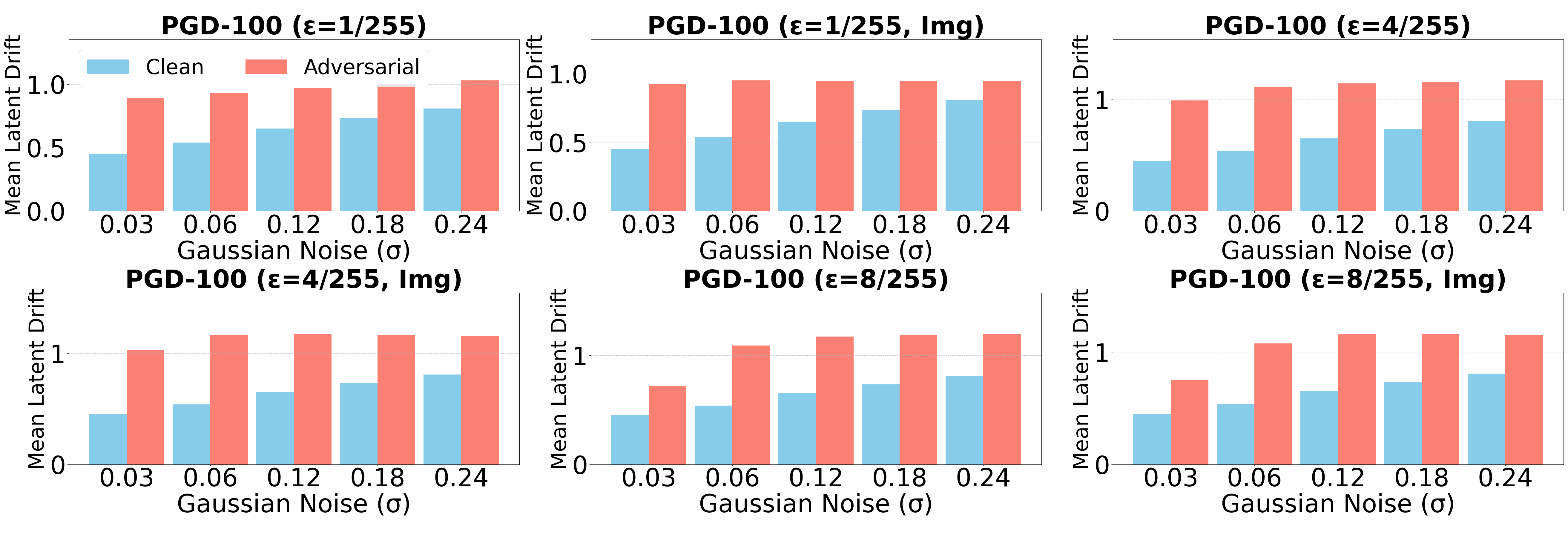}
\caption{\textbf{ViT-L-14(DataComp-1B) at PGD-100 ($\epsilon=\frac{X}{255}$)}. Mean latent drift ($\tau$) versus gaussian noise strength for clean and adversarial samples averaged across eight fine-grained datasets. Adversarial examples are generated at different perturbation budgets and using two attack objectives: the standard objective that maximizes cross-entropy loss, and a vision-only objective (Img) that maximizes the discrepancy in visual features. Under weak noise, adversarial samples exhibit slightly lower drift than clean samples (\emph{false stability}). As noise strength increases, adversarial drift grows more rapidly and eventually exceeds clean drift, yielding a clear high-noise separation signal.}
    \label{fig:app_gaussian_noise_tau_anlalysis_eps_all_vit_l_14_datacomp}
    \vspace{-1em}

\end{figure}

\begin{figure}[t]
    \centering    
    \includegraphics[width=\linewidth,keepaspectratio]{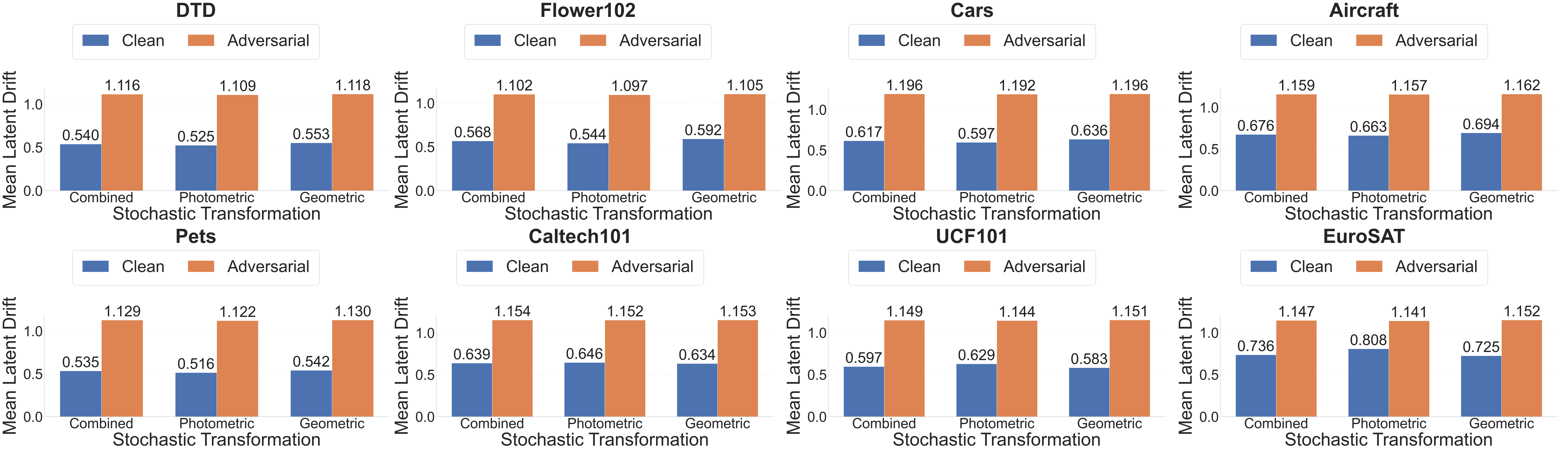}
    \caption{\textbf{ViT-L-14(DataComp-1B) at PGD-100 ($\epsilon=\frac{4}{255}$)}. Mean latent drift ($\tau$) versus stochastic transformations~\cite{abdul2023align,sheng2025r} for clean and adversarial samples across eight fine-grained datasets. Under  different transformations, adversarial drift becomes substantially larger than clean drift, yielding a reliable high separation signal.}
    \label{fig:app_tpt_aug_tau_anlalysis_eps4_vit_l_14_datacomp}
    \vspace{-1em}

\end{figure}

\begin{figure}[t]
    \centering
    \includegraphics[width=\linewidth]{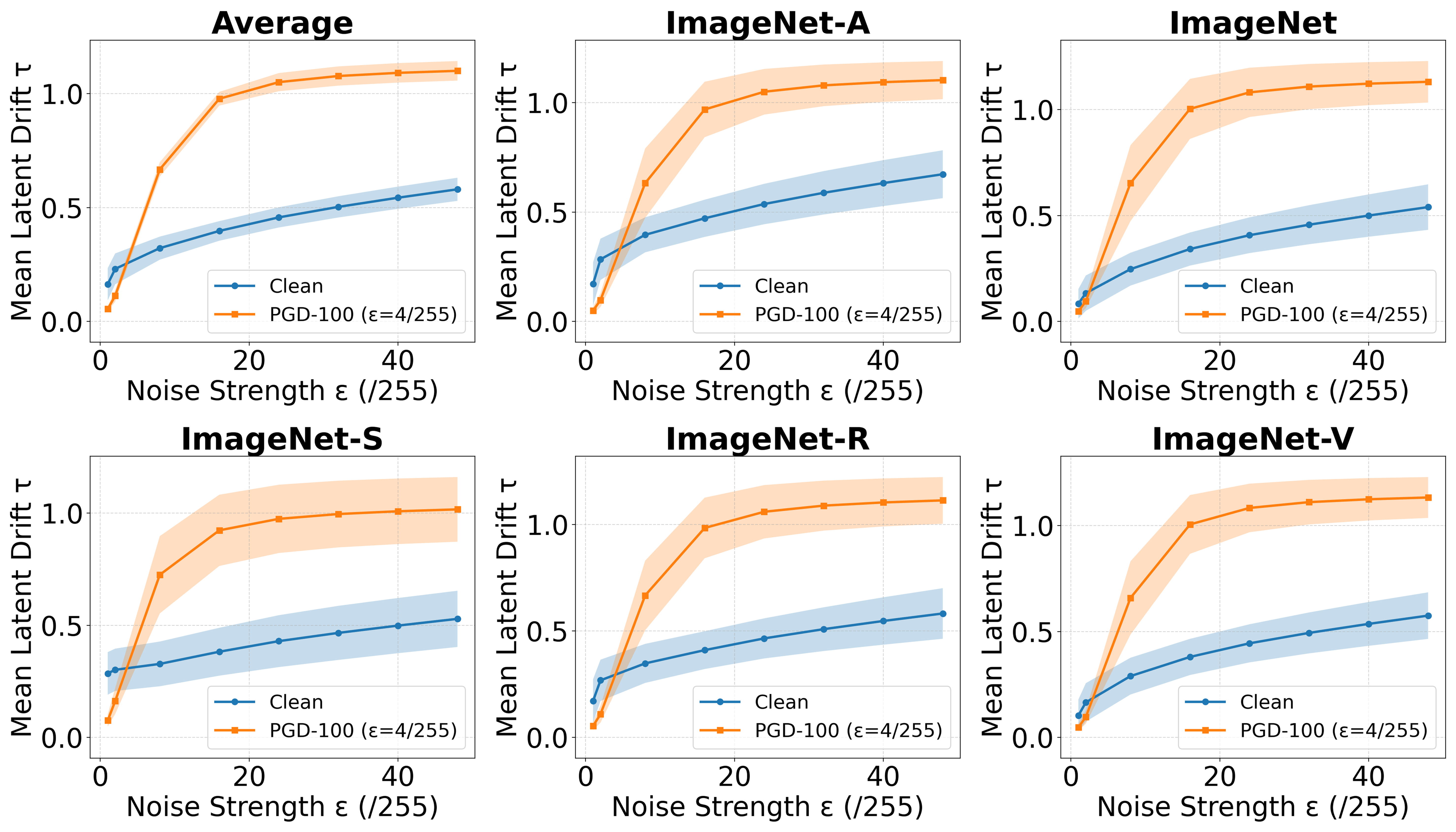}
    \caption{\textbf{ViT-L-14(DataComp-1B) at PGD-100 ($\epsilon=\frac{4}{100}$)}. Mean latent drift ($\tau$) versus uniform noise strength ($\epsilon$) for clean and adversarial samples across ImageNet and its four out of distribution datasets. Under weak noise, adversarial samples exhibit slightly lower drift than clean samples (\emph{false stability}). As noise strength increases, the curves cross and adversarial drift becomes substantially larger than clean drift, yielding a reliable high-noise separation signal.}
    \label{fig:app_uniform_noise_tau_anlalysis_eps4_vit_l_14_datacomp_imagenet}
    \vspace{-1em}

\end{figure}

\begin{figure}[t]
    \centering
    \includegraphics[width=\linewidth]{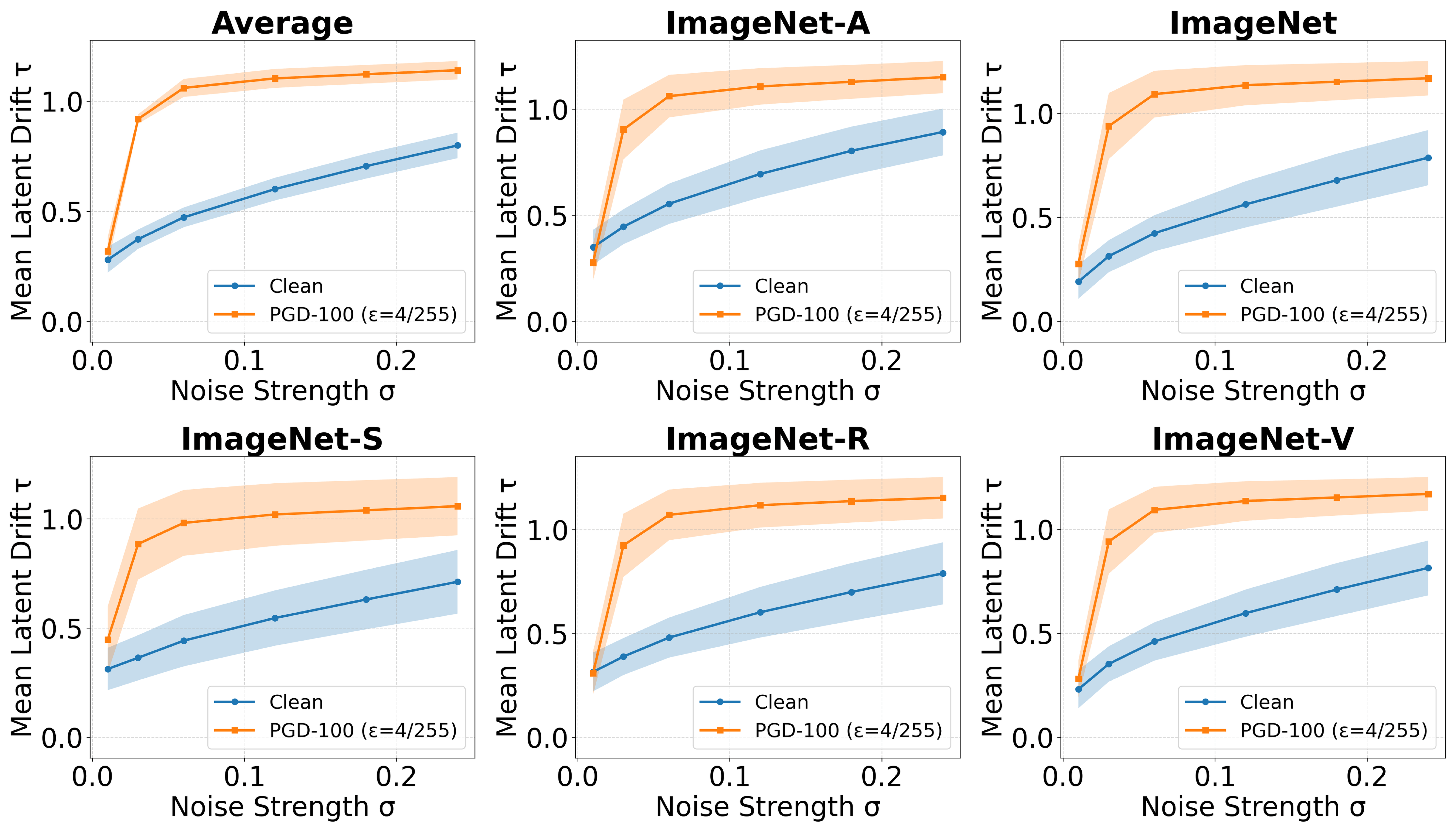}
    \caption{\textbf{ViT-L-14(DataComp-1B) at PGD-100 ($\epsilon=\frac{4}{100}$)}. Mean latent drift ($\tau$) versus gaussian noise strength ($\sigma$) for clean and adversarial samples across ImageNet and its four out of distribution datasets. Under weak noise, adversarial samples exhibit slightly lower drift than clean samples (\emph{false stability}). As noise strength increases, the curves cross and adversarial drift becomes substantially larger than clean drift, yielding a reliable high-noise separation signal.}
    \label{fig:app_gaussian_noise_tau_anlalysis_eps4_vit_l_14_datacomp_imagenet}
    \vspace{-1em}

\end{figure}

We analyze mean latent drift for CLIP 
ViT-L/14 pretrained on 
DataComp-1B~\cite{liu2024clips} under 
different stochastic perturbations and attack 
settings. 
Figs.~\ref{fig:app_uniform_noise_tau_anlalysis_eps4_vit_l_14_datacomp}, 
\ref{fig:app_uniform_noise_tau_anlalysis_eps8_vit_l_14_datacomp}, 
\ref{fig:app_gaussian_noise_tau_anlalysis_eps4_vit_l_14_datacomp}, 
and~\ref{fig:app_gaussian_noise_tau_anlalysis_eps8_vit_l_14_datacomp} 
report drift curves for clean and adversarial 
samples across the eight fine-grained 
datasets under increasing strengths of 
uniform and Gaussian noise injected onto 
input samples, for PGD-100 adversarial 
examples at $\epsilon \in \{4/255, 8/255\}$. 
Across all datasets and noise types, 
adversarial samples exhibit slightly lower 
drift than clean samples under weak noise 
injections (\emph{false stability}), with 
the curves crossing as noise strength 
increases until adversarial drift becomes 
substantially larger---a pattern that holds 
consistently across attack budgets and 
objectives, as confirmed by the 
dataset-averaged summaries in 
Figs.~\ref{fig:app_uniform_noise_tau_anlalysis_eps_all_vit_l_14_datacomp} 
and~\ref{fig:app_gaussian_noise_tau_anlalysis_eps_all_vit_l_14_datacomp}. 
The same drift crossover is observed on 
ImageNet and its four OOD variants 
(Figs.~\ref{fig:app_uniform_noise_tau_anlalysis_eps4_vit_l_14_datacomp_imagenet}--\ref{fig:app_gaussian_noise_tau_anlalysis_eps4_vit_l_14_datacomp_imagenet}), 
indicating the phenomenon is not restricted 
to fine-grained settings. 
Fig.~\ref{fig:app_tpt_aug_tau_anlalysis_eps4_vit_l_14_datacomp} 
further confirms that a similar separation 
emerges under photometric, geometric, and 
combined transformations, demonstrating that 
the high-noise separability signal is not 
tied to a particular perturbation 
distribution.

\subsubsection{ViT-L/14}

\begin{figure}[t]
    \centering
    \includegraphics[width=\linewidth]{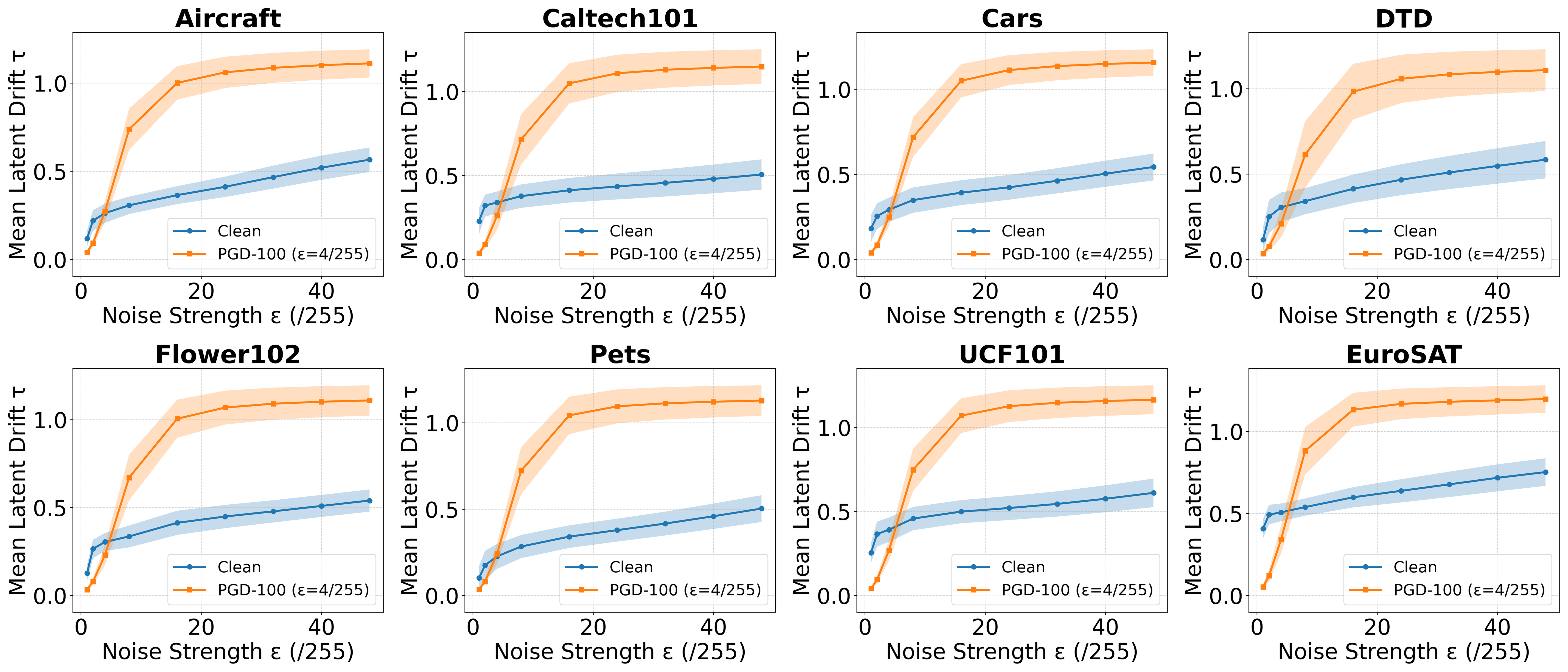}
    \caption{\textbf{ViT-L-14 at PGD-100 ($\epsilon=\frac{4}{100}$)}. Mean latent drift ($\tau$) versus uniform noise strength ($\sigma$) for clean and adversarial samples across eight fine-grained datasets. Under weak noise, adversarial samples exhibit slightly lower drift than clean samples (\emph{false stability}). As noise strength increases, the curves cross and adversarial drift becomes substantially larger than clean drift, yielding a reliable high-noise separation signal.}
    \label{fig:app_uniform_noise_tau_anlalysis_eps4_vit_l_14}
    \vspace{-1em}

\end{figure}

\begin{figure}[t]
    \centering
    \includegraphics[width=\linewidth]{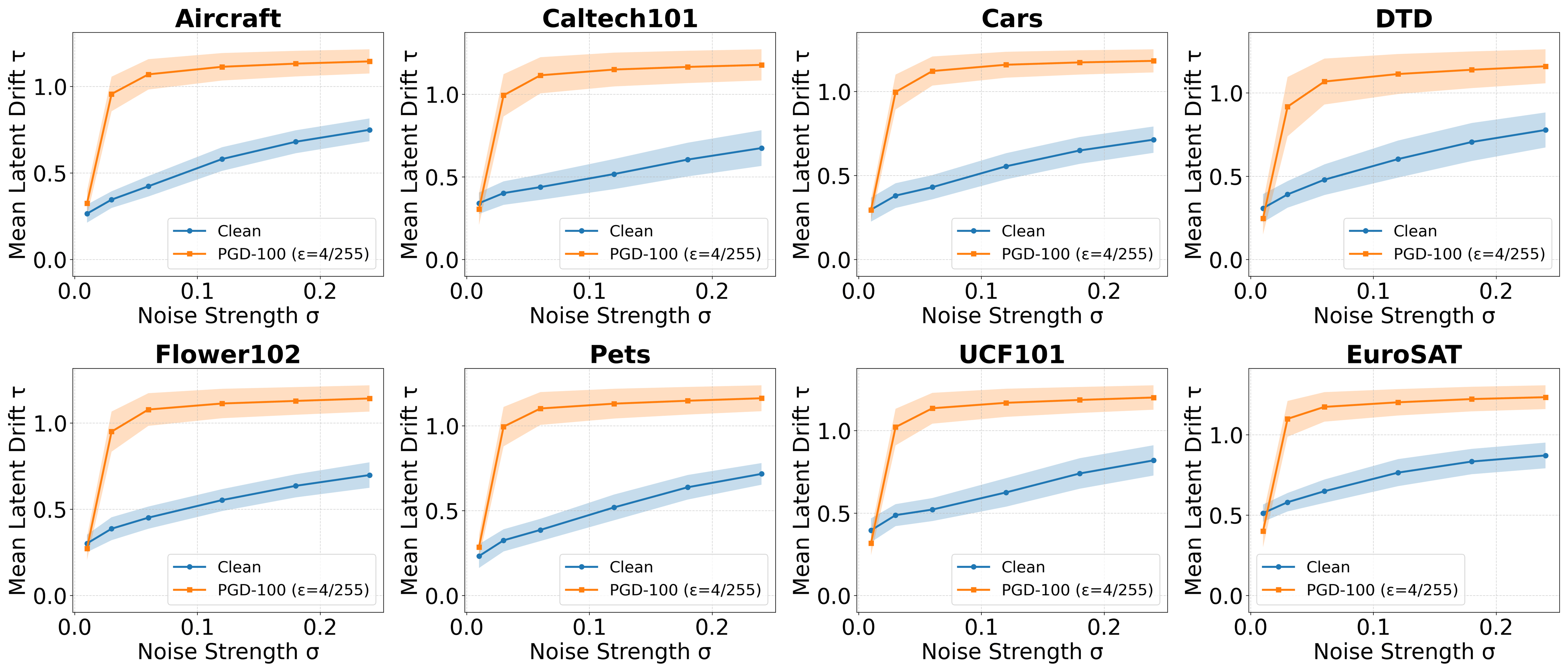}
    \caption{\textbf{ViT-L-14 at PGD-100 ($\epsilon=\frac{4}{100}$)}. Mean latent drift ($\tau$) versus gaussian noise strength ($\sigma$) for clean and adversarial samples across eight fine-grained datasets. Under weak noise, adversarial samples exhibit slightly lower drift than clean samples (\emph{false stability}). As noise strength increases, the curves cross and adversarial drift becomes substantially larger than clean drift, yielding a reliable high-noise separation signal.}
    \label{fig:app_gaussian_noise_tau_anlalysis_eps4_vit_l_14}
    \vspace{-1em}

\end{figure}

\begin{figure}[t]
    \centering
    \includegraphics[width=\linewidth]{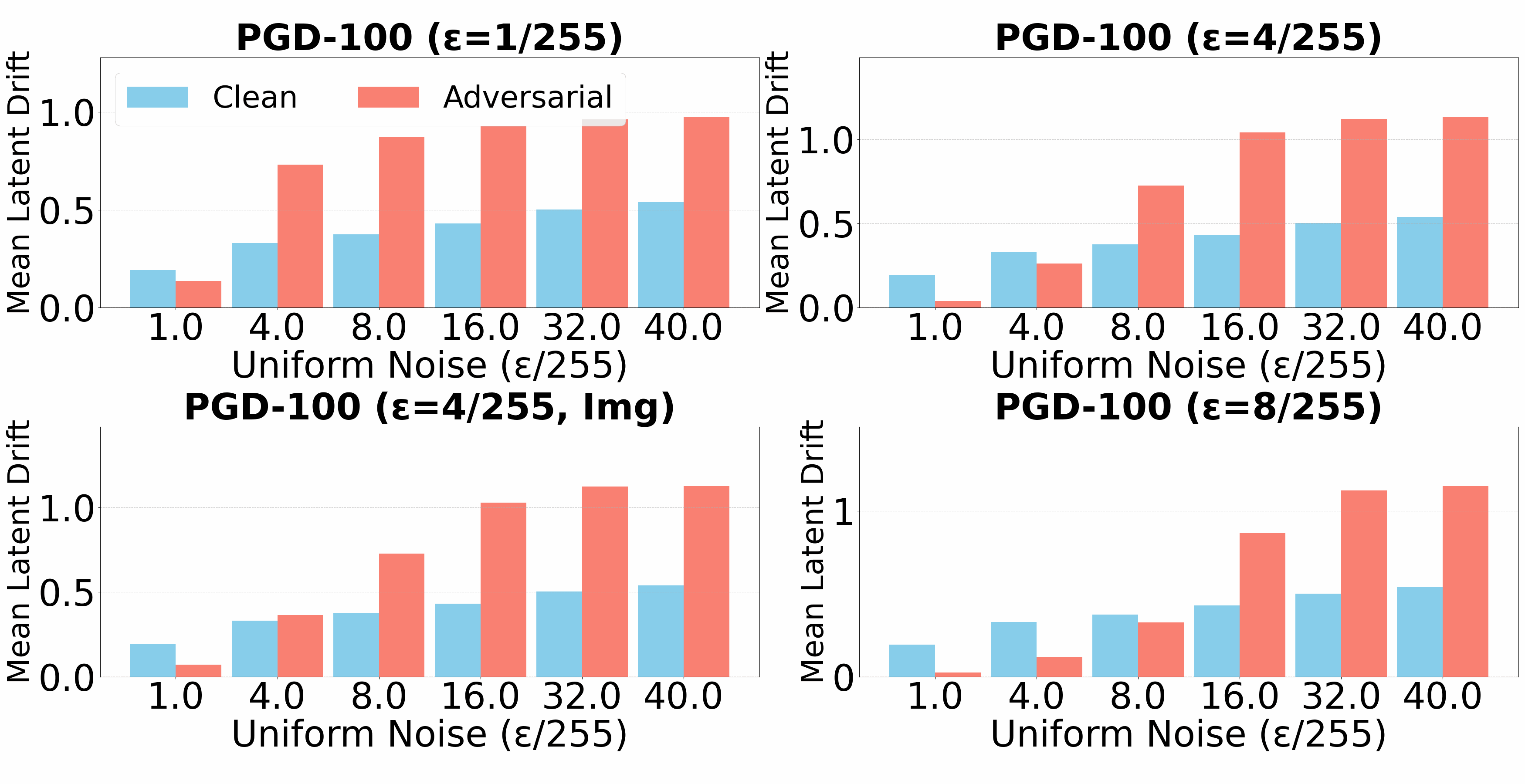}
\caption{\textbf{ViT-L-14 at PGD-100 ($\epsilon=\frac{X}{255}$)}. Mean latent drift ($\tau$) versus uniform noise strength for clean and adversarial samples averaged across eight fine-grained datasets. Adversarial examples are generated at different perturbation budgets and using two attack objectives: the standard objective that maximizes cross-entropy loss, and a vision-only objective (Img) that maximizes the discrepancy in visual features. Under weak noise, adversarial samples exhibit slightly lower drift than clean samples (\emph{false stability}). As noise strength increases, adversarial drift grows more rapidly and eventually exceeds clean drift, yielding a clear high-noise separation signal.}
    \label{fig:app_uniform_noise_tau_anlalysis_eps_all_vit_l_14}
    \vspace{-1em}

\end{figure}

\begin{figure}[t]
    \centering
    \includegraphics[width=\linewidth]{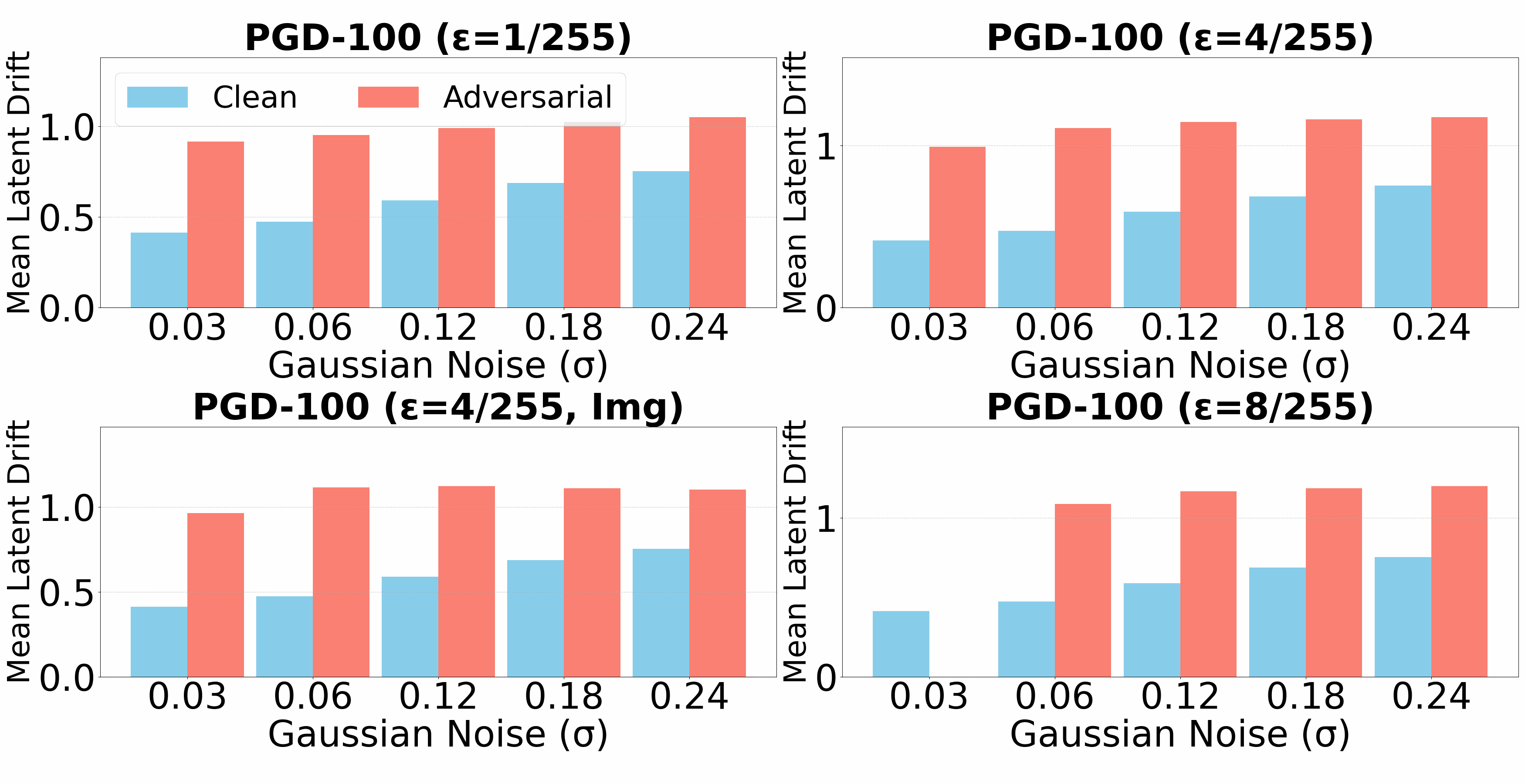}
\caption{\textbf{ViT-L-14 at PGD-100 ($\epsilon=\frac{X}{255}$)}. Mean latent drift ($\tau$) versus gaussian noise strength for clean and adversarial samples averaged across eight fine-grained datasets. Adversarial examples are generated at different perturbation budgets and using two attack objectives: the standard objective that maximizes cross-entropy loss, and a vision-only objective (Img) that maximizes the discrepancy in visual features. Under weak noise, adversarial samples exhibit slightly lower drift than clean samples (\emph{false stability}). As noise strength increases, adversarial drift grows more rapidly and eventually exceeds clean drift, yielding a clear high-noise separation signal.}
    \label{fig:app_gaussian_noise_tau_anlalysis_eps_all_vit_l_14}
    \vspace{-1em}

\end{figure}

Figs.~\ref{fig:app_uniform_noise_tau_anlalysis_eps4_vit_l_14} 
and~\ref{fig:app_gaussian_noise_tau_anlalysis_eps4_vit_l_14} 
report mean latent drift curves for clean 
and adversarial samples across the eight 
fine-grained datasets for the original CLIP 
ViT-L/14~\cite{radford2021learning}, with 
adversarial examples generated using PGD-100 
at $\epsilon=4/255$ and drift computed under 
increasing uniform and Gaussian noise 
injections. The same qualitative behavior as 
the DataComp-pretrained model is observed: 
adversarial samples exhibit false stability 
under weak noise, with the curves crossing 
as noise strength increases to produce a 
clear high-noise separation signal. 
Figs.~\ref{fig:app_uniform_noise_tau_anlalysis_eps_all_vit_l_14} 
and~\ref{fig:app_gaussian_noise_tau_anlalysis_eps_all_vit_l_14} 
confirm the same regime transition across 
attack budgets and objectives.

\subsubsection{FARE}

\begin{figure}[t]
    \centering
    \includegraphics[width=\linewidth]{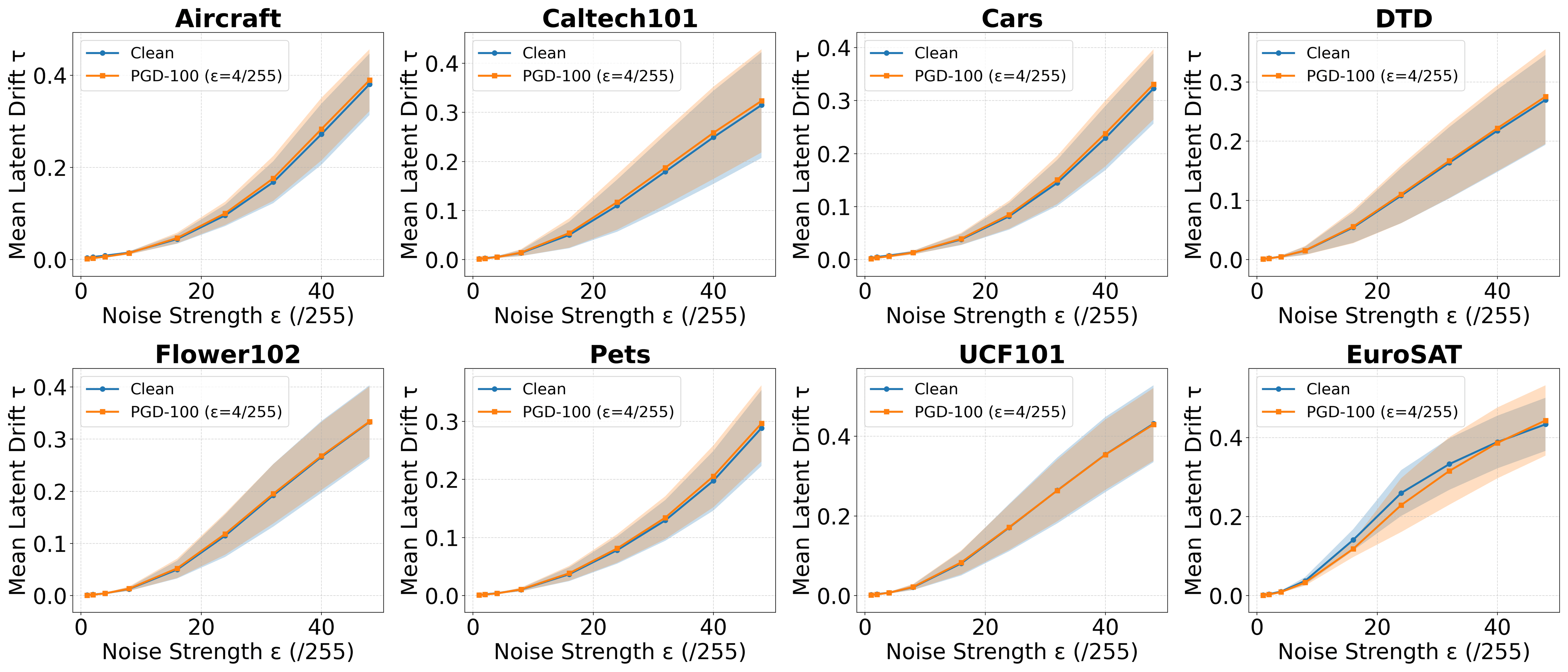}
    \caption{\textbf{FARE~\cite{schlarmann2024robust} at PGD-100 ($\epsilon=\frac{4}{100}$)}. Mean latent drift ($\tau$) versus uniform noise strength ($\sigma$) for clean and adversarial samples across eight fine-grained datasets. In contrast to standard CLIP models, clean and adversarial samples exhibit closely aligned drift curves across perturbation strengths, with little pronounced separation. This behavior is consistent with adversarial training, which explicitly encourages alignment between clean and adversarial feature distributions, thereby reducing the distinct high-noise drift signature observed in non-robust models.}
    \label{fig:app_uniform_noise_tau_anlalysis_eps4_fare}
    \vspace{-1em}

\end{figure}

\begin{figure}[t]
    \centering
    \includegraphics[width=\linewidth]{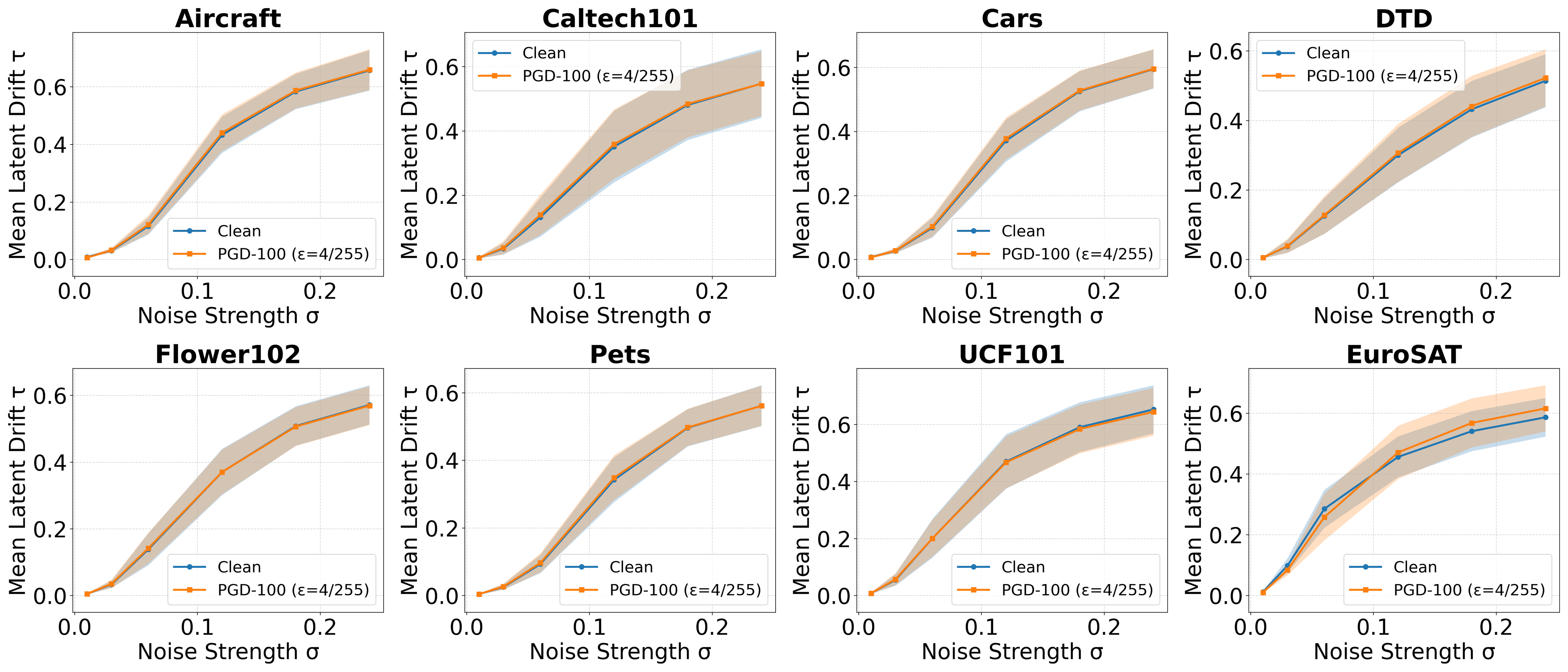}
    \caption{\textbf{FARE~\cite{schlarmann2024robust} at PGD-100 ($\epsilon=\frac{4}{100}$)}. Mean latent drift ($\tau$) versus gaussian noise strength ($\sigma$) for clean and adversarial samples across eight fine-grained datasets.In contrast to standard CLIP models, clean and adversarial samples exhibit closely aligned drift curves across perturbation strengths, with little pronounced separation. This behavior is consistent with adversarial training, which explicitly encourages alignment between clean and adversarial feature distributions, thereby reducing the distinct high-noise drift signature observed in non-robust models.}
    \label{fig:app_gaussian_noise_tau_anlalysis_eps4_fare}
    \vspace{-1em}

\end{figure}

\begin{figure}[t]
    \centering
    \includegraphics[width=\linewidth]{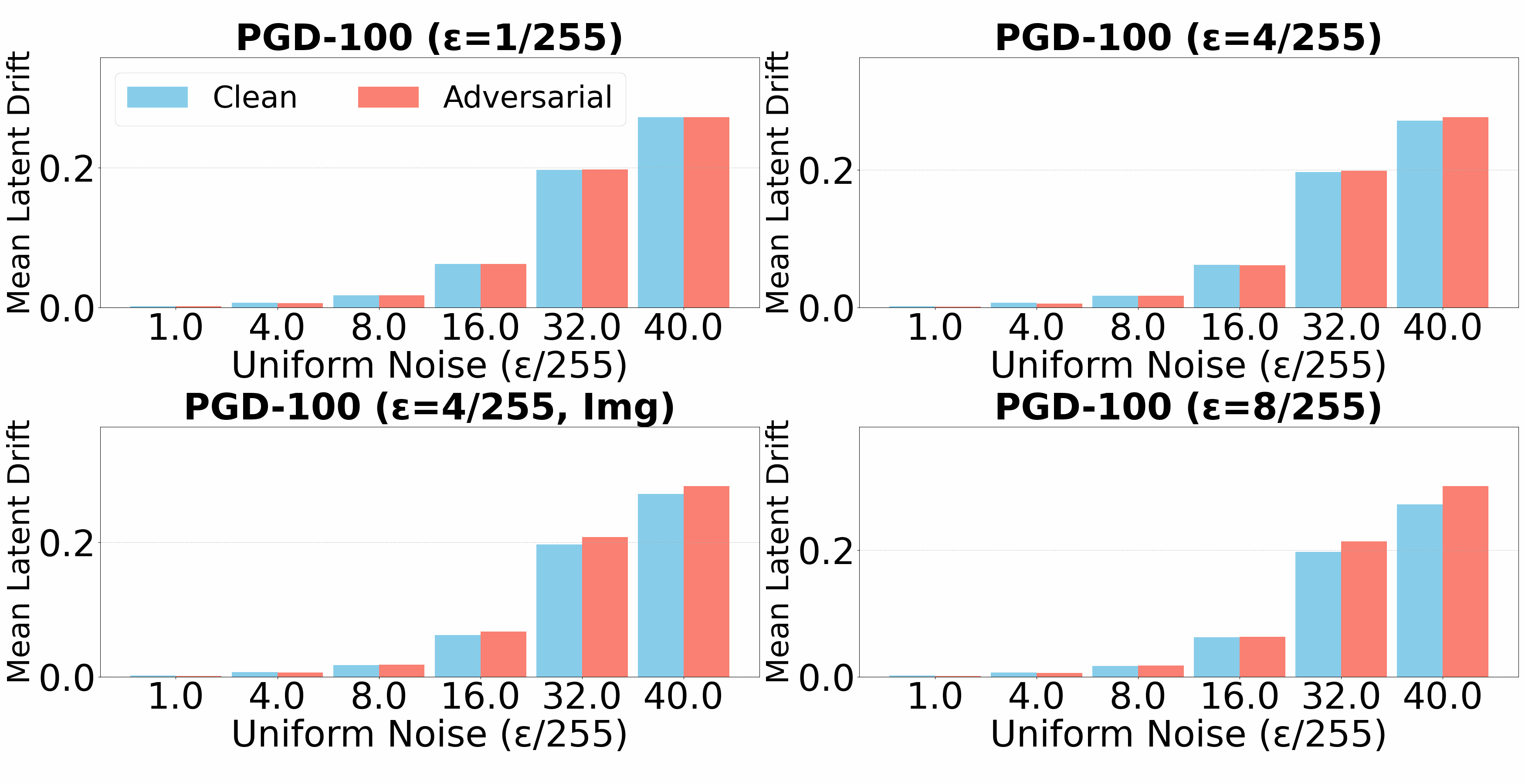}
\caption{\textbf{FARE~\cite{schlarmann2024robust} at PGD-100 ($\epsilon=\frac{X}{255}$)}. Mean latent drift ($\tau$) versus uniform noise strength for clean and adversarial samples averaged across eight fine-grained datasets. Adversarial examples are generated at different perturbation budgets and using two attack objectives: the standard objective that maximizes cross-entropy loss, and a vision-only objective (Img) that maximizes the discrepancy in visual features. In contrast to standard CLIP models, clean and adversarial samples exhibit closely aligned drift curves across perturbation strengths, with little pronounced separation. This behavior is consistent with adversarial training, which explicitly encourages alignment between clean and adversarial feature distributions, thereby reducing the distinct high-noise drift signature observed in non-robust models.}
    \label{fig:app_uniform_noise_tau_anlalysis_eps_all_fare}
    \vspace{-1em}

\end{figure}

\begin{figure}[t]
    \centering
    \includegraphics[width=\linewidth]{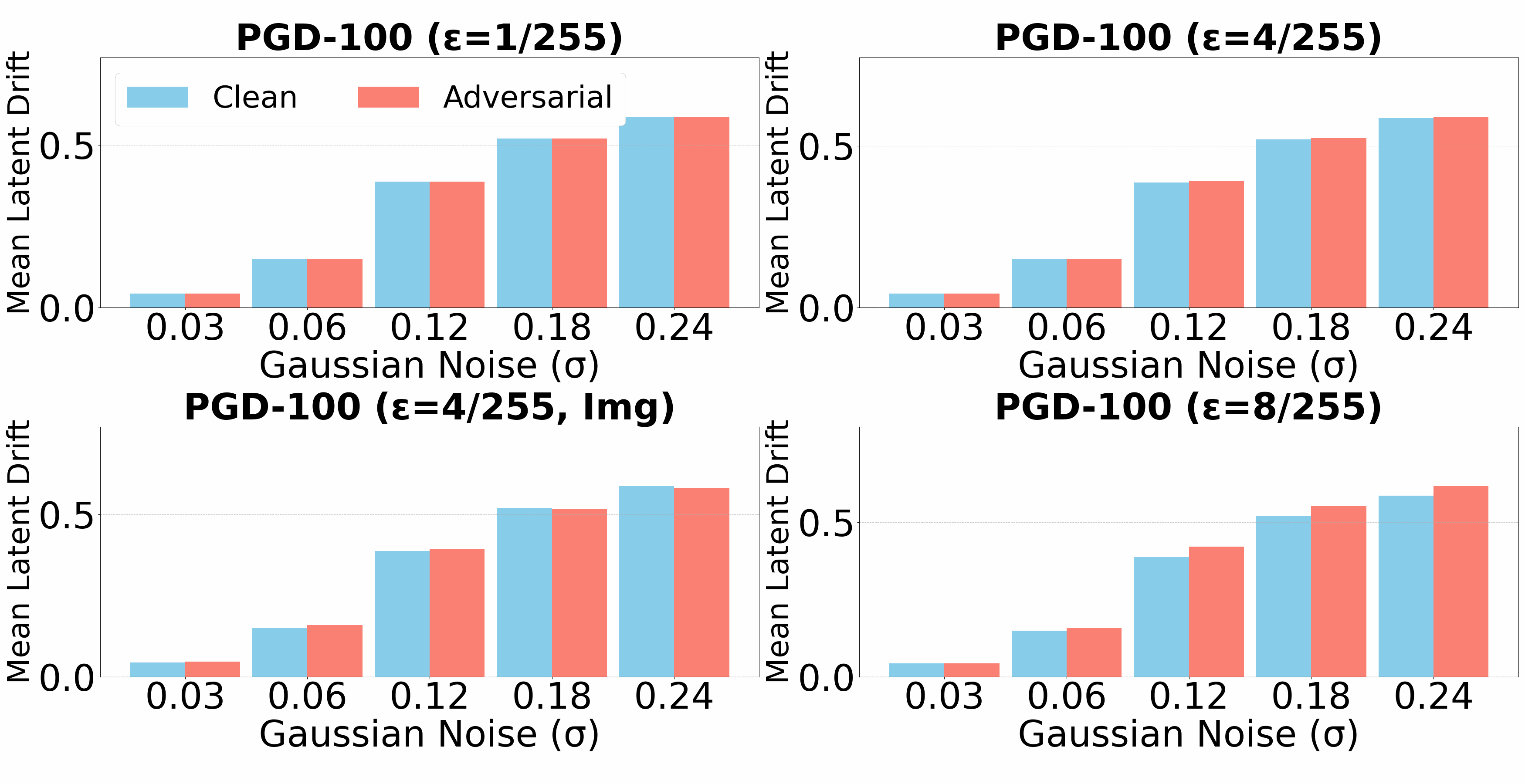}
\caption{\textbf{FARE~\cite{schlarmann2024robust} at PGD-100 ($\epsilon=\frac{X}{255}$)}. Mean latent drift ($\tau$) versus gaussian noise strength for clean and adversarial samples averaged across eight fine-grained datasets. Adversarial examples are generated at different perturbation budgets and using two attack objectives: the standard objective that maximizes cross-entropy loss, and a vision-only objective (Img) that maximizes the discrepancy in visual features. In contrast to standard CLIP models, clean and adversarial samples exhibit closely aligned drift curves across perturbation strengths, with little pronounced separation. This behavior is consistent with adversarial training, which explicitly encourages alignment between clean and adversarial feature distributions, thereby reducing the distinct high-noise drift signature observed in non-robust models.}
    \label{fig:app_gaussian_noise_tau_anlalysis_eps_all_fare}
    \vspace{-1em}

\end{figure}

Figs.~\ref{fig:app_uniform_noise_tau_anlalysis_eps4_fare} 
and~\ref{fig:app_gaussian_noise_tau_anlalysis_eps4_fare} 
report mean latent drift curves across the 
eight fine-grained datasets for 
FARE~\cite{schlarmann2024robust}, with 
adversarial examples generated using PGD-100 
at $\epsilon=4/255$ and drift computed under 
increasing uniform and Gaussian noise 
injections. In contrast to non-robust CLIP 
models, clean and adversarial samples exhibit 
closely aligned drift trajectories across all 
datasets---the characteristic crossover is 
largely absent, with both conditions 
responding to noise injections in a similar 
manner. 
Figs.~\ref{fig:app_uniform_noise_tau_anlalysis_eps_all_fare} 
and~\ref{fig:app_gaussian_noise_tau_anlalysis_eps_all_fare} 
confirm this alignment across attack budgets 
and objectives, consistent with adversarial 
training explicitly encouraging clean and 
adversarial representations to occupy similar 
regions of feature space.

\subsubsection{DeltaCLIP-L}

\begin{figure}[t]
    \centering
    \includegraphics[width=\linewidth]{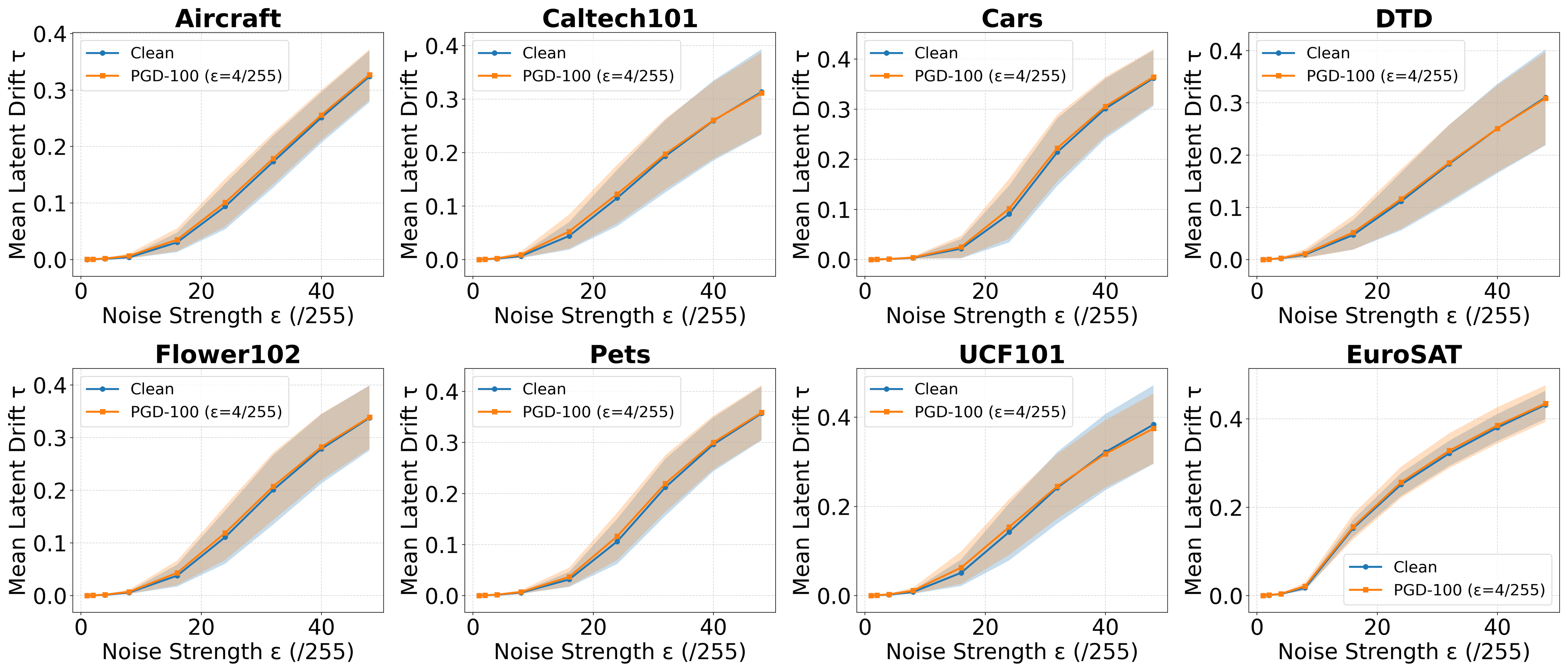}
    \caption{\textbf{DeltaCLIP-L~\cite{wang2025double} at PGD-100 ($\epsilon=\frac{4}{100}$)}. Mean latent drift ($\tau$) versus uniform noise strength ($\sigma$) for clean and adversarial samples across eight fine-grained datasets. In contrast to standard CLIP models, clean and adversarial samples exhibit closely aligned drift curves across perturbation strengths, with little pronounced separation. This behavior is consistent with adversarial training, which explicitly encourages alignment between clean and adversarial feature distributions, thereby reducing the distinct high-noise drift signature observed in non-robust models.}
    \label{fig:app_uniform_noise_tau_anlalysis_eps4_deltaclip}
    \vspace{-1em}

\end{figure}

\begin{figure}[t]
    \centering
    \includegraphics[width=\linewidth]{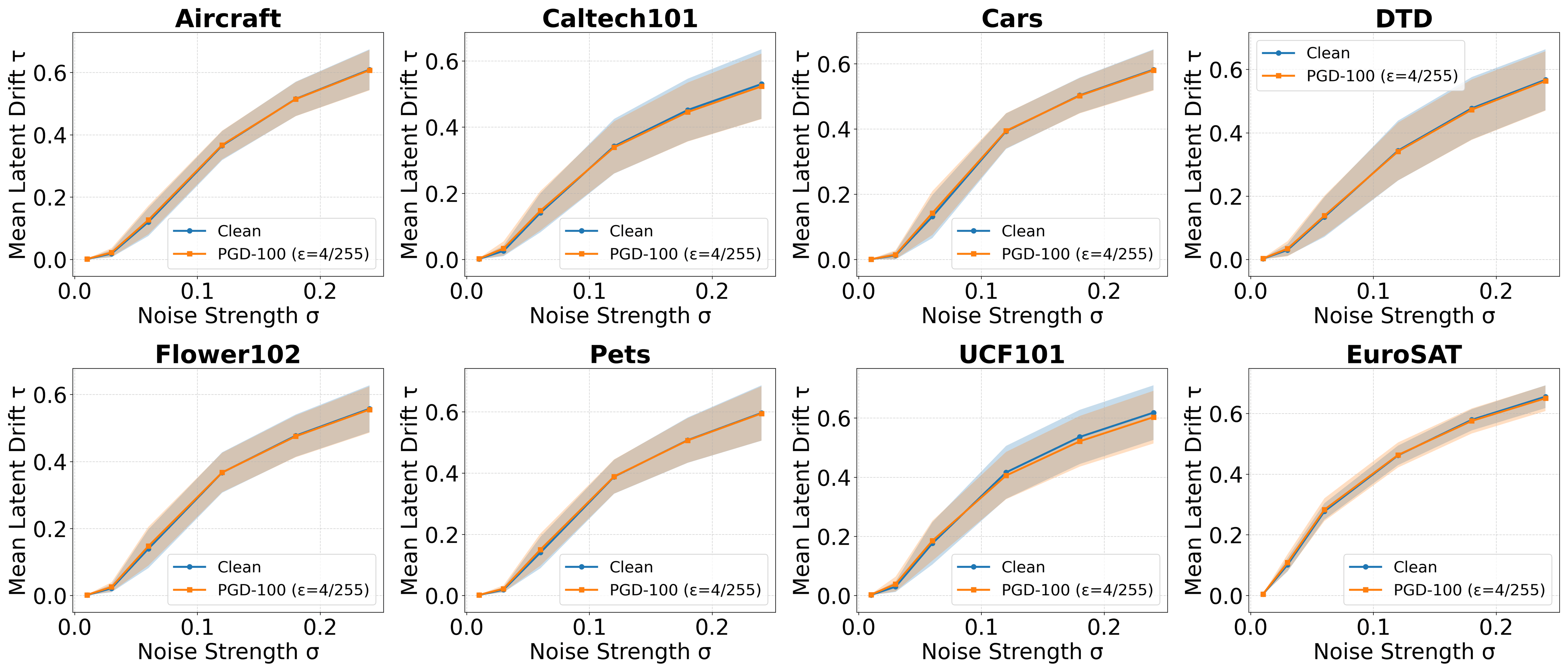}
    \caption{\textbf{DeltaCLIP-L~\cite{wang2025double} at PGD-100 ($\epsilon=\frac{4}{100}$)}. Mean latent drift ($\tau$) versus gaussian noise strength ($\sigma$) for clean and adversarial samples across eight fine-grained datasets. In contrast to standard CLIP models, clean and adversarial samples exhibit closely aligned drift curves across perturbation strengths, with little pronounced separation. This behavior is consistent with adversarial training, which explicitly encourages alignment between clean and adversarial feature distributions, thereby reducing the distinct high-noise drift signature observed in non-robust models.}
    \label{fig:app_gaussian_noise_tau_anlalysis_eps4_deltaclip}
    \vspace{-1em}

\end{figure}

\begin{figure}[t]
    \centering
    \includegraphics[width=\linewidth]{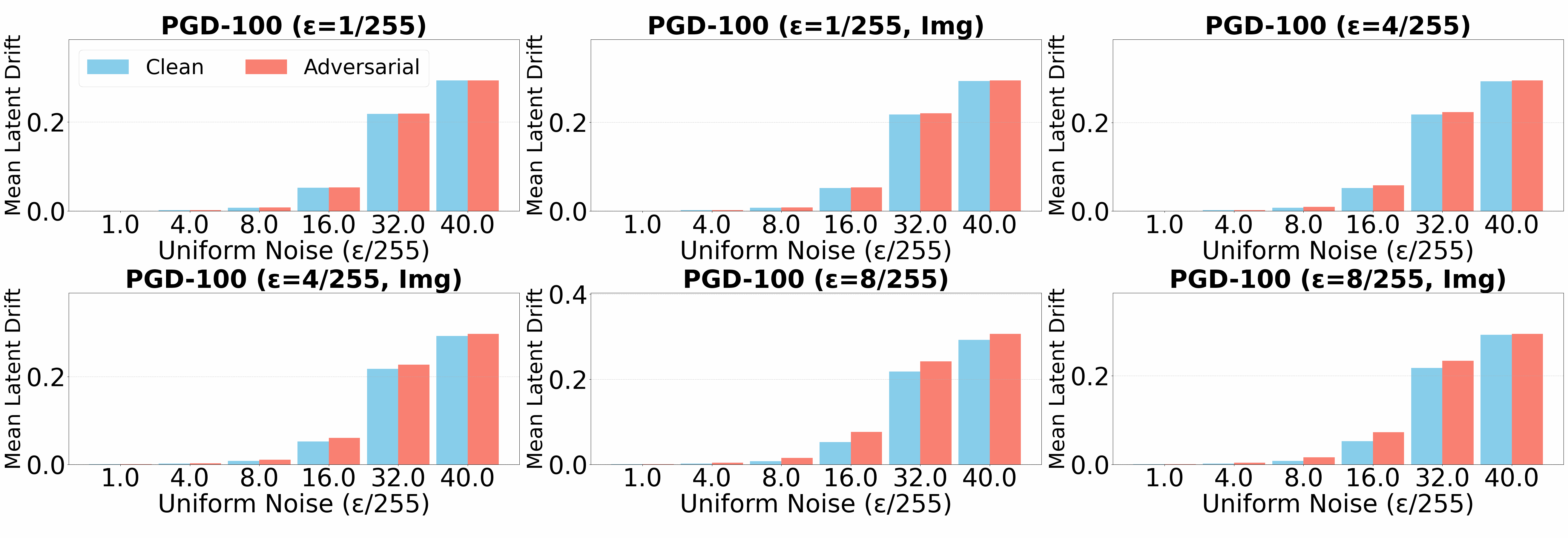}
\caption{\textbf{DeltaCLIP-L~\cite{wang2025double} at PGD-100 ($\epsilon=\frac{X}{255}$)}. Mean latent drift ($\tau$) versus uniform noise strength for clean and adversarial samples averaged across eight fine-grained datasets. Adversarial examples are generated at different perturbation budgets and using two attack objectives: the standard objective that maximizes cross-entropy loss, and a vision-only objective (Img) that maximizes the discrepancy in visual features. In contrast to standard CLIP models, clean and adversarial samples exhibit closely aligned drift curves across perturbation strengths, with little pronounced separation. This behavior is consistent with adversarial training, which explicitly encourages alignment between clean and adversarial feature distributions, thereby reducing the distinct high-noise drift signature observed in non-robust models.}
    \label{fig:app_uniform_noise_tau_anlalysis_eps_all_deltaclip}
    \vspace{-1em}

\end{figure}

\begin{figure}[t]
    \centering
    \includegraphics[width=\linewidth]{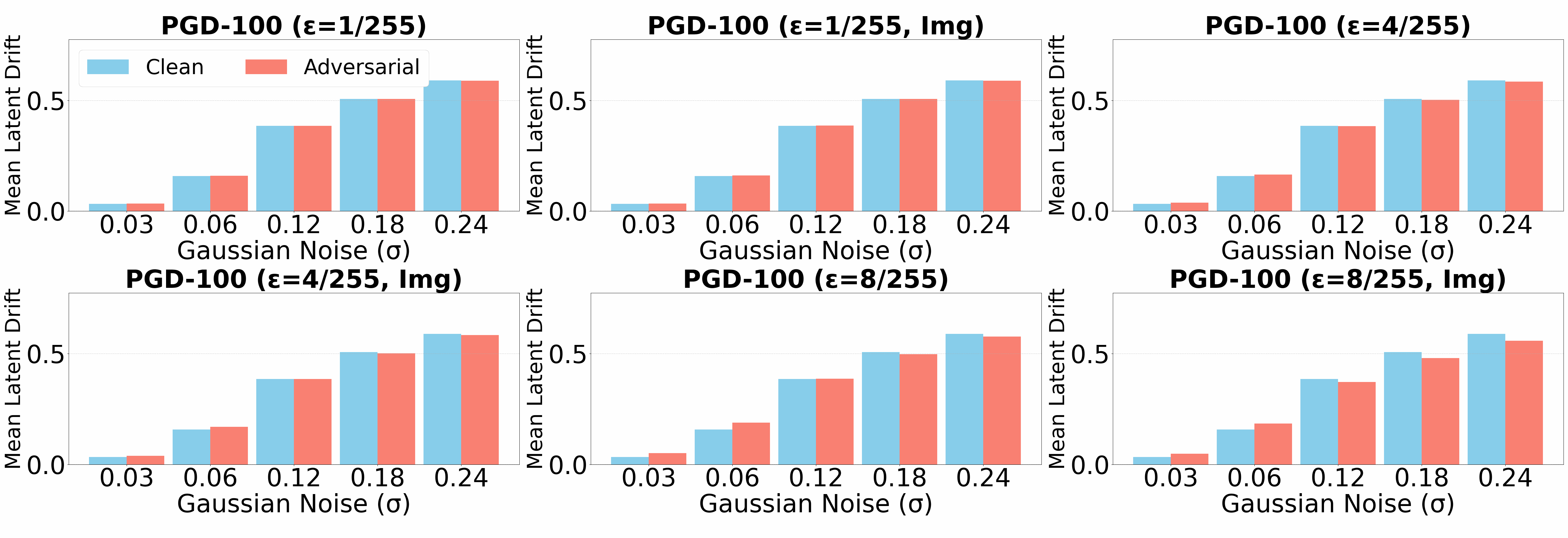}
\caption{\textbf{DeltaCLIP-L~\cite{wang2025double} at PGD-100 ($\epsilon=\frac{X}{255}$)}. Mean latent drift ($\tau$) versus gaussian noise strength for clean and adversarial samples averaged across eight fine-grained datasets. Adversarial examples are generated at different perturbation budgets and using two attack objectives: the standard objective that maximizes cross-entropy loss, and a vision-only objective (Img) that maximizes the discrepancy in visual features. In contrast to standard CLIP models, clean and adversarial samples exhibit closely aligned drift curves across perturbation strengths, with little pronounced separation. This behavior is consistent with adversarial training, which explicitly encourages alignment between clean and adversarial feature distributions, thereby reducing the distinct high-noise drift signature observed in non-robust models.}
    \label{fig:app_gaussian_noise_tau_anlalysis_eps_all_deltaclip}
    \vspace{-1em}

\end{figure}

Figs.~\ref{fig:app_uniform_noise_tau_anlalysis_eps4_deltaclip} 
and~\ref{fig:app_gaussian_noise_tau_anlalysis_eps4_deltaclip} 
report mean latent drift curves across the 
eight fine-grained datasets for 
DeltaCLIP-L~\cite{wang2025double}, with 
adversarial examples generated using PGD-100 
at $\epsilon=4/255$ and drift computed under 
increasing uniform and Gaussian noise 
injections. As with FARE, clean and 
adversarial samples follow closely aligned 
drift trajectories across all datasets, 
with no meaningful crossover at higher 
noise strengths. 
Figs.~\ref{fig:app_uniform_noise_tau_anlalysis_eps_all_deltaclip} 
and~\ref{fig:app_gaussian_noise_tau_anlalysis_eps_all_deltaclip} 
confirm this alignment across attack budgets 
and objectives. Together with the FARE 
results, these findings indicate that 
adversarial training eliminates the 
local-basin structure responsible for the 
high-noise drift separation observed in 
non-robust CLIP models.

\subsection{Evaluating Performance with 
Random Stochastic Transformations}
\label{sec:app_eval_noise}

We next evaluate the direct impact of 
injecting stochastic perturbations onto 
input samples on clean and adversarial 
classification accuracy, complementing the 
representation-level analysis of the 
previous section with a performance-level 
view across different CLIP model variants.

\subsubsection{ViT-L/14 (DataComp-1B)}

\begin{figure}[t]
    \centering
    \includegraphics[width=\linewidth]{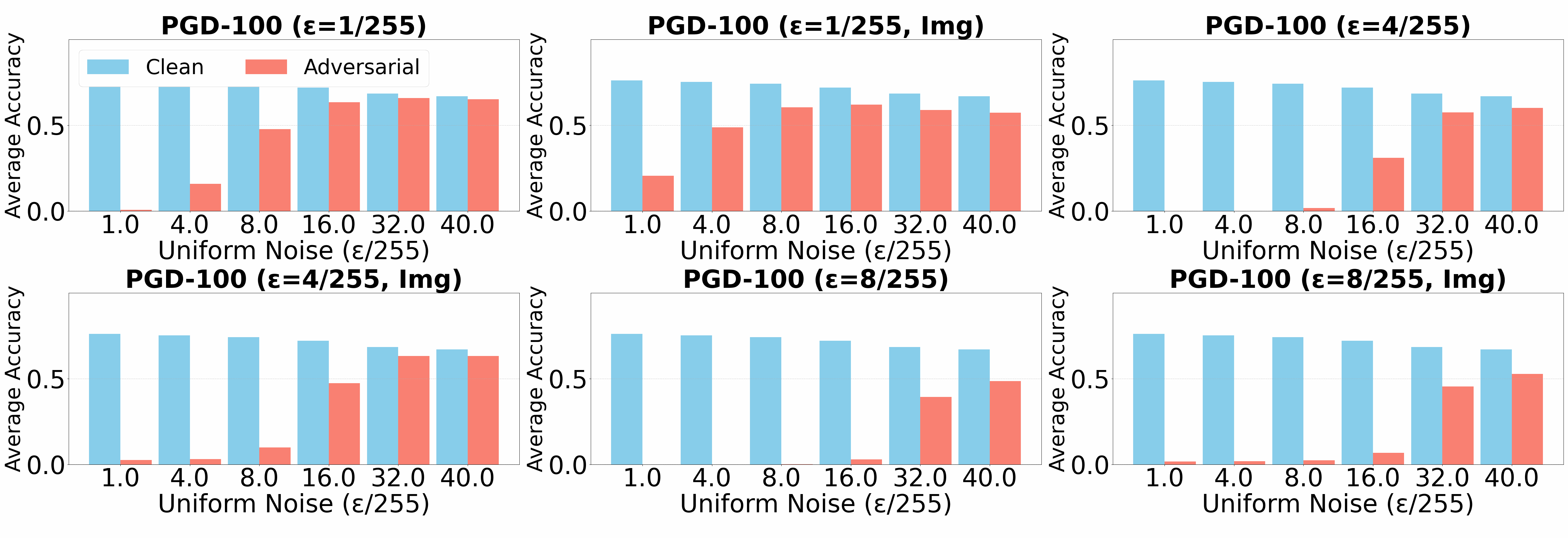}
\caption{\textbf{ViT-L-14(DataComp-1B).} Clean and adversarial accuracy versus uniform noise strength, averaged across eight fine-grained datasets. Adversarial examples (PGD-100) are generated at different perturbation budgets and with two attack objectives: maximizing cross-entropy loss and maximizing the discrepancy in visual features (Img). Across all settings, stronger noise leads to clear improvements in adversarial accuracy, while weak noise yields little robustness benefit and is insufficient to suppress adversarial perturbations.}
    \label{fig:app_uniform_noise_accuracy_anlalysis_eps_all_vit_l_14_datacomp}
    \vspace{-1em}

\end{figure}

\begin{figure}[t]
    \centering
    \includegraphics[width=\linewidth]{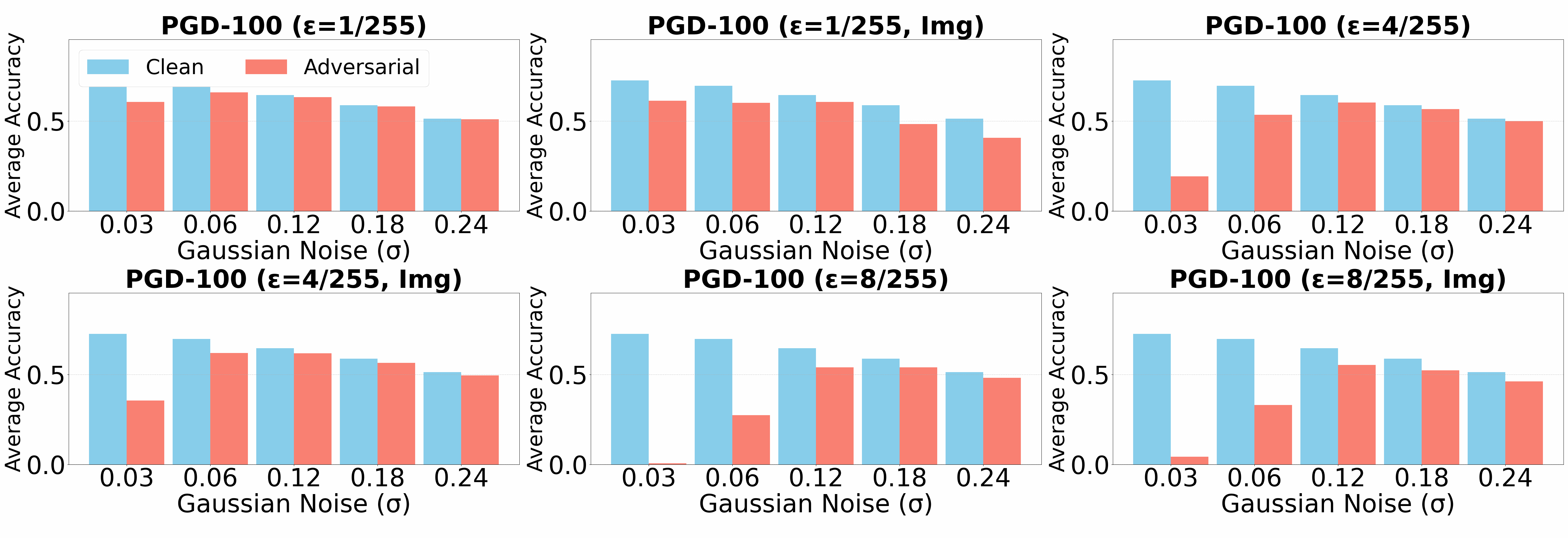}
\caption{\textbf{ViT-L-14(DataComp-1B).} Clean and adversarial accuracy versus Gaussian noise strength, averaged across eight fine-grained datasets. Adversarial examples (PGD-100) are generated at different perturbation budgets and with two attack objectives: maximizing cross-entropy loss and maximizing the discrepancy in visual features (Img). Across settings, moderate Gaussian noise improves adversarial accuracy, whereas stronger noise causes both clean and adversarial performance to drop and gradually converge, reflecting the loss of useful image semantics at high noise levels.}
    \label{fig:app_gaussian_noise_accuracy_anlalysis_eps_all_vit_l_14_datacomp}
    \vspace{-1em}

\end{figure}

Figs.~\ref{fig:app_uniform_noise_accuracy_anlalysis_eps_all_vit_l_14_datacomp} 
and~\ref{fig:app_gaussian_noise_accuracy_anlalysis_eps_all_vit_l_14_datacomp} 
report clean and adversarial accuracy as 
uniform and Gaussian noise injection 
strength increases, averaged across the 
eight fine-grained datasets under multiple 
attack budgets and objectives. Weak noise 
injections provide little robustness benefit; 
as strength increases, adversarial accuracy 
improves substantially while clean accuracy 
degrades only gradually, reflecting the 
asymmetric destabilization of adversarial 
representations discussed in the main paper.

\begin{figure}[t]
    \small \centering
    \begin{minipage}{0.77\textwidth}
        \centering
        \includegraphics[width=\linewidth]{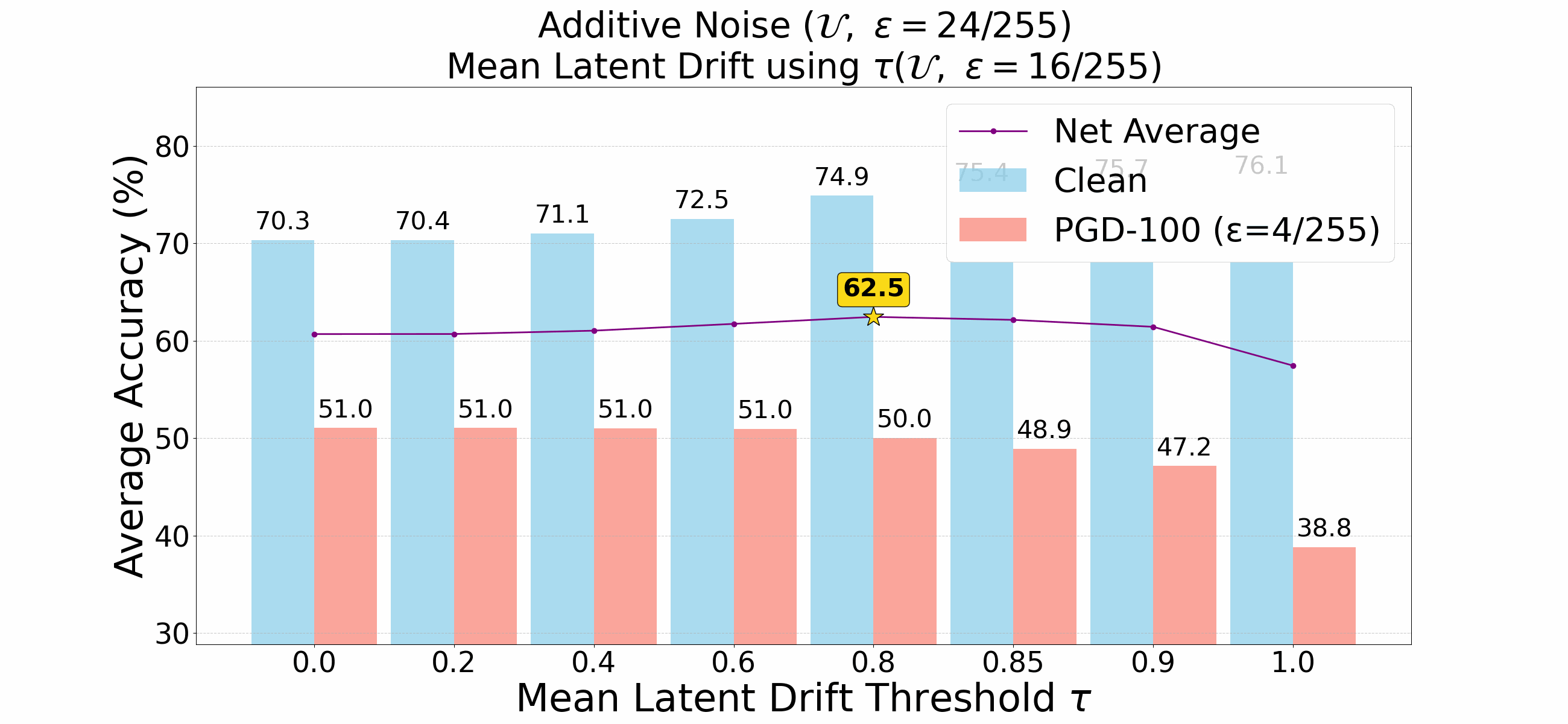}

        \vspace{0.5em}

        \includegraphics[width=\linewidth]{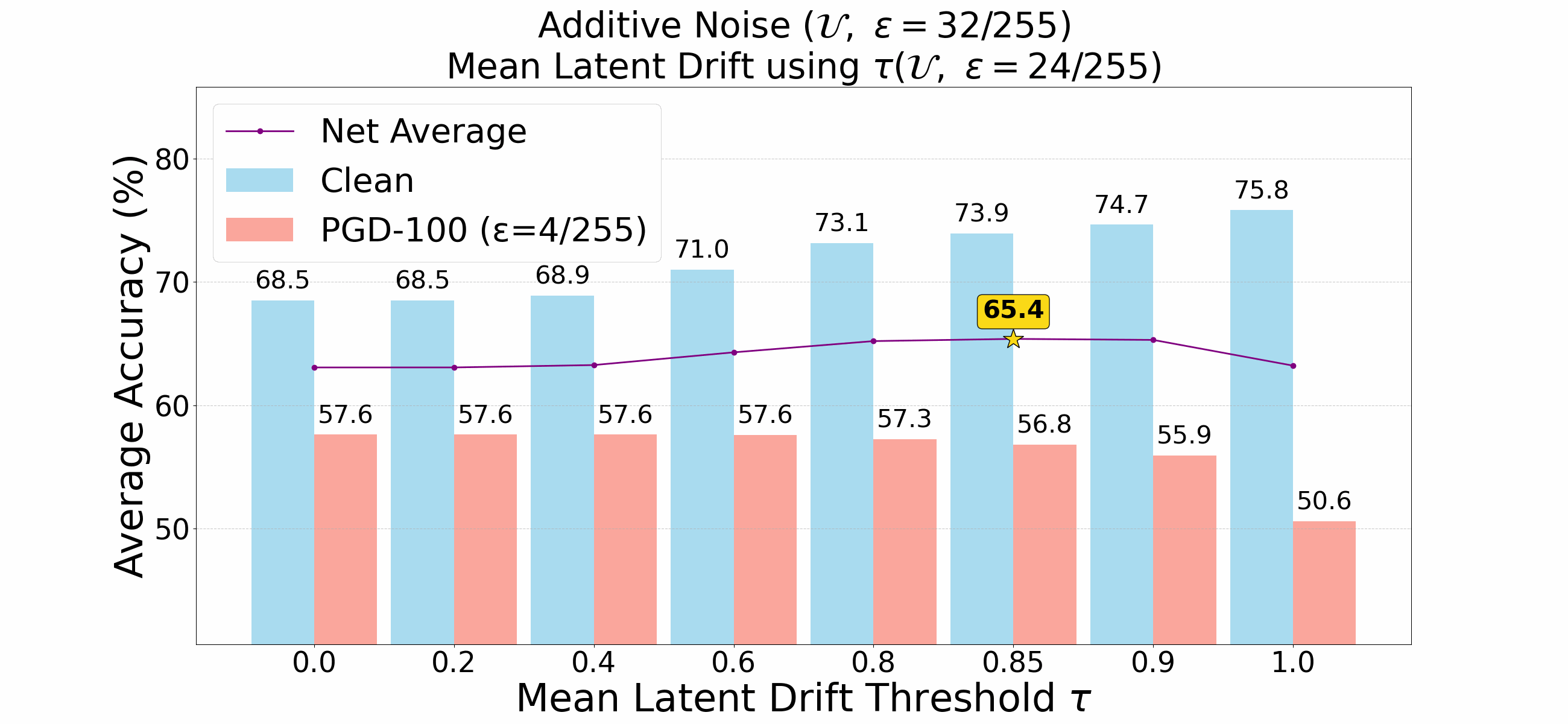}

                \vspace{0.5em}

        \includegraphics[width=\linewidth]{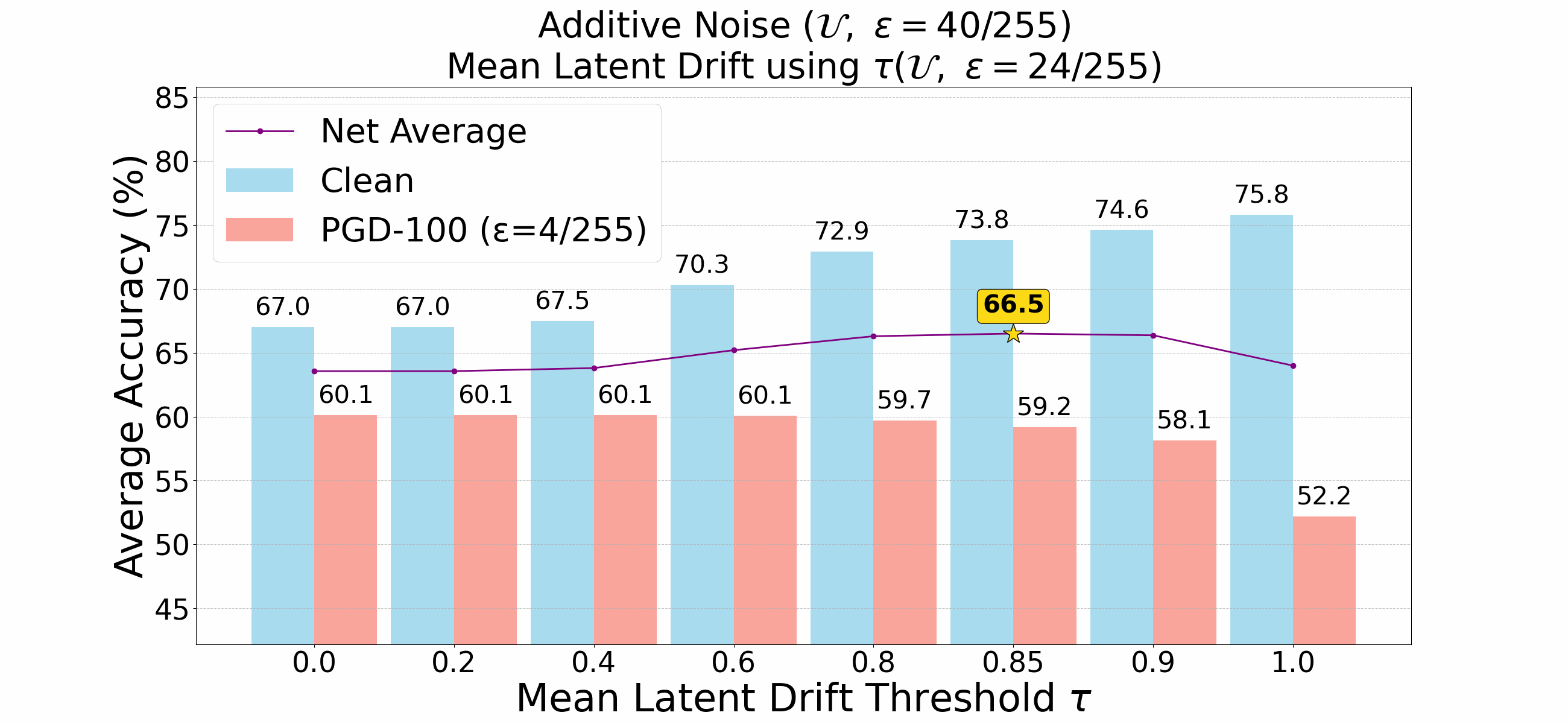}

                \vspace{0.5em}

        \includegraphics[width=\linewidth]{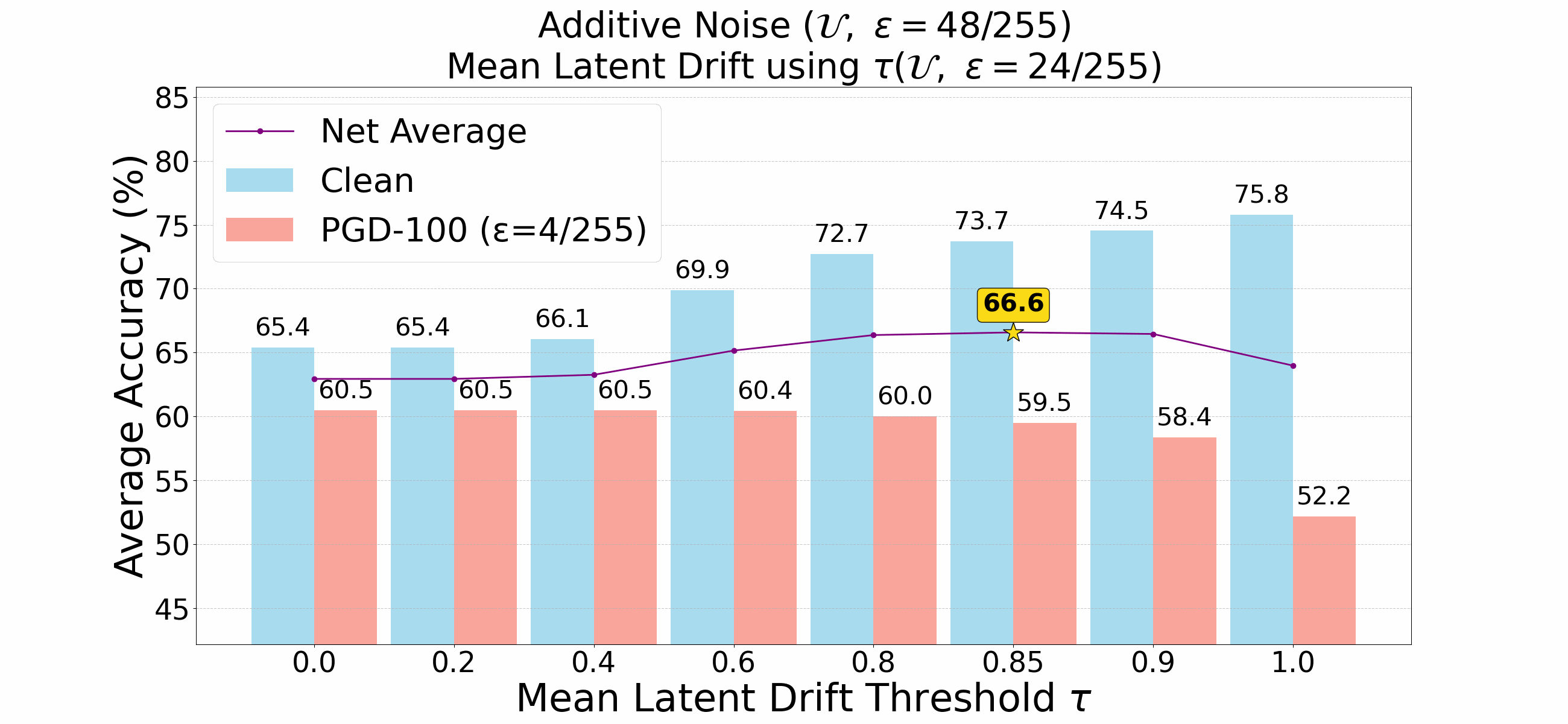}
    \end{minipage}
    \begin{minipage}{0.22\textwidth}
\caption{\small Performance of drift-gated stochastic intervention using uniform noise. Clean (blue), adversarial (red; PGD-100, $\epsilon=4/255$), and average accuracy (purple) are reported after selectively injecting uniform noise into the input. The intervention is applied only to samples whose mean latent drift exceeds a threshold $\tau$, where latent drift is computed under a uniform noise best for that setting. Each plot corresponds to a different level of  noise strength ($\epsilon \in \{24, 32, 40, 48\}/255$) injected on to the samples. For every intervention setting, the probe noise used to compute mean latent drift is chosen to provide the best clean--robust trade-off (indicated in the figure). As $\tau$ increases, the intervention is applied more selectively, improving clean accuracy while largely preserving adversarial robustness. The best trade-off occurs near $\tau_{\text{threshold}} \approx 0.85$ for stronger intervention strengths ($\epsilon=40/255$ and $48/255$).}
        \label{fig:results_eval_uniform_noise_threshold_vit_l_14_datacomp}
    \end{minipage}
    \vspace{-1em}
\end{figure}

\begin{figure}[t]
\centering
\small

\includegraphics[width=\linewidth,height=0.28\textheight,keepaspectratio]{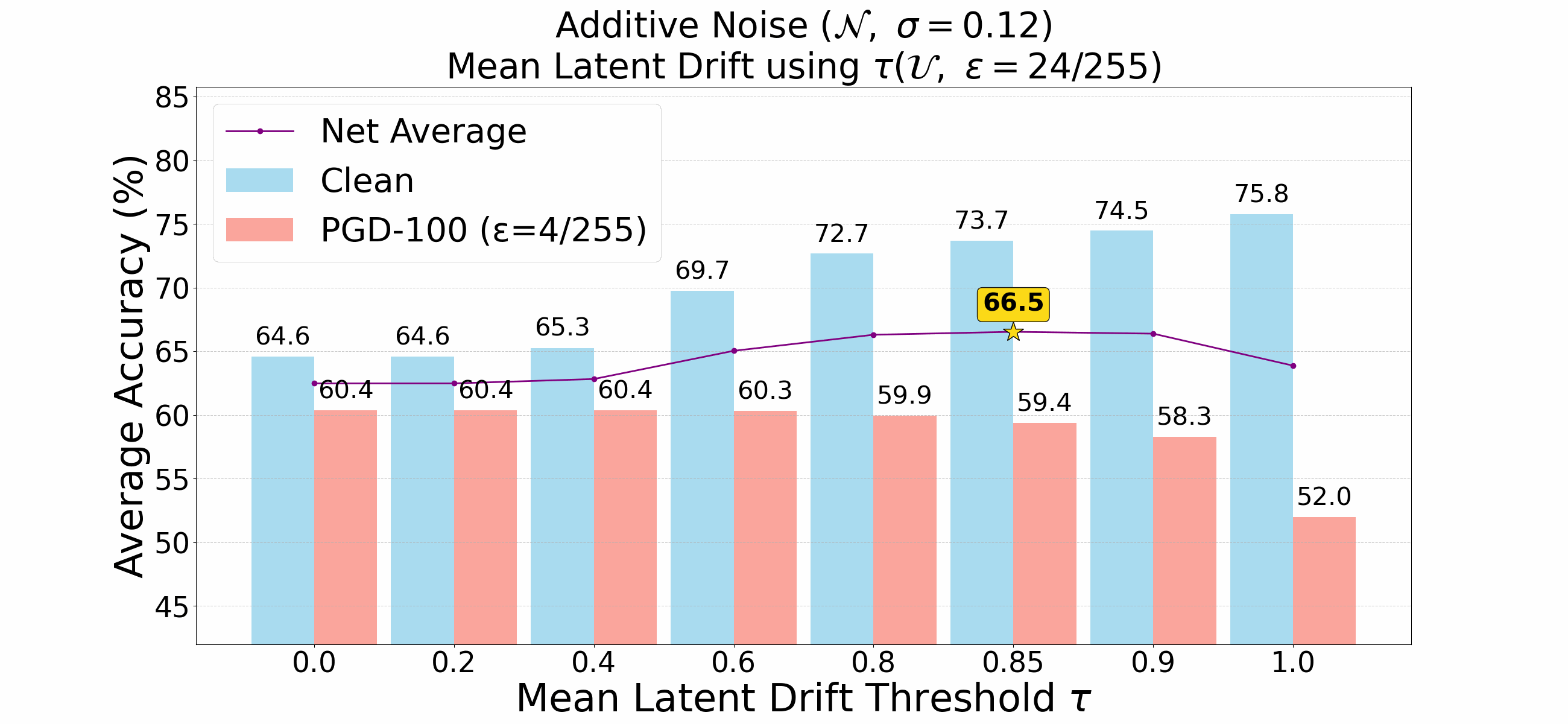}

\includegraphics[width=\linewidth]{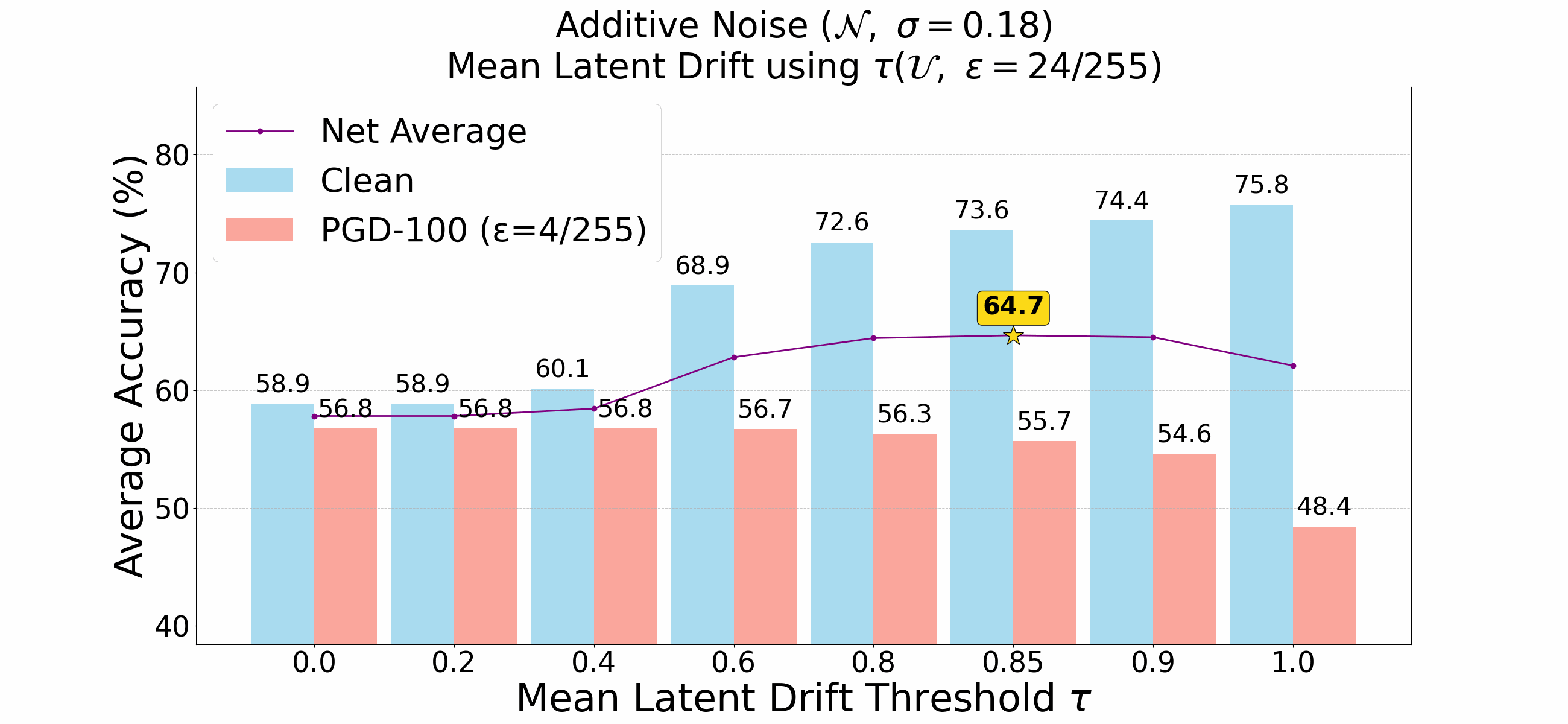}

\includegraphics[width=\linewidth]{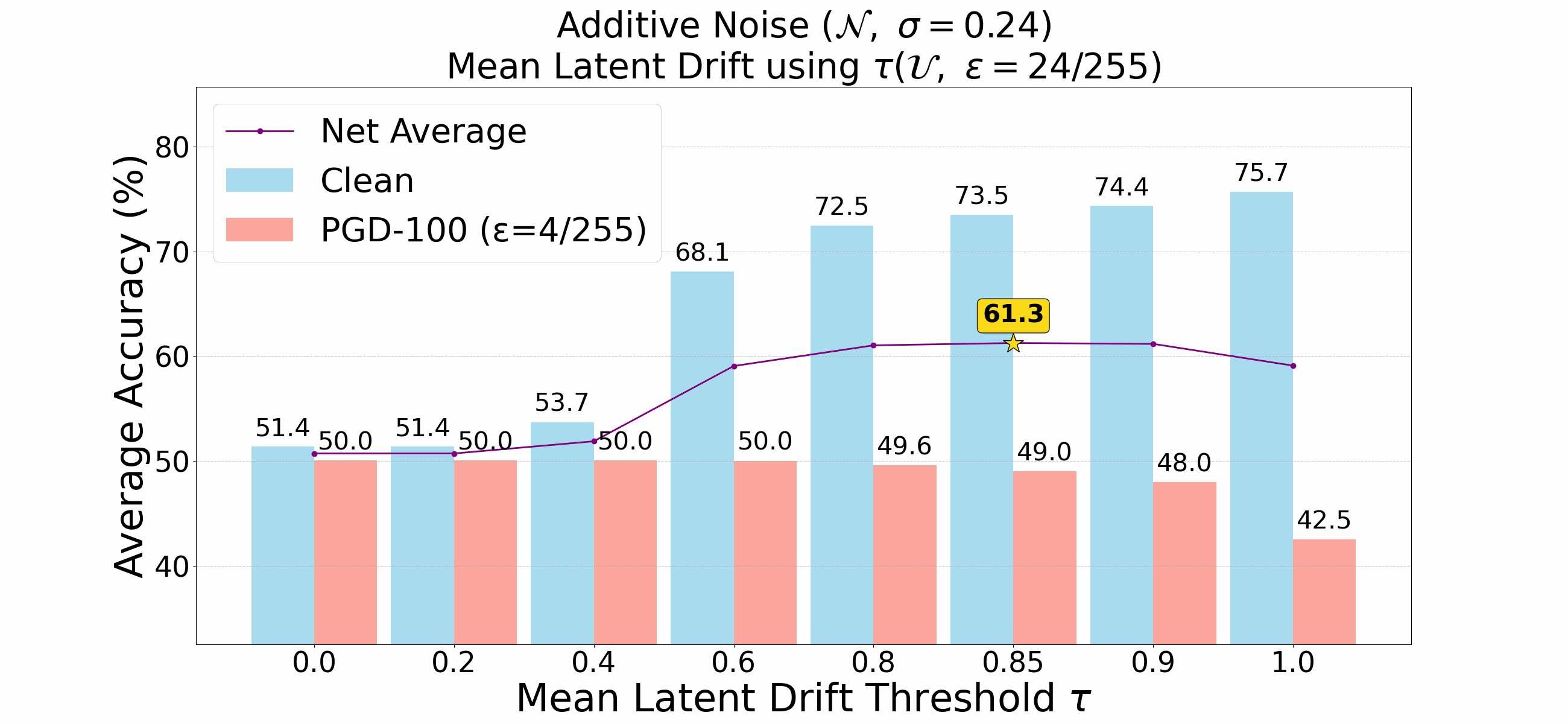}

\caption{\small Performance of drift-gated stochastic intervention using Gaussian noise. Clean (blue), adversarial (red; PGD-100, $\epsilon=4/255$), and average accuracy (purple) are reported after selectively injecting Gaussian noise into the input.  Each plot corresponds to a different level of Gaussian noise strength ($\sigma \in \{0.12, 0.18, 0.24\}$) injected onto the samples. For every intervention setting, the probe noise used to compute mean latent drift is chosen to provide the best clean--robust trade-off (indicated in the figure). The best clean--robust trade-off is achieved at $\sigma=0.12$ with $\tau_{\text{threshold}} \approx 0.85$ (highlighted).}

\label{fig:results_eval_gaussian_noise_threshold_vit_l_14_datacomp}

\vspace{-1em}
\end{figure}

Motivated by the clean--adversarial drift 
separation in the high-noise regime, 
Figs.~\ref{fig:results_eval_uniform_noise_threshold_vit_l_14_datacomp} 
and~\ref{fig:results_eval_gaussian_noise_threshold_vit_l_14_datacomp} 
evaluate our drift-gated intervention, 
where noise is selectively injected only 
onto inputs exhibiting adversarial-like 
instability. As the drift threshold $\gamma$ 
increases, the intervention is applied more 
selectively, recovering clean accuracy while 
largely preserving adversarial robustness.

\subsubsection{ViT-L/14}

\begin{figure}[t]
    \centering
    \includegraphics[width=\linewidth]{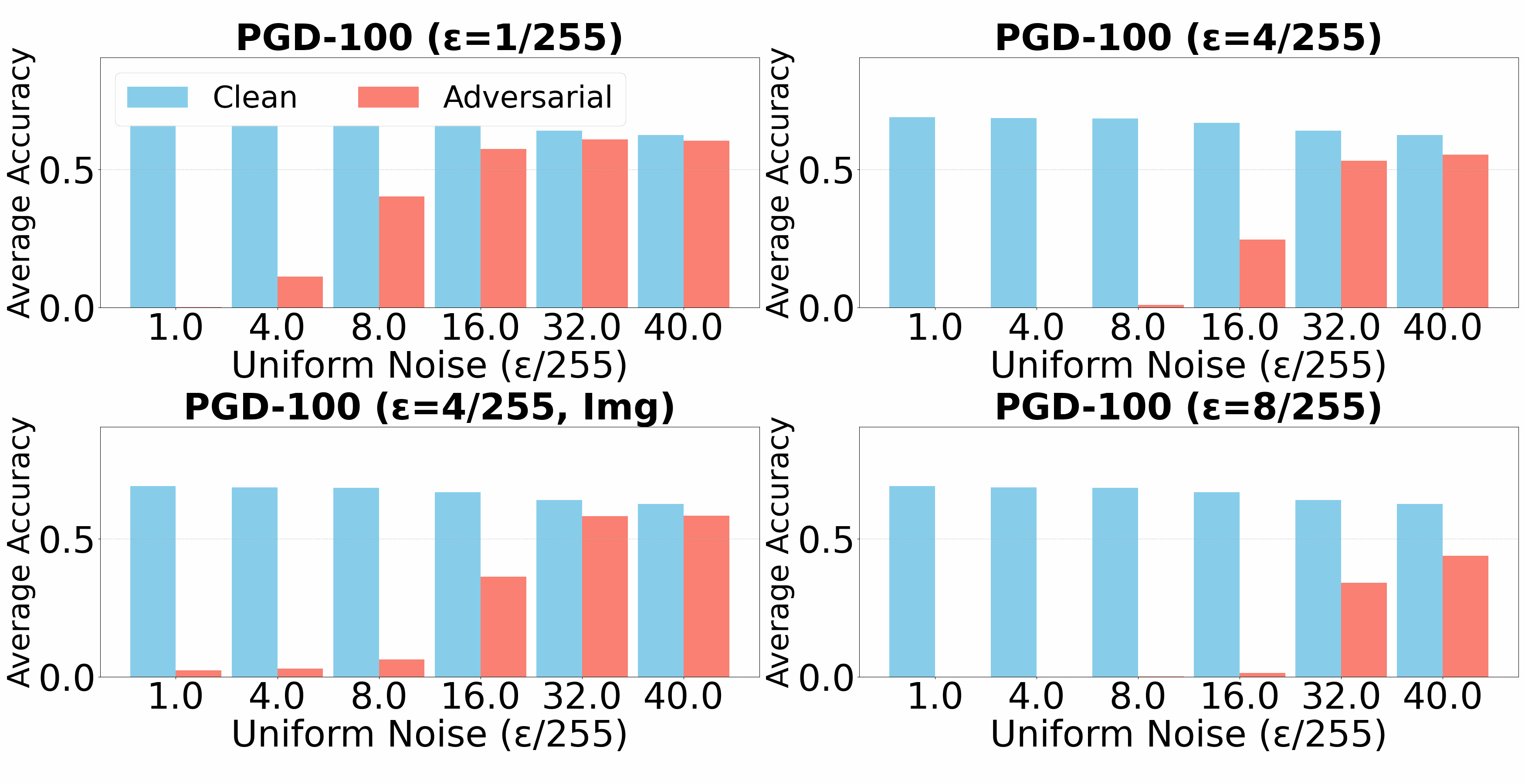}
\caption{\textbf{ViT-L-14.} Clean and adversarial accuracy versus uniform noise strength, averaged across eight fine-grained datasets. Adversarial examples (PGD-100) are generated at different perturbation budgets and with two attack objectives: maximizing cross-entropy loss and maximizing the discrepancy in visual features (Img). Across all settings, stronger noise leads to clear improvements in adversarial accuracy, while weak noise yields little robustness benefit and is insufficient to suppress adversarial perturbations.}
    \label{fig:app_uniform_noise_accuracy_anlalysis_eps_all_vit_l_14}
    \vspace{-1em}

\end{figure}

\begin{figure}[t]
    \centering
    \includegraphics[width=\linewidth]{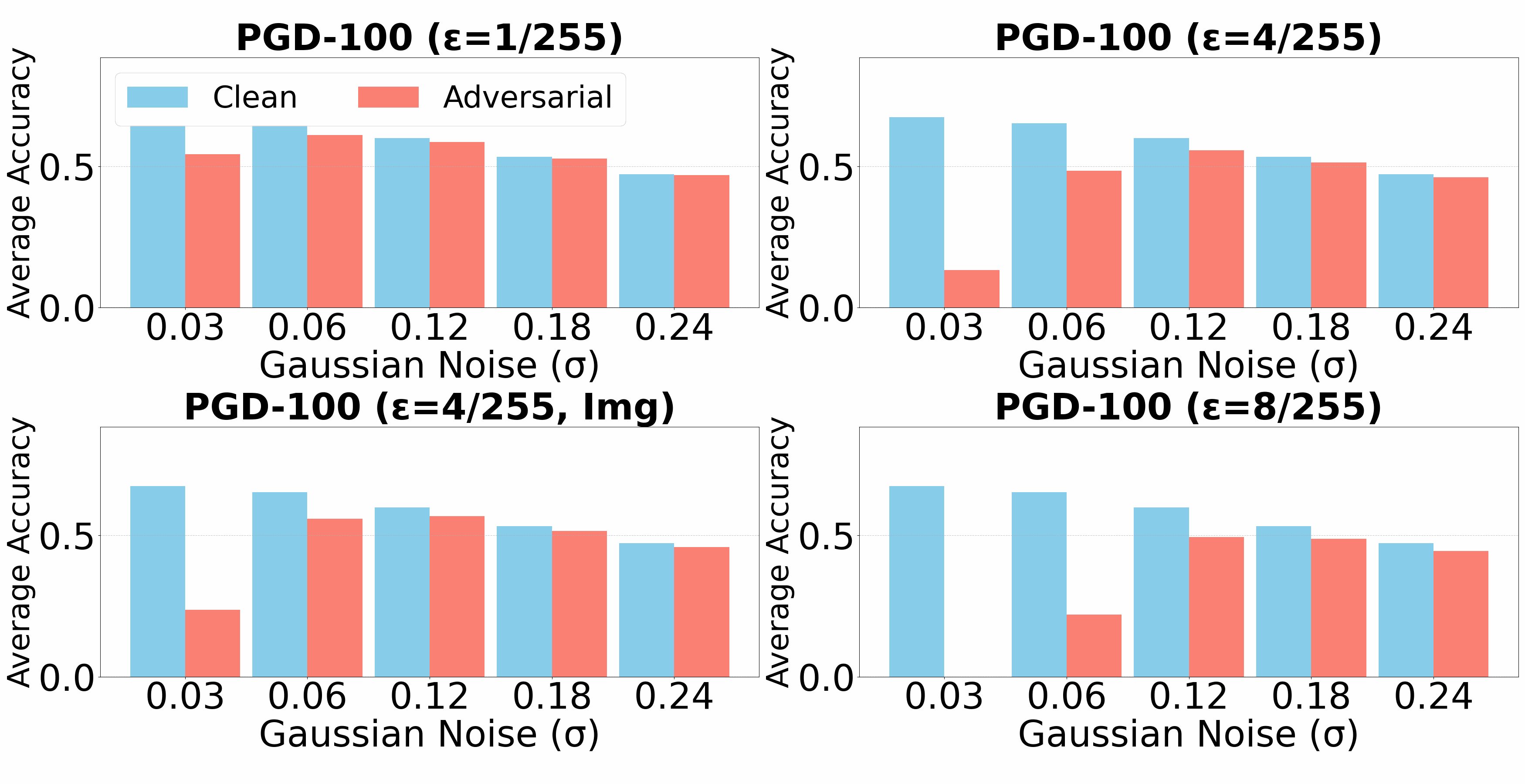}
\caption{\textbf{ViT-L-14.} Clean and adversarial accuracy versus Gaussian noise strength, averaged across eight fine-grained datasets. Adversarial examples (PGD-100) are generated at different perturbation budgets and with two attack objectives: maximizing cross-entropy loss and maximizing the discrepancy in visual features (Img). Across settings, moderate Gaussian noise improves adversarial accuracy, whereas stronger noise causes both clean and adversarial performance to drop and gradually converge, reflecting the loss of useful image semantics at high noise levels.}
    \label{fig:app_gaussian_noise_accuracy_anlalysis_eps_all_vit_l_14}
    \vspace{-1em}

\end{figure}

Figs.~\ref{fig:app_uniform_noise_accuracy_anlalysis_eps_all_vit_l_14} 
and~\ref{fig:app_gaussian_noise_accuracy_anlalysis_eps_all_vit_l_14} 
report the same evaluation for the original 
CLIP ViT-L/14. The same trends are observed: 
weak noise injections provide little 
robustness benefit, while stronger 
injections substantially improve adversarial 
accuracy at the cost of gradually degrading 
clean performance, consistently across noise 
types and attack settings.

\subsubsection{FARE}

\begin{figure}[t]
    \centering
    \includegraphics[width=\linewidth]{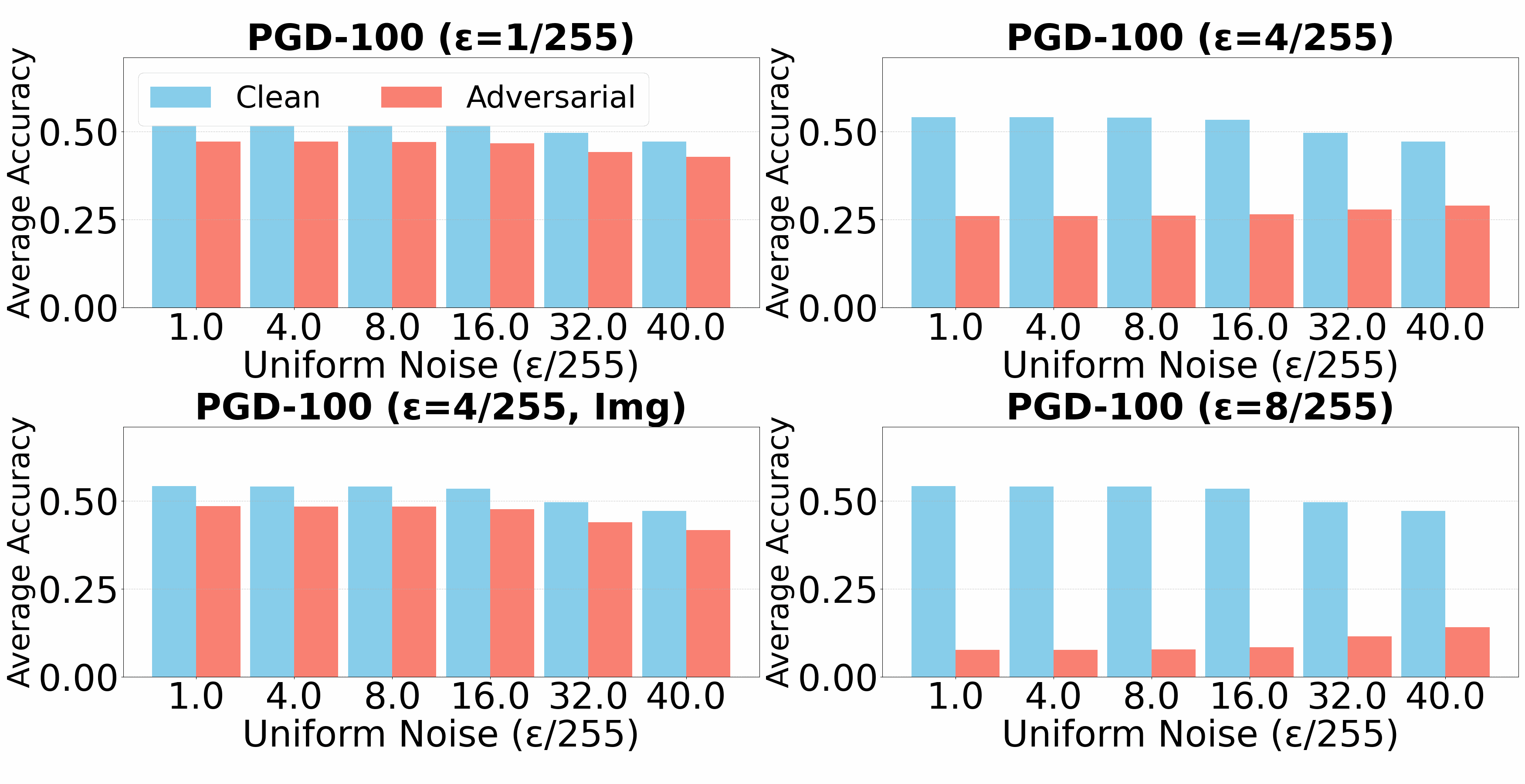}
\caption{\textbf{FARE~\cite{schlarmann2024robust}.} Clean and adversarial accuracy versus uniform noise strength, averaged across eight fine-grained datasets. Adversarial examples (PGD-100) are generated at different perturbation budgets and with two attack objectives: maximizing cross-entropy loss and maximizing the discrepancy in visual features (Img). In contrast to standard CLIP models, noise injection does not improve adversarial accuracy; clean and adversarial performance vary similarly with increasing noise, indicating that adversarially trained models are less sensitive to the noise-regime effects observed in non-robust models.}
    \label{fig:app_uniform_noise_accuracy_anlalysis_eps_all_fare4}
    \vspace{-1em}

\end{figure}

\begin{figure}[t]
    \centering
    \includegraphics[width=\linewidth]{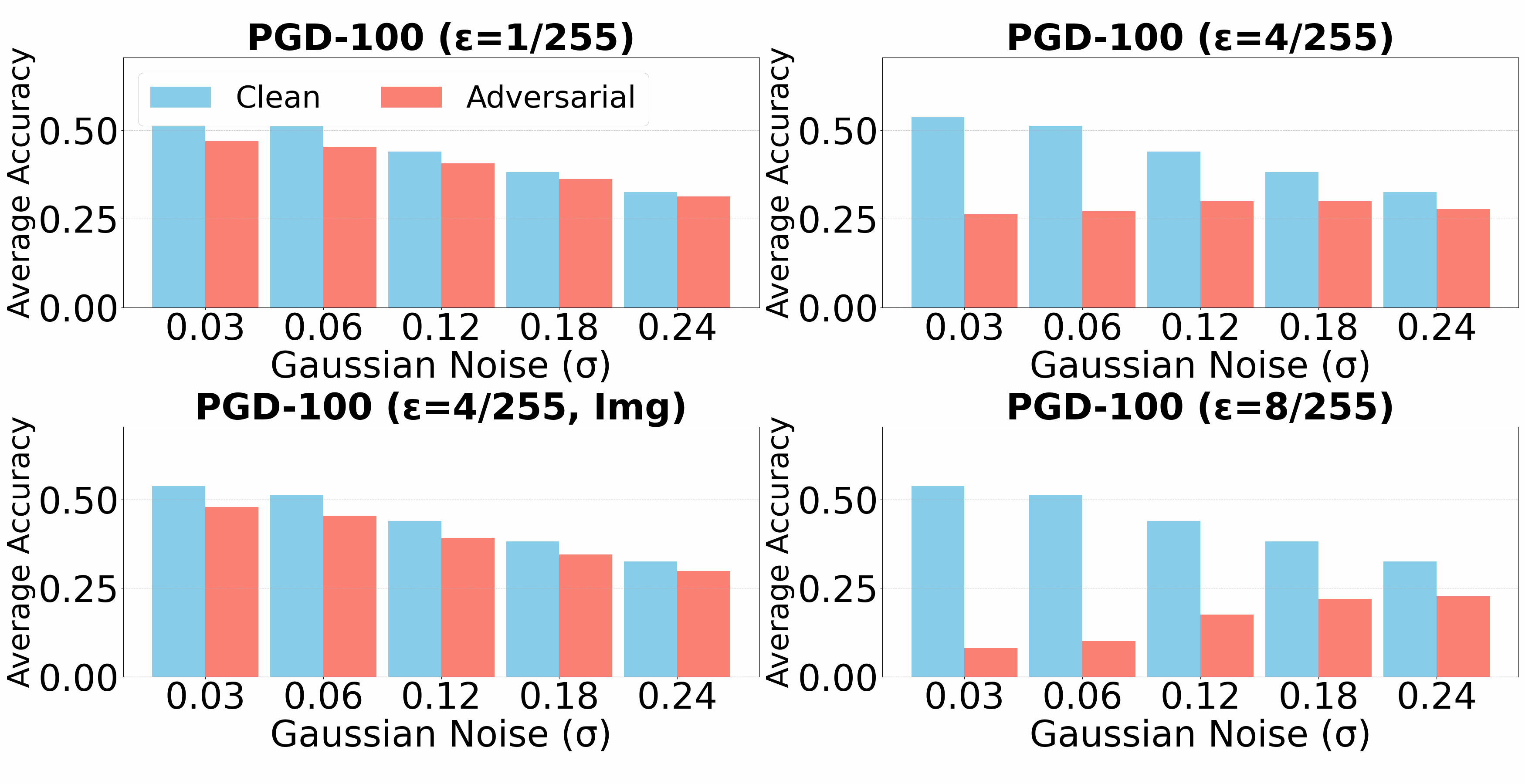}
\caption{\textbf{FARE~\cite{schlarmann2024robust}.} Clean and adversarial accuracy versus Gaussian noise strength, averaged across eight fine-grained datasets. Adversarial examples (PGD-100) are generated at different perturbation budgets and with two attack objectives: maximizing cross-entropy loss and maximizing the discrepancy in visual features (Img). In contrast to standard CLIP models, noise injection does not improve adversarial accuracy; clean and adversarial performance vary similarly with increasing noise, indicating that adversarially trained models are less sensitive to the noise-regime effects observed in non-robust models.}
    \label{fig:app_gaussian_noise_accuracy_anlalysis_eps_all_fare4}
    \vspace{-1em}

\end{figure}

Figs.~\ref{fig:app_uniform_noise_accuracy_anlalysis_eps_all_fare4} 
and~\ref{fig:app_gaussian_noise_accuracy_anlalysis_eps_all_fare4} 
report the same evaluation for 
FARE~\cite{schlarmann2024robust}. In contrast 
to non-robust CLIP models, injecting 
stochastic perturbations onto inputs does 
not meaningfully improve adversarial 
accuracy; clean and adversarial performance 
vary similarly with noise strength, 
consistent with adversarial training 
reducing sensitivity to the noise-regime 
effects observed in non-robust models.

\subsubsection{DeltaCLIP-L}

\begin{figure}[t]
    \centering
    \includegraphics[width=\linewidth]{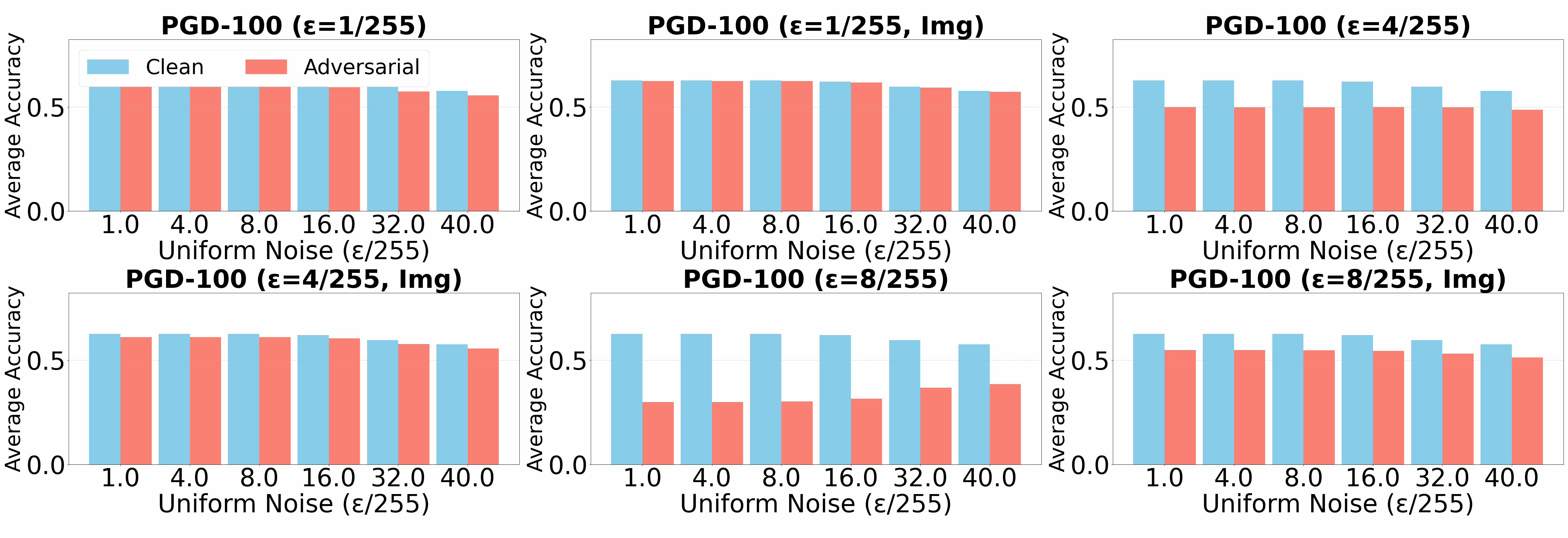}
\caption{\textbf{DeltaCLIP-L~\cite{wang2025double}.} Clean and adversarial accuracy versus uniform noise strength, averaged across eight fine-grained datasets. Adversarial examples (PGD-100) are generated at different perturbation budgets and with two attack objectives: maximizing cross-entropy loss and maximizing the discrepancy in visual features (Img). In contrast to standard CLIP models, noise injection does not improve adversarial accuracy; clean and adversarial performance vary similarly with increasing noise, indicating that adversarially trained models are less sensitive to the noise-regime effects observed in non-robust models.}
    \label{fig:app_uniform_noise_accuracy_anlalysis_eps_all_deltaclip}
    \vspace{-1em}

\end{figure}

\begin{figure}[t]
    \centering
    \includegraphics[width=\linewidth]{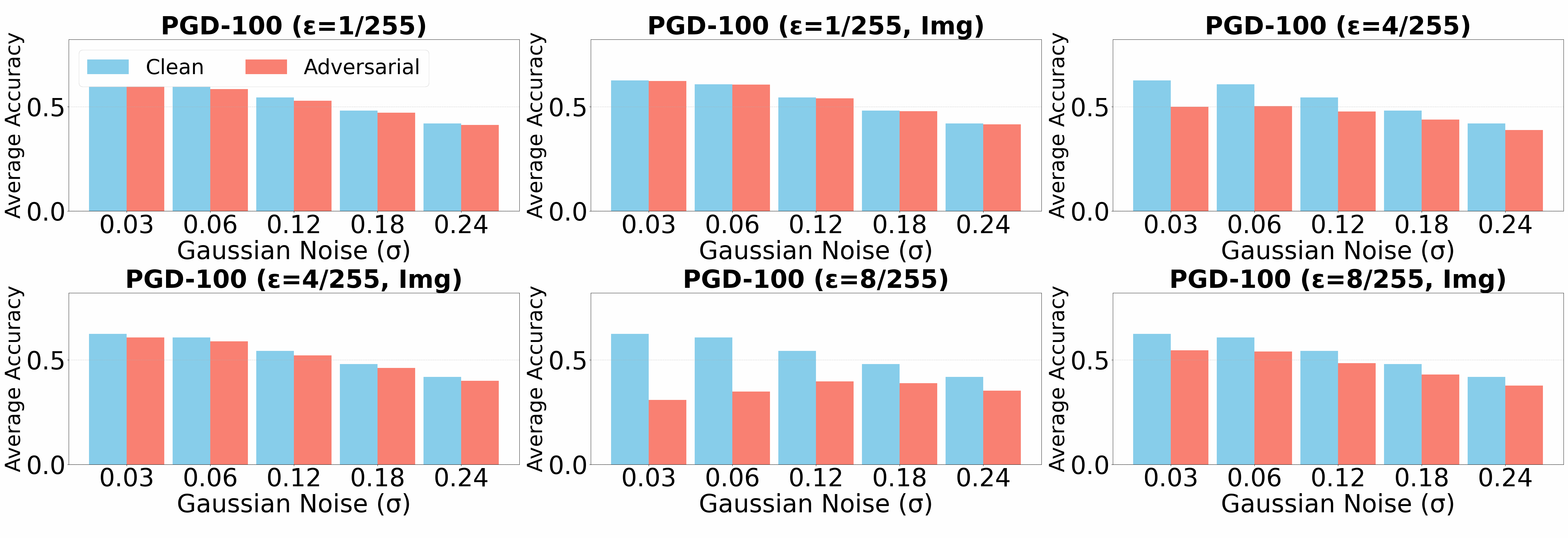}
\caption{\textbf{DeltaCLIP-L~\cite{wang2025double}.} Clean and adversarial accuracy versus Gaussian noise strength, averaged across eight fine-grained datasets. Adversarial examples (PGD-100) are generated at different perturbation budgets and with two attack objectives: maximizing cross-entropy loss and maximizing the discrepancy in visual features (Img). In contrast to standard CLIP models, noise injection does not improve adversarial accuracy; clean and adversarial performance vary similarly with increasing noise, indicating that adversarially trained models are less sensitive to the noise-regime effects observed in non-robust models.}
    \label{fig:app_gaussian_noise_accuracy_anlalysis_eps_all_deltaclip}
    \vspace{-1em}

\end{figure}

Figs.~\ref{fig:app_uniform_noise_accuracy_anlalysis_eps_all_deltaclip} 
and~\ref{fig:app_gaussian_noise_accuracy_anlalysis_eps_all_deltaclip} 
report the same evaluation for 
DeltaCLIP-L~\cite{wang2025double}. As with 
FARE, injecting stochastic perturbations 
onto inputs does not produce clear 
robustness improvements; clean and 
adversarial accuracy follow similar trends 
as noise strength increases. Together, 
these results confirm that adversarial 
training aligns the behavior of clean and 
adversarial representations, reducing the 
effectiveness of noise-based interventions.

\subsection{Evaluating Performance with Test-Time Counter Attacks}
\label{sec:app_eval_ttc}

\begin{figure}[t]
    \small \centering
    \begin{minipage}{0.75\textwidth}
        \centering


        \includegraphics[width=\linewidth]{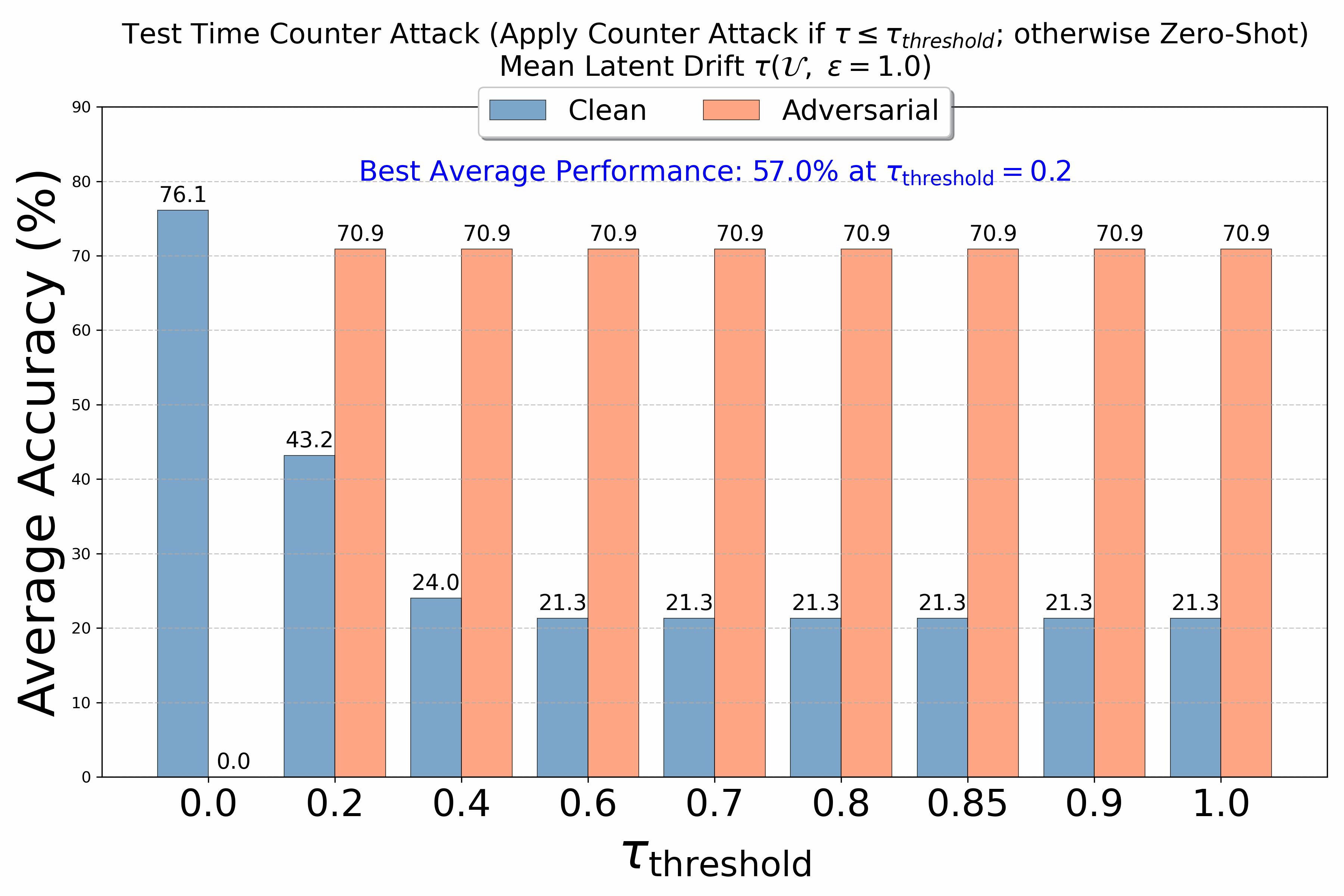}

                \vspace{0.5em}

        \includegraphics[width=\linewidth]{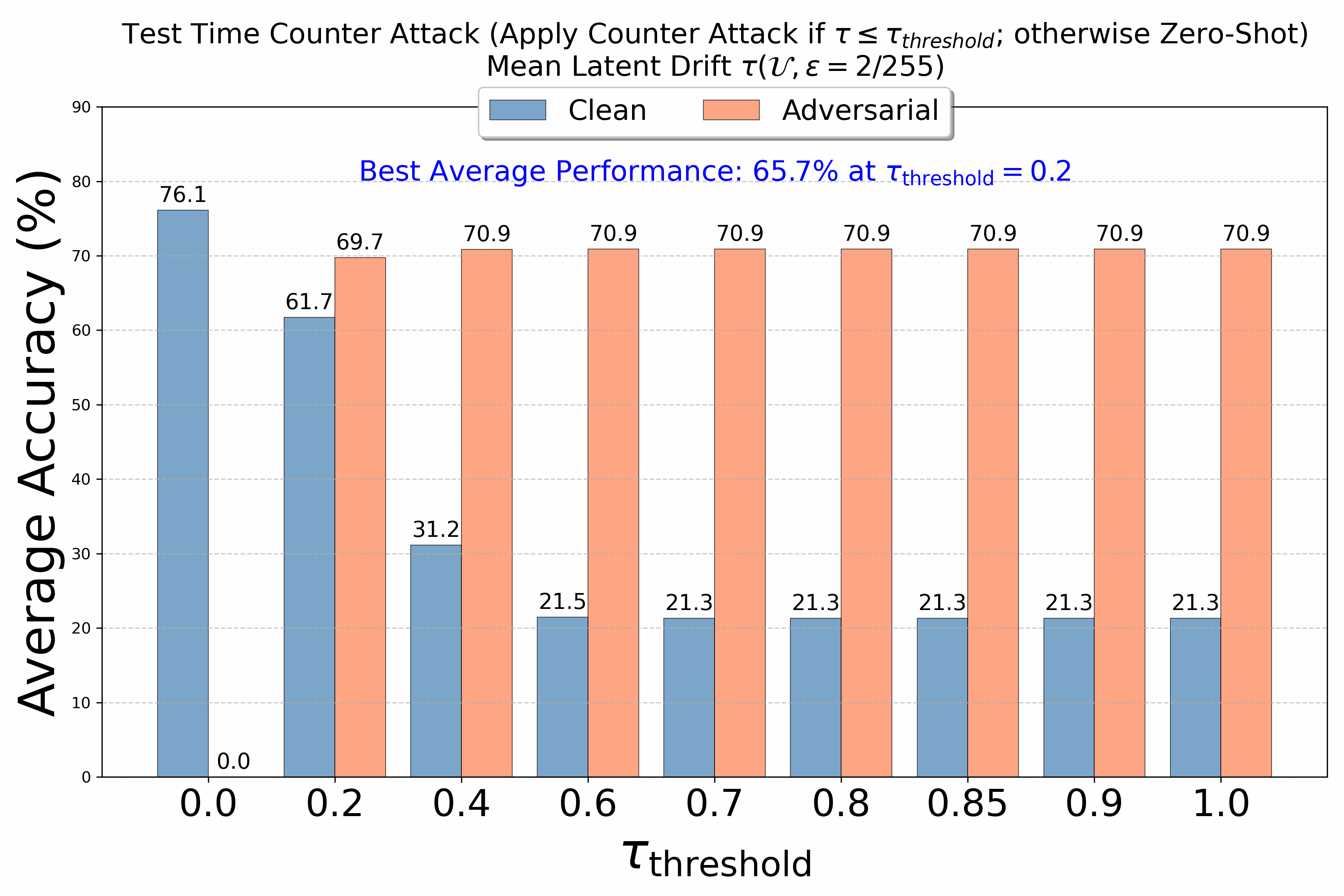}

                \vspace{0.5em}

        \includegraphics[width=\linewidth]{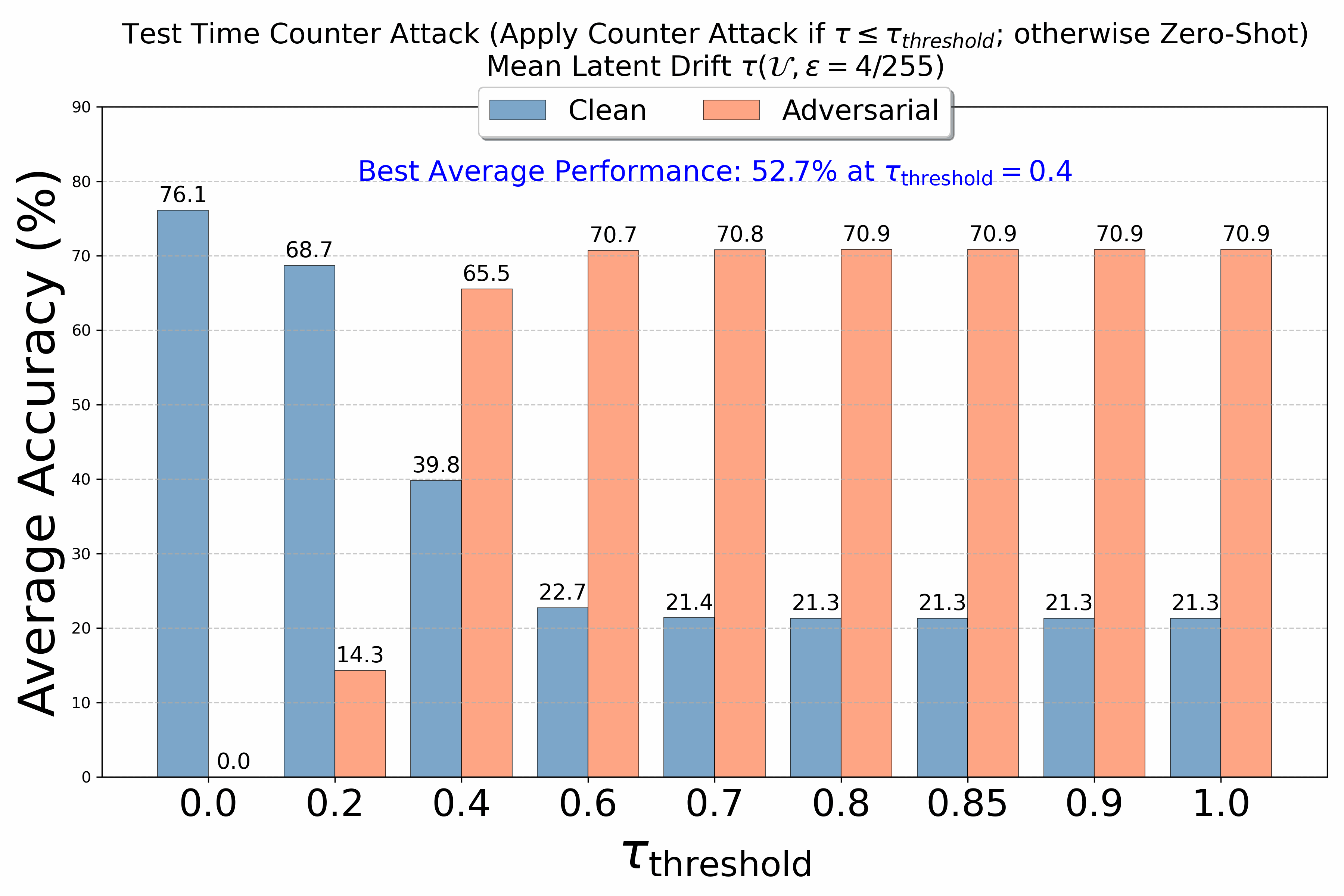}
    \end{minipage}
    \begin{minipage}{0.24\textwidth}
\caption{Standard TTC~\cite{xing2025clip} 
evaluation averaged across eight fine-grained 
datasets. Clean (blue) and adversarial (red; 
PGD-100, $\epsilon=4/255$) accuracy are shown 
as a function of the false-stability threshold 
$\tau_{\text{TTC}}$, below which the 
counterattack is triggered. The gating signal 
is computed under weak uniform probe noise 
$\epsilon \in \{1, 2, 4\}/255$, following the 
original TTC protocol~\cite{xing2025clip}. 
The best clean--robust trade-off is achieved 
at $\epsilon=2/255$ with 
$\tau_{\text{TTC}}=0.2$, yielding an average 
accuracy of $65.70\%$ (highlighted), which 
is adopted as the TTC baseline in the main 
paper.}
        \label{fig:results_eval_uniform_noise_ttc_default_vit_l_14_datacomp}
    \end{minipage}
    \vspace{-1em}
\end{figure}

We begin by evaluating the original TTC 
strategy~\cite{xing2025clip}, which computes 
a false-stability signal $\tau_{\text{TTC}}$ 
under weak uniform noise to differentiate 
between clean and adversarial samples and 
trigger the counterattack accordingly. 
Figure~\ref{fig:results_eval_uniform_noise_ttc_default_vit_l_14_datacomp} 
reports results across the eight fine-grained 
datasets, sweeping the false-stability 
threshold under probe noise strengths 
$\epsilon \in \{1/255, 2/255, 4/255\}$ as used 
in~\cite{xing2025clip}. Based on these results, 
$\epsilon=2/255$ with threshold 
$\tau_{\text{TTC}}=0.2$ yields the best average 
performance of $65.70\%$, which we adopt as 
the TTC baseline reported in the main paper. 
While this configuration represents the best 
achievable under the weak-noise false-stability 
signal, the probe noise at this strength 
provides an unreliable separation between 
clean and adversarial inputs, resulting in a 
suboptimal clean--robust trade-off. More than $25\%$ of the clean samples are incorrectly selected for counterattack intervention, leading to significant drop in clean performance.

\begin{figure}[t]
    \small \centering
    \begin{minipage}{0.75\textwidth}
        \centering


        \includegraphics[width=\linewidth]{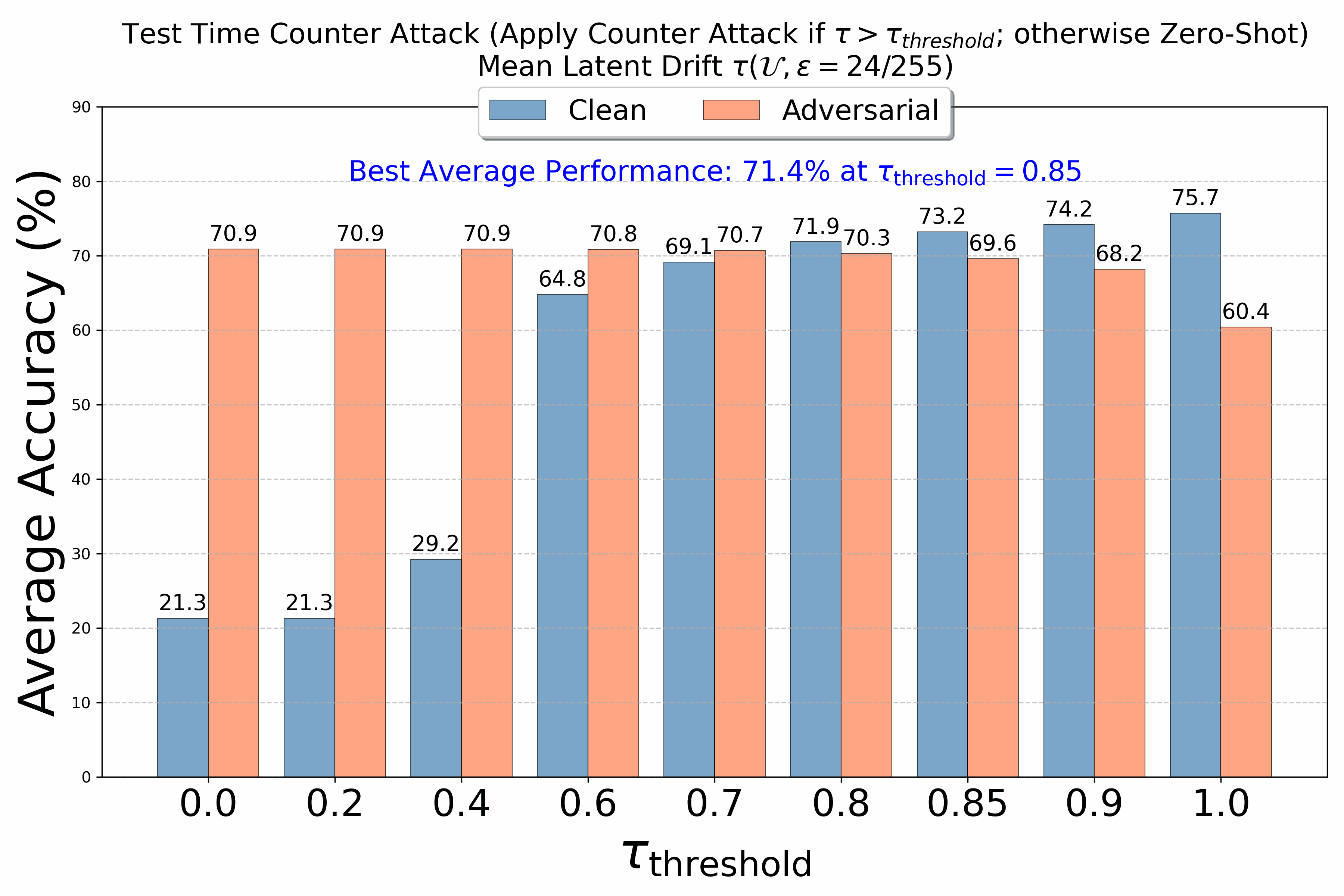}

                \vspace{0.5em}

        \includegraphics[width=\linewidth]{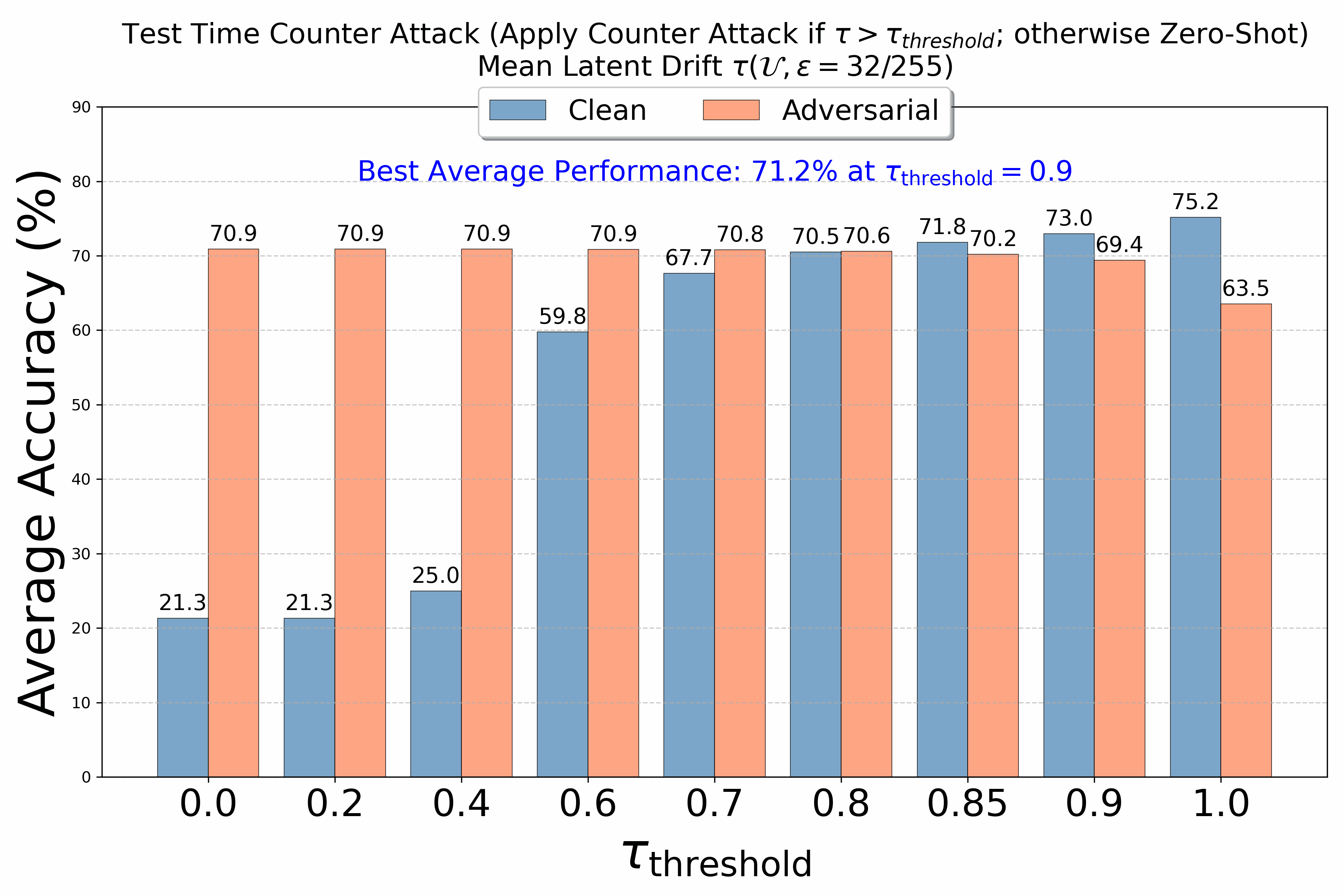}

                \vspace{0.5em}

        \includegraphics[width=\linewidth]{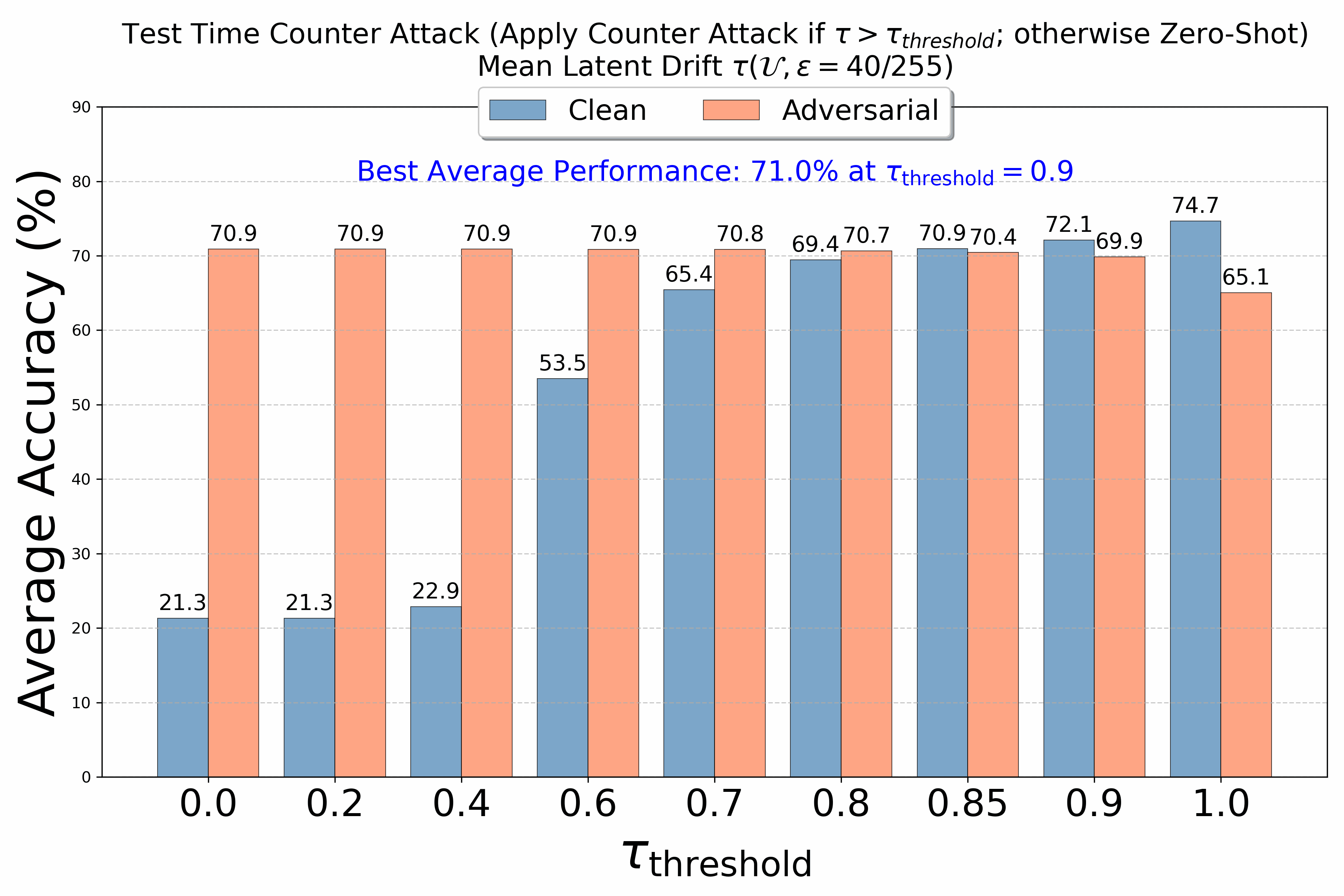}
    \end{minipage}
    \begin{minipage}{0.24\textwidth}
\caption{\textbf{Ours:} Drift-gated TTC 
averaged across eight fine-grained datasets 
using uniform probe noise. Clean (blue) and 
adversarial (red; PGD-100, $\epsilon=4/255$) 
accuracy are shown as a function of the drift 
threshold $\gamma$, above which the 
counterattack is triggered. Unlike TTC's 
weak-noise false-stability trigger, our 
gating signal is computed under high-noise 
uniform probe strengths 
$\epsilon \in \{24, 32, 40\}/255$, providing 
a more reliable clean--adversarial separation. 
As $\gamma$ increases, TTC is applied more 
selectively, improving clean accuracy while 
largely preserving adversarial robustness. 
The best average accuracy and corresponding 
threshold for each setting are highlighted.}
        \label{fig:results_eval_uniform_noise_ttc_vit_l_14_datacomp}
    \end{minipage}
    \vspace{-1em}
\end{figure}

\begin{figure}[t]
    \small \centering
    \begin{minipage}{0.75\textwidth}
        \centering


        \includegraphics[width=\linewidth]{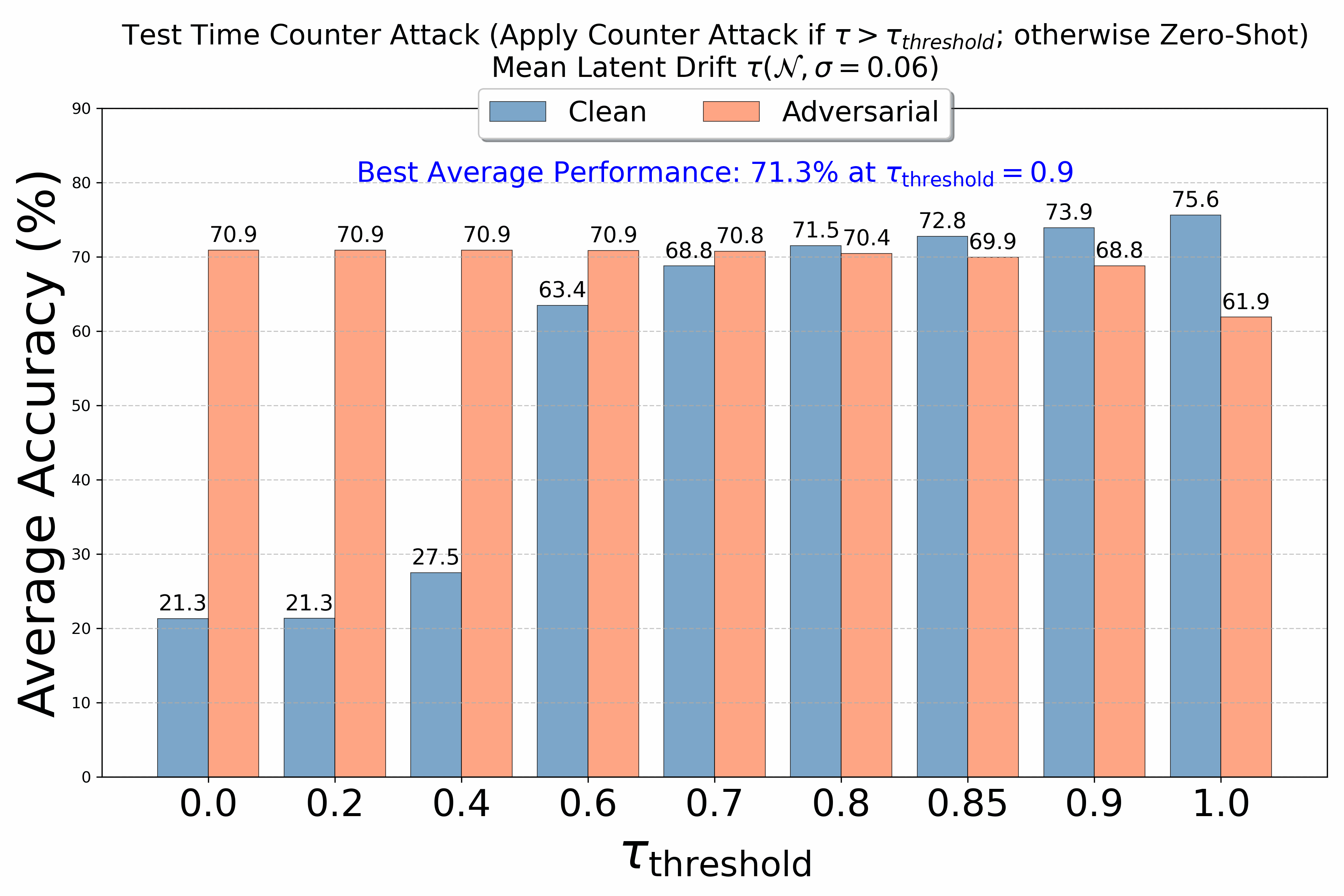}

                \vspace{0.5em}

        \includegraphics[width=\linewidth]{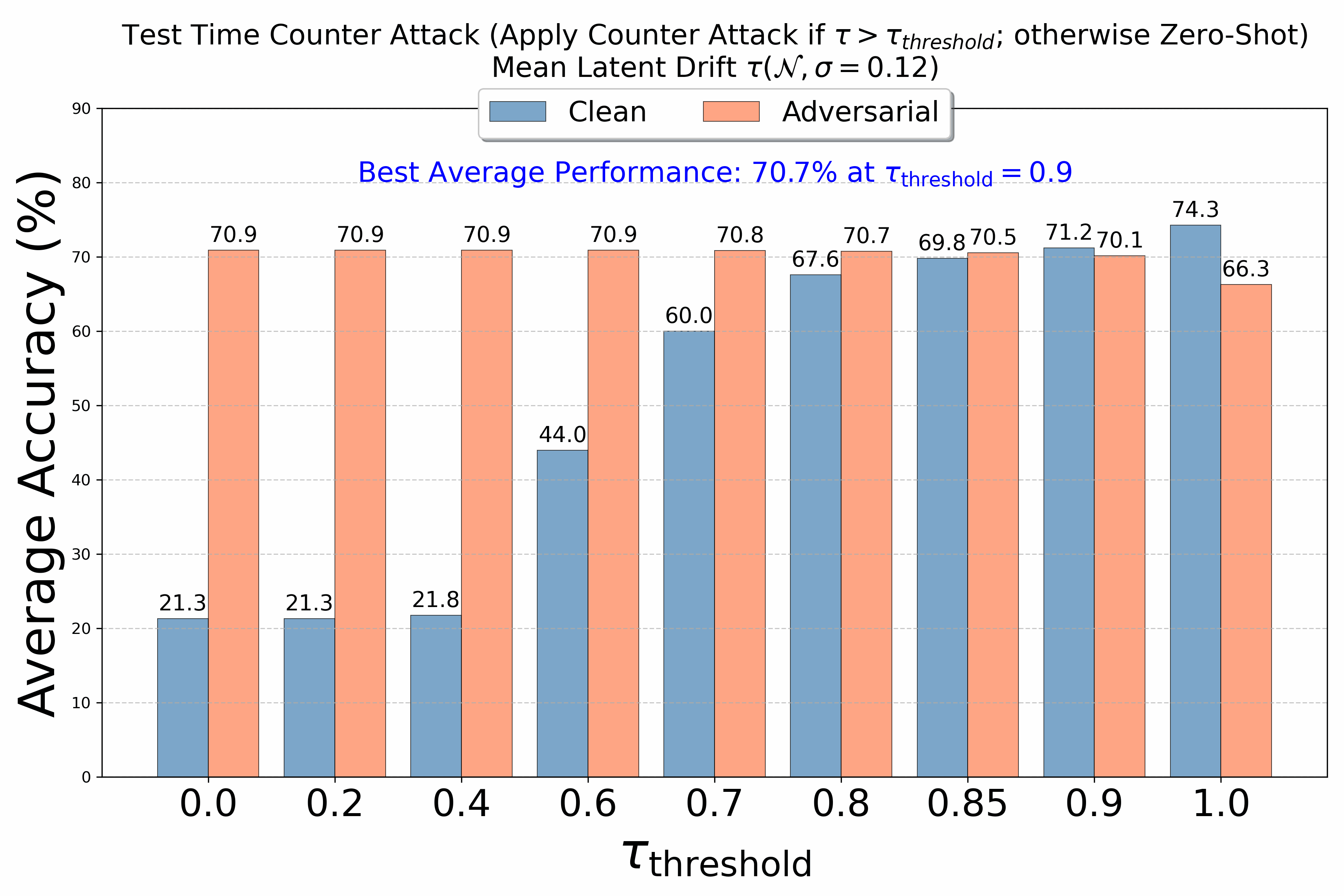}

                \vspace{0.5em}

        \includegraphics[width=\linewidth]{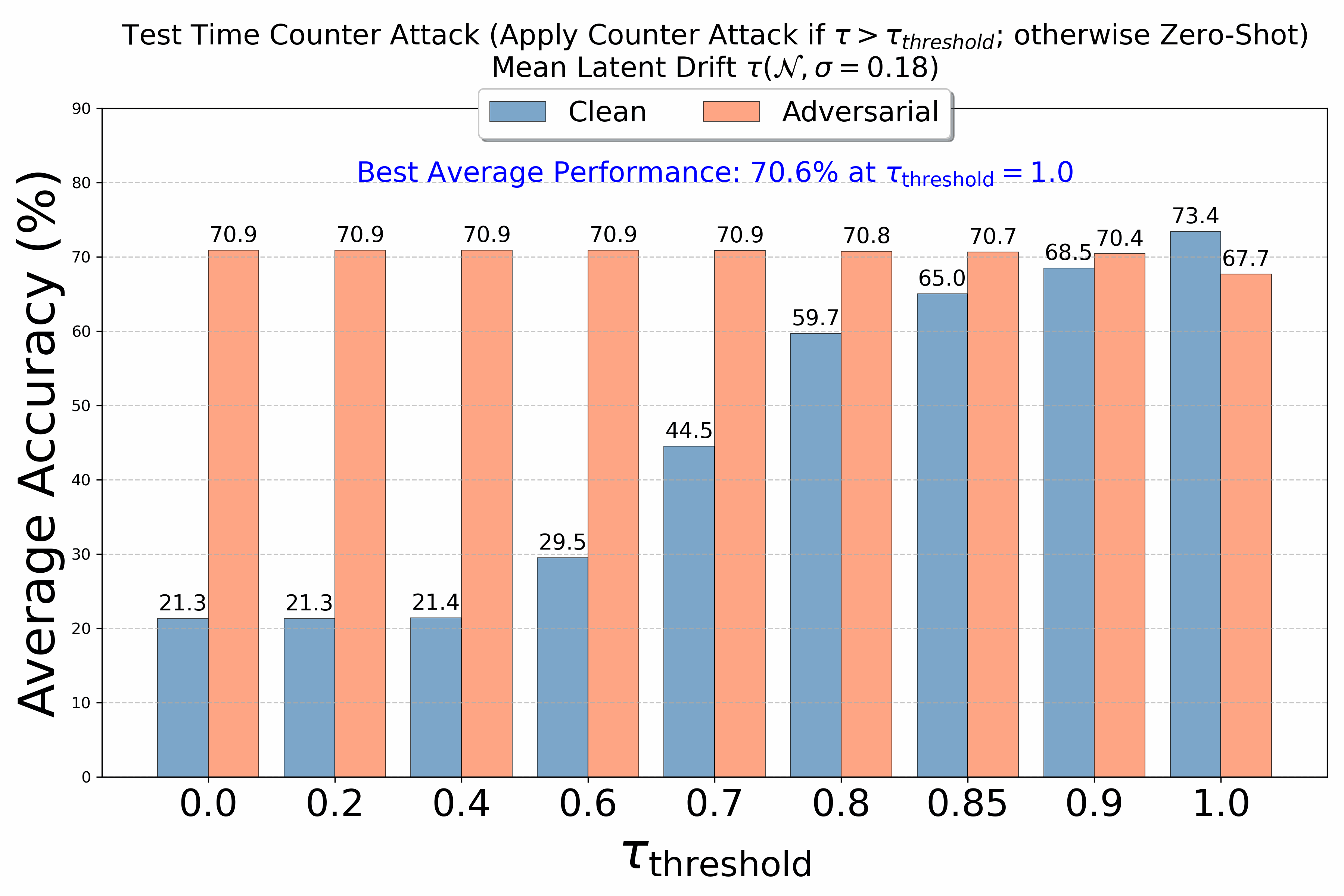}

    \end{minipage}
    \begin{minipage}{0.24\textwidth}
\caption{\textbf{Ours:} Drift-gated TTC 
averaged across eight fine-grained datasets 
using Gaussian probe noise. Clean (blue) and 
adversarial (red; PGD-100, $\epsilon=4/255$) 
accuracy are shown as a function of the drift 
threshold $\gamma$, above which the 
counterattack is triggered. Unlike TTC's 
weak-noise false-stability trigger, our 
gating signal is computed under high-noise 
Gaussian probe strengths 
$\sigma \in \{0.06, 0.12, 0.18\}$, providing 
a more reliable clean--adversarial separation. 
As $\gamma$ increases, TTC is applied more 
selectively, improving clean accuracy while 
largely preserving adversarial robustness. 
The best average accuracy and corresponding 
threshold for each setting are highlighted.}
        \label{fig:results_eval_gaussian_noise_ttc_vit_l_14_datacomp}
    \end{minipage}
    \vspace{-1em}
\end{figure}

We then evaluate our high-noise drift-gated 
variant, which replaces the weak-noise trigger 
with a gating signal computed in the high-noise 
regime. This regime provides a clearer 
separation between clean and adversarial 
inputs, allowing TTC to be applied more 
selectively. 
Figures~\ref{fig:results_eval_uniform_noise_ttc_vit_l_14_datacomp} 
and~\ref{fig:results_eval_gaussian_noise_ttc_vit_l_14_datacomp} 
report results averaged across the eight 
fine-grained datasets for uniform and Gaussian 
probe noise respectively. As the threshold 
$\gamma$ increases, the intervention becomes 
more selective, improving clean accuracy while 
largely preserving adversarial robustness. The 
best average accuracy and corresponding 
threshold for each configuration are 
highlighted in the figures. Using our high noise drift-gated strategy, leads to less than $10\%$ of clean samples being selected for defensive intervention.

\begin{figure}[t]
    \small \centering
    \begin{minipage}{0.75\textwidth}
        \centering


        \includegraphics[width=\linewidth]{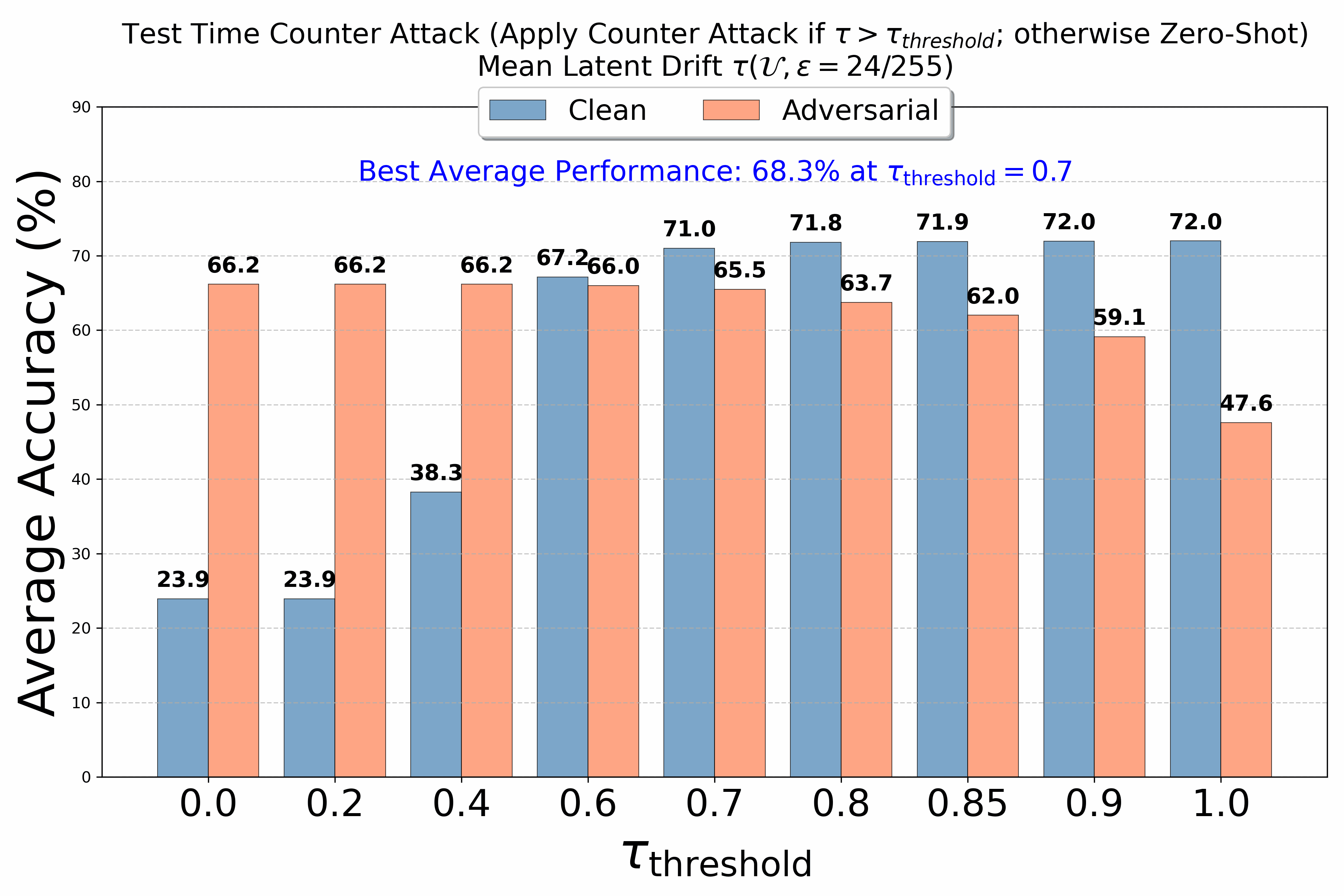}

                \vspace{0.5em}

        \includegraphics[width=\linewidth]{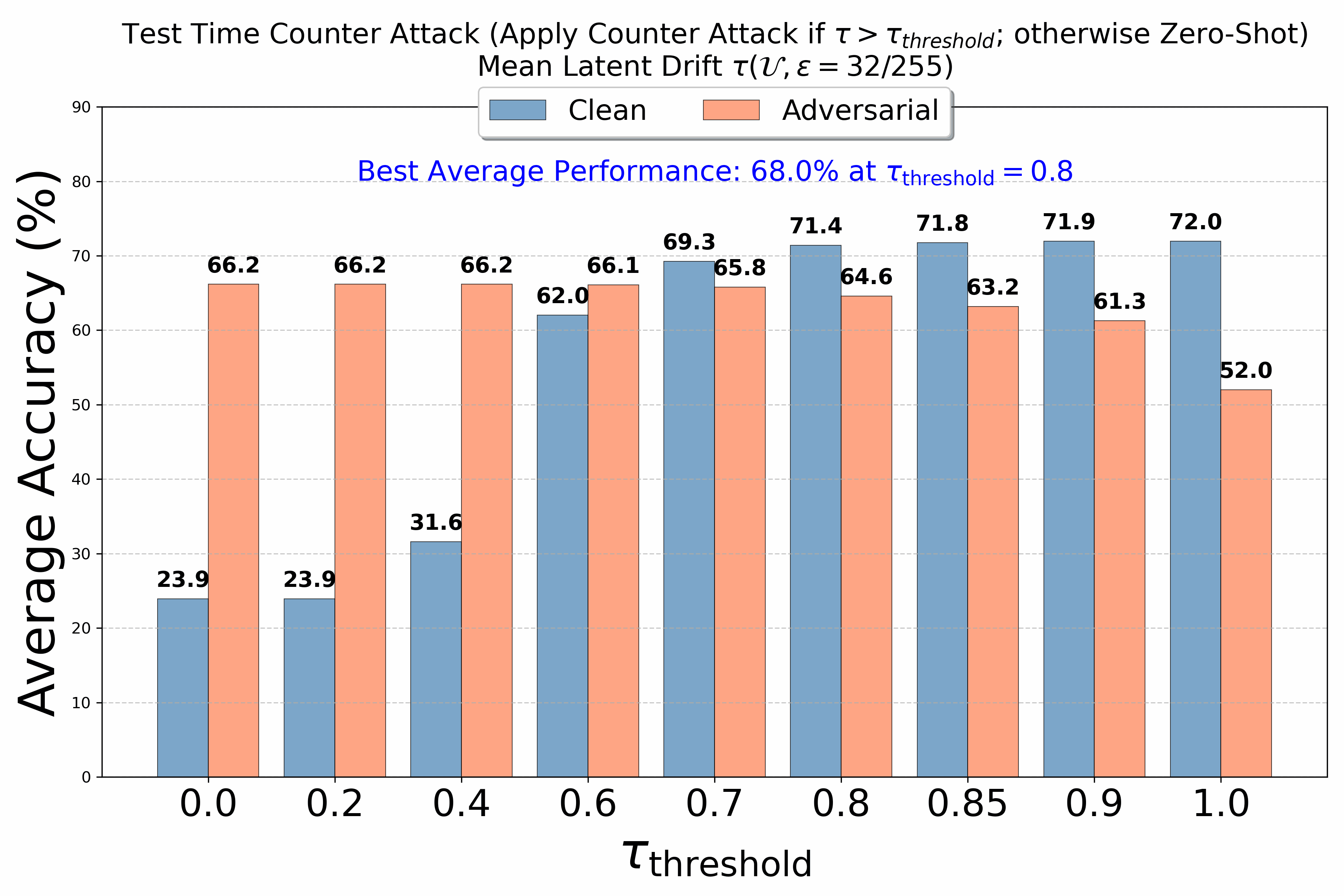}

                \vspace{0.5em}

        \includegraphics[width=\linewidth]{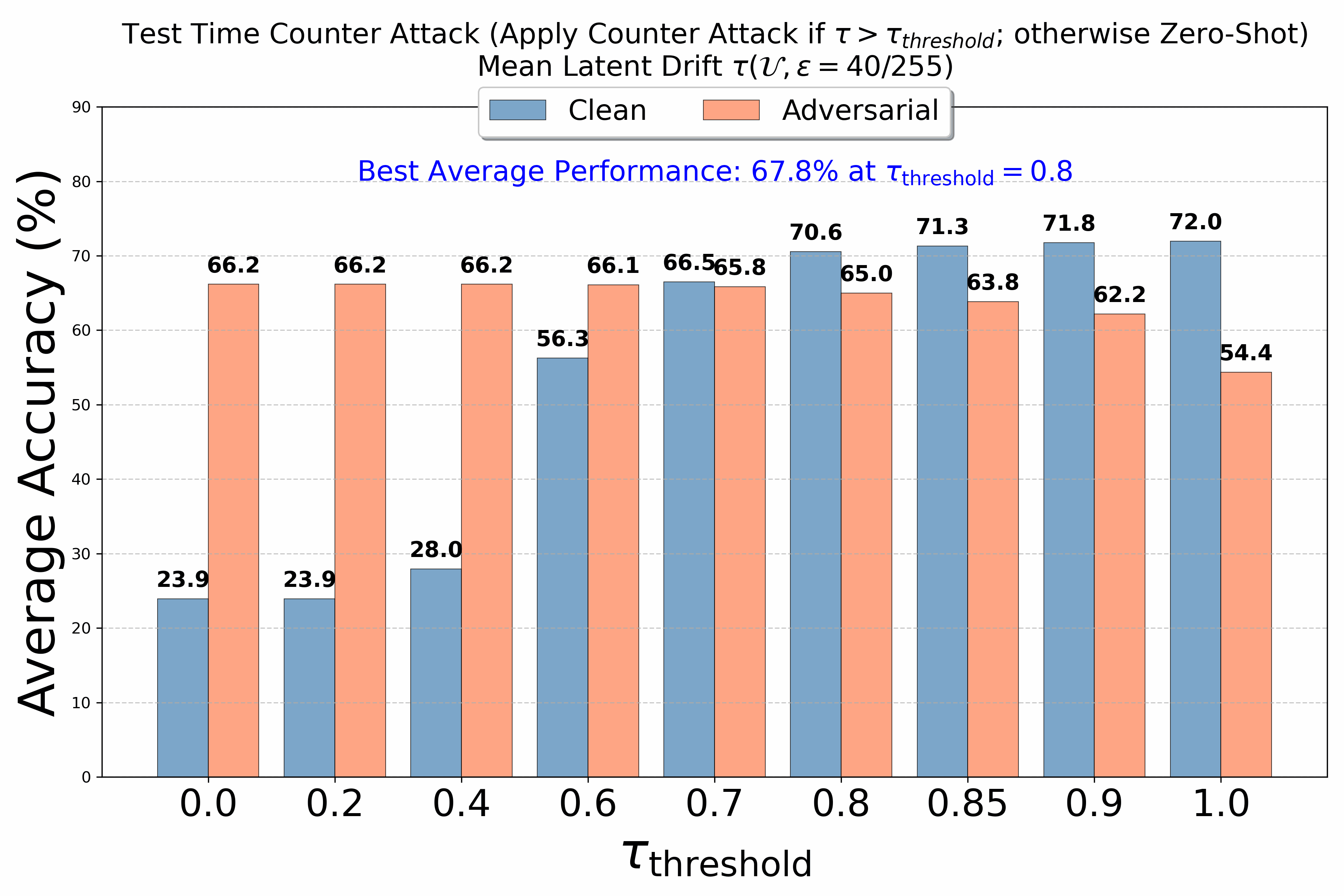}
    \end{minipage}
    \begin{minipage}{0.24\textwidth}
\caption{\textbf{Ours:} Drift-gated TTC 
averaged across ImageNet and its four 
out-of-distribution variants using uniform 
probe noise. Clean (blue) and adversarial 
(red; PGD-100, $\epsilon=4/255$) accuracy 
are shown as a function of the drift threshold 
$\gamma$, above which the counterattack is 
triggered. The gating signal is computed under 
high-noise uniform probe strengths 
$\epsilon \in \{24, 32, 40\}/255$. As $\gamma$ 
increases, TTC is applied more selectively, 
improving clean accuracy while largely 
preserving adversarial robustness. The best 
average accuracy and corresponding threshold 
for each setting are highlighted.}
        \label{fig:results_eval_uniform_noise_ttc_vit_l_14_datacomp_imagenet}
    \end{minipage}
    \vspace{-1em}
\end{figure}

\begin{figure}[t]
    \small \centering
    \begin{minipage}{0.75\textwidth}
        \centering


        \includegraphics[width=\linewidth]{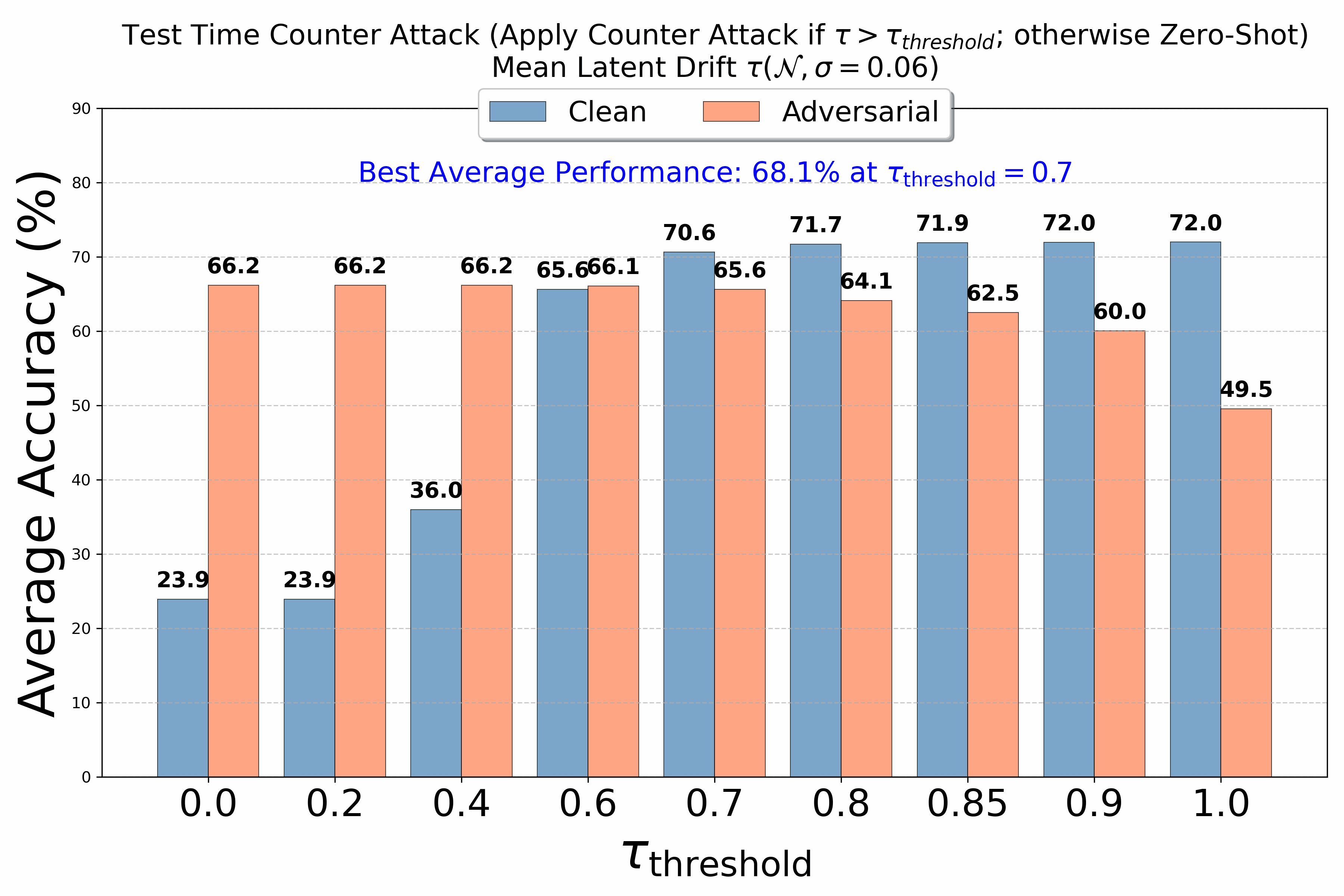}

                \vspace{0.5em}

        \includegraphics[width=\linewidth]{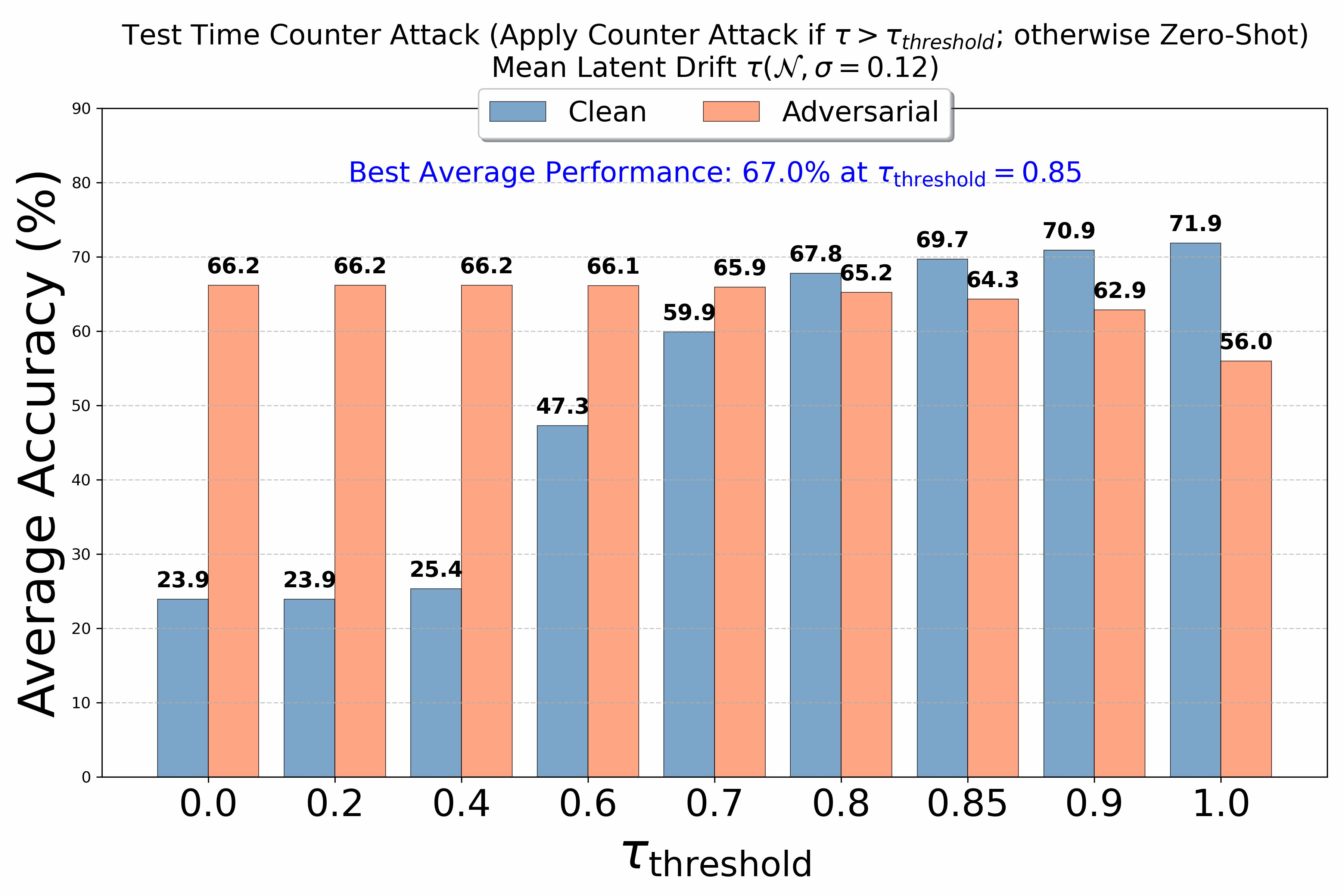}

                \vspace{0.5em}

        \includegraphics[width=\linewidth]{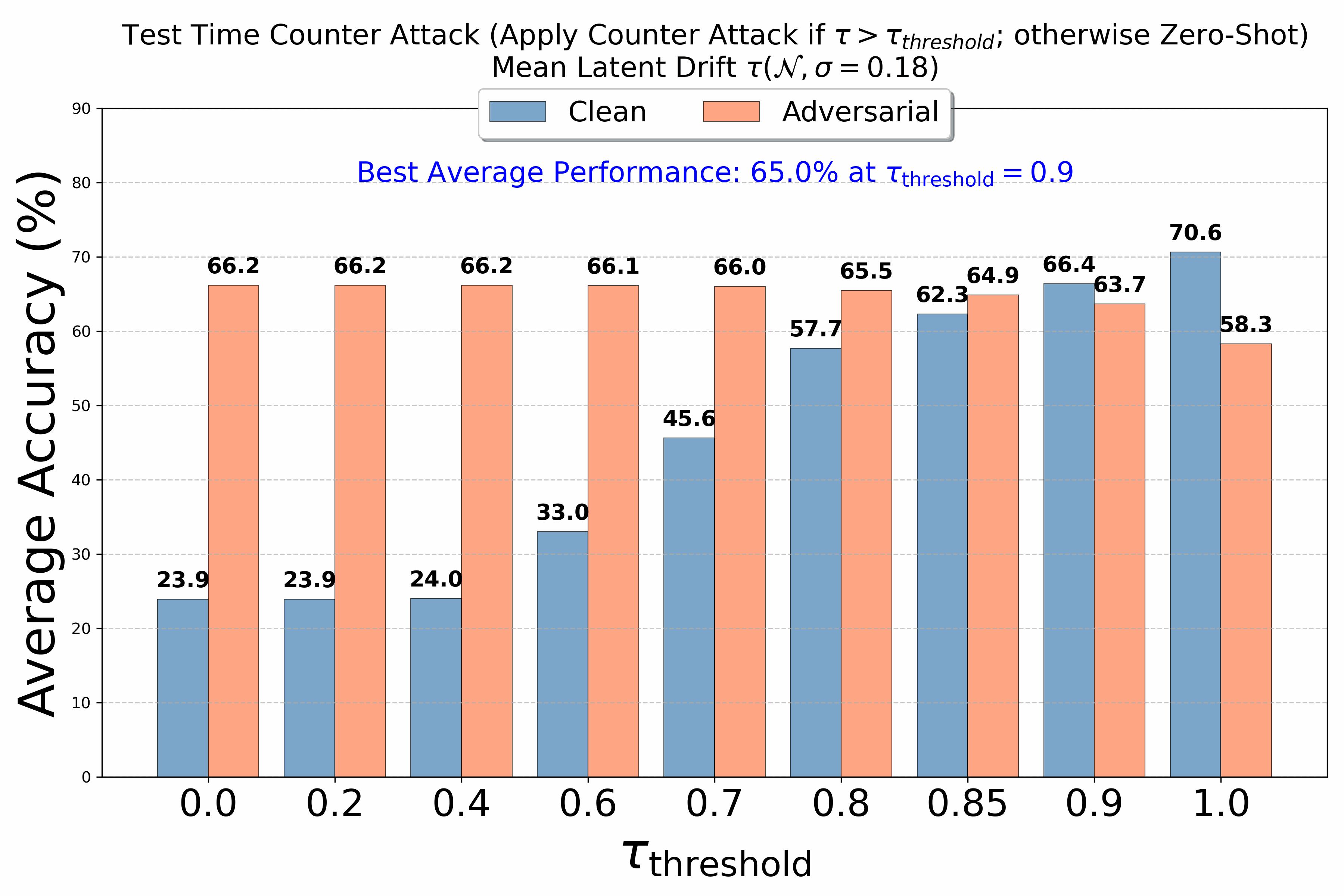}
    \end{minipage}
    \begin{minipage}{0.24\textwidth}
\caption{\textbf{Ours:} Drift-gated TTC 
averaged across ImageNet and its four 
out-of-distribution variants using Gaussian 
probe noise. Clean (blue), adversarial (red; 
PGD-100, $\epsilon=4/255$), and average 
(purple) accuracy are shown as a function of 
the drift threshold $\gamma$, above which the 
counterattack is triggered. The gating signal 
is computed under high-noise Gaussian probe 
strengths $\sigma \in \{0.06, 0.12, 0.18\}$. 
As $\gamma$ increases, TTC is applied more 
selectively, improving clean accuracy while 
largely preserving adversarial robustness. 
The best average accuracy and corresponding 
threshold for each setting are highlighted.}
        \label{fig:results_eval_gaussian_noise_ttc_vit_l_14_datacomp_imagenet}
    \end{minipage}
    \vspace{-1em}
\end{figure}

We additionally evaluate on ImageNet and its 
four out-of-distribution variants, where the 
original TTC achieves $56.10\%$ average 
performance. 
Figures~\ref{fig:results_eval_uniform_noise_ttc_vit_l_14_datacomp_imagenet} 
and~\ref{fig:results_eval_gaussian_noise_ttc_vit_l_14_datacomp_imagenet} 
show that our drift-gated strategy consistently 
improves the clean--robust trade-off across 
all ImageNet variants, confirming that the 
high-noise separation signal generalises 
beyond fine-grained datasets.

All experiments in this section focus on the ViT-L/14 (DataComp-1B) model, which demonstrated the strongest clean and robust performance under stochastic perturbations in the previous subsection.

\subsection{Evaluating Performance with AOM}
\label{sec:app_eval_aom}

We next evaluate how our drift-gated strategy improves the Anchor-guided One-step Linear Movement Method (AOM)~\cite{tong2025zero}. AOM improves robustness by interpolating visual representations toward a noisy anchor constructed from averaging the noisy original samples. In the original formulation, this interpolation is applied uniformly to all inputs, without distinguishing between clean and adversarial samples. As a result, while the method can suppress adversarial perturbations, it may also degrade clean performance.

Our approach introduces a gating mechanism based on mean latent drift. Instead of applying anchor interpolation to all samples, the intervention is triggered only when the measured latent drift exceeds a threshold $\tau$. The gating signal is computed using stochastic perturbations in the high-noise regime identified in our latent drift analysis, enabling a more reliable separation between clean and adversarial inputs.

\begin{figure}[t]
\centering
\small

\includegraphics[width=0.95\linewidth]{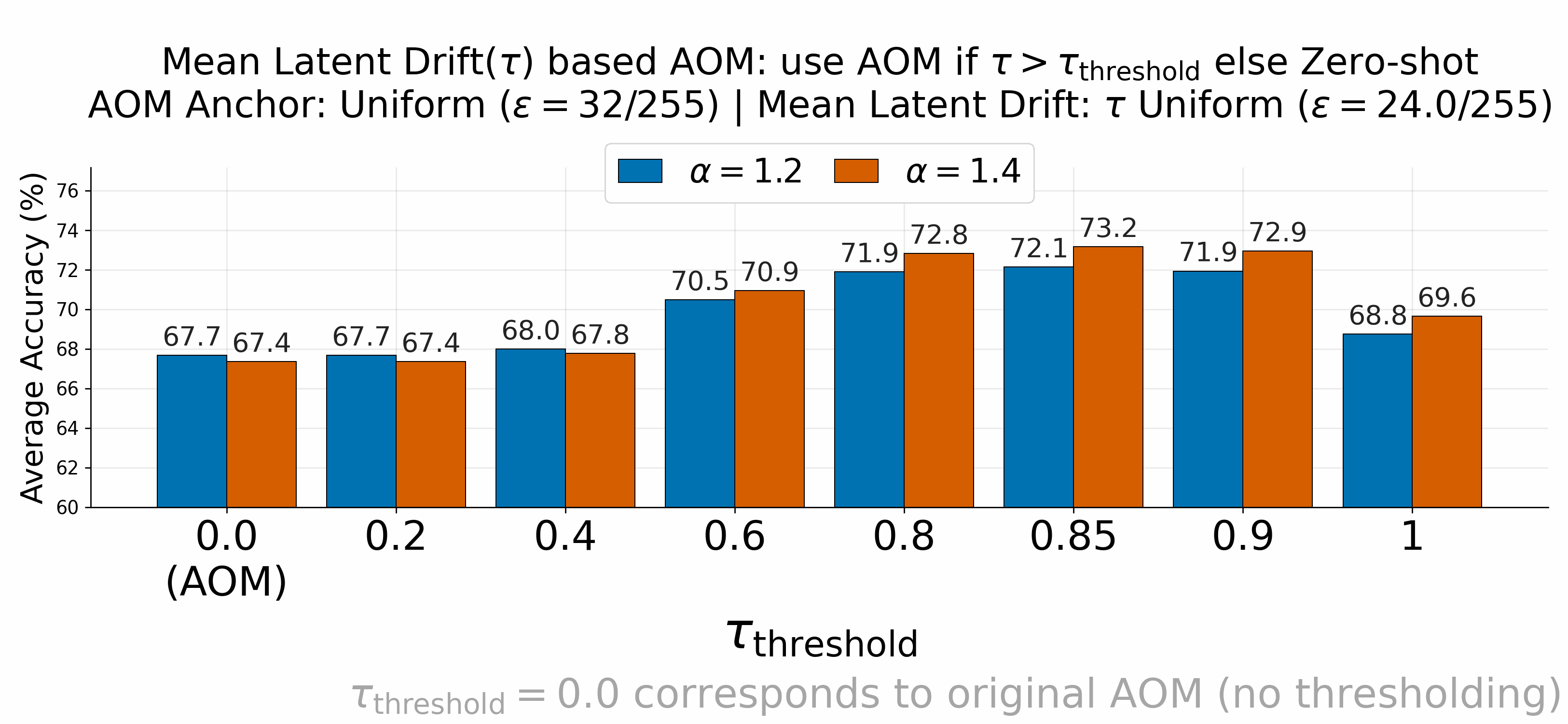}
\includegraphics[width=\linewidth]{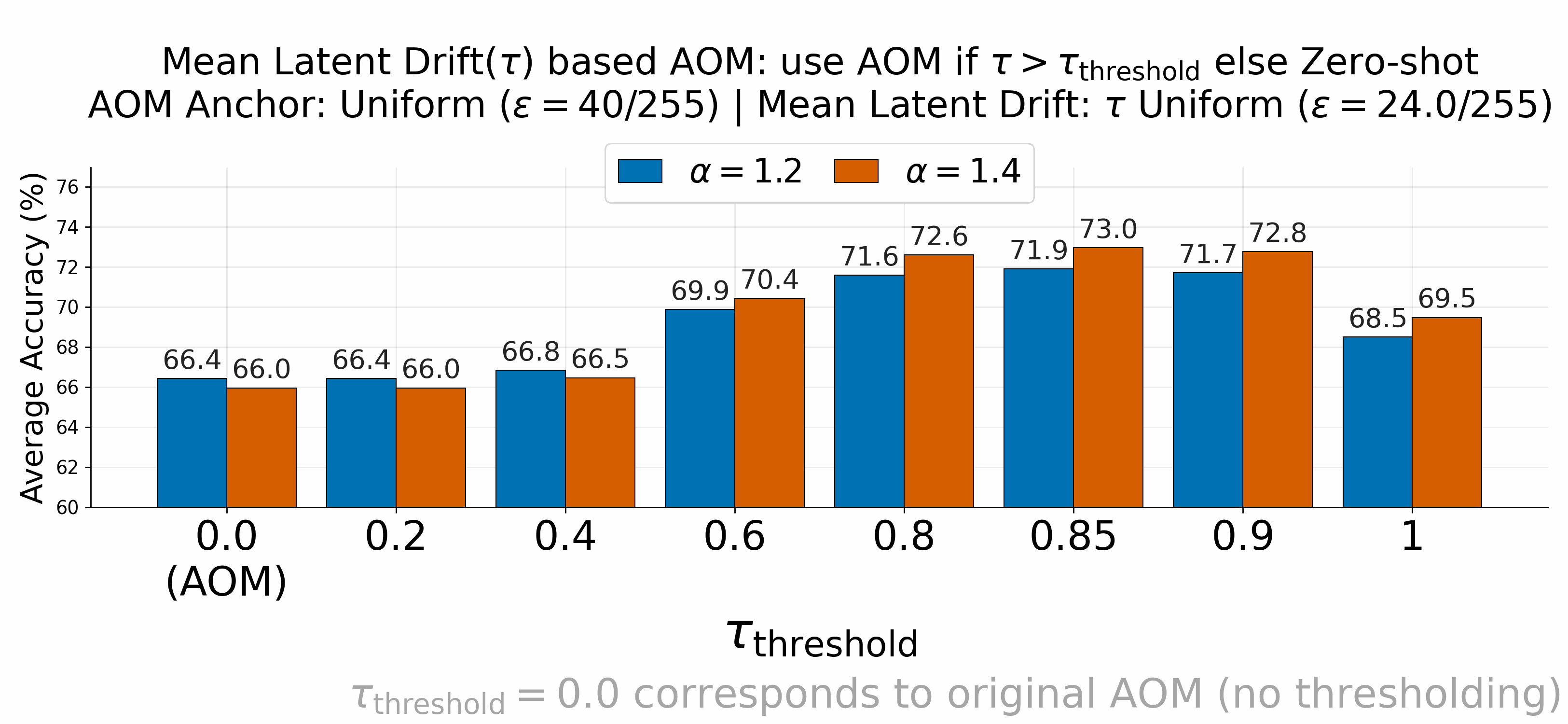}
\includegraphics[width=\linewidth]{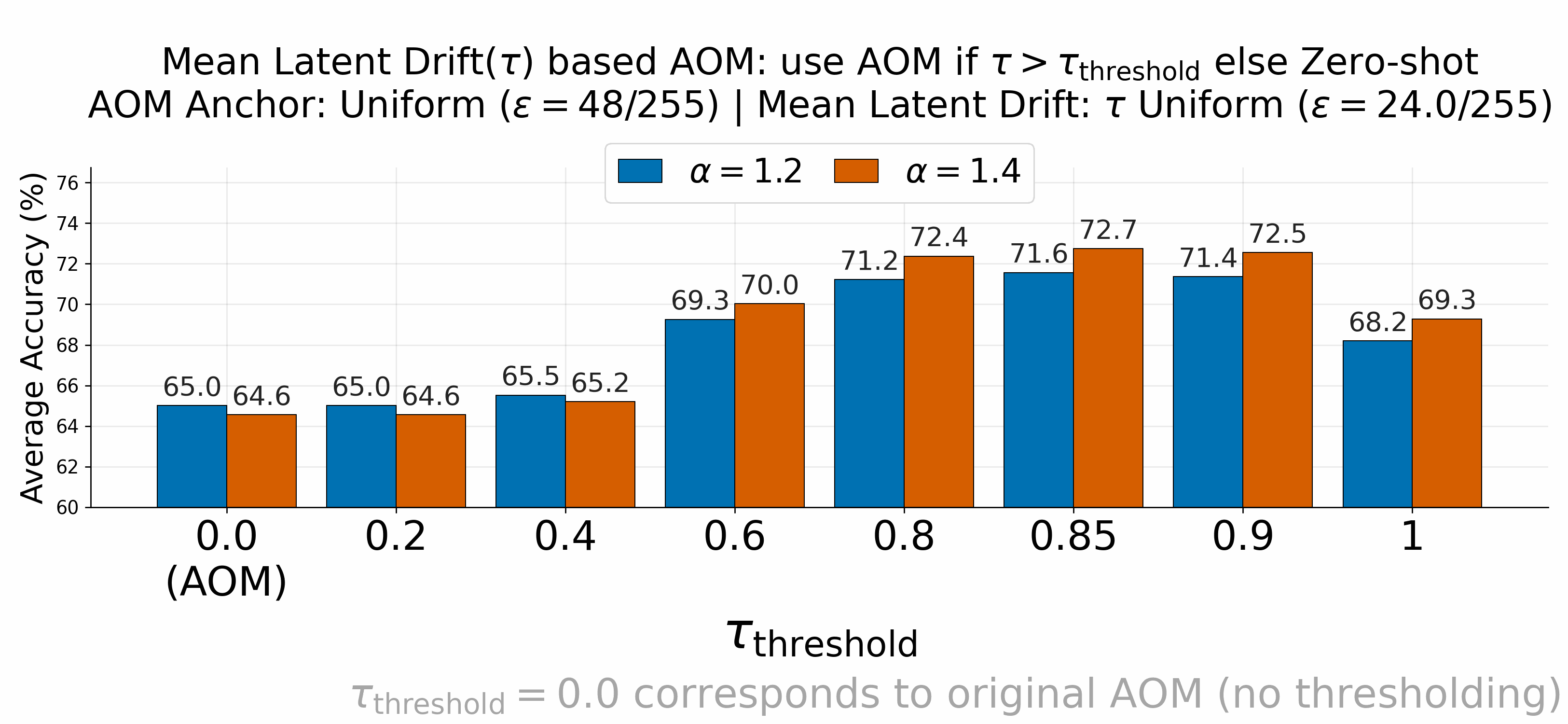}
\caption{\scriptsize Evaluation averaged across eight fine-grained datasets. Performance of our mean latent-drift--gated AOM~\cite{tong2025zero}. Average accuracy is reported as a function of the latent drift threshold $\tau$. Uniform noisy anchors with different perturbation strengths are used for feature interpolation, while the type and strength of noise used to compute mean latent drift are specified in each figure. A threshold $\tau=0$ corresponds to the original AOM formulation, where no distinction is made between clean and adversarial samples and all representations are interpolated toward the noisy anchor. The interpolation is controlled by the factor $\alpha$. In this figure we report the resulting average accuracy for $\alpha \in \{1.2, 1.4\}$. As $\tau$ increases, the interpolation is applied more selectively based on mean latent drift, improving clean accuracy while largely preserving adversarial robustness.}

\label{fig:results_eval_uniform_anchor_aom_vit_l_14_datacomp}

\vspace{-1em}
\end{figure}

\begin{figure}[t]
\centering
\small

\includegraphics[width=0.95\linewidth]{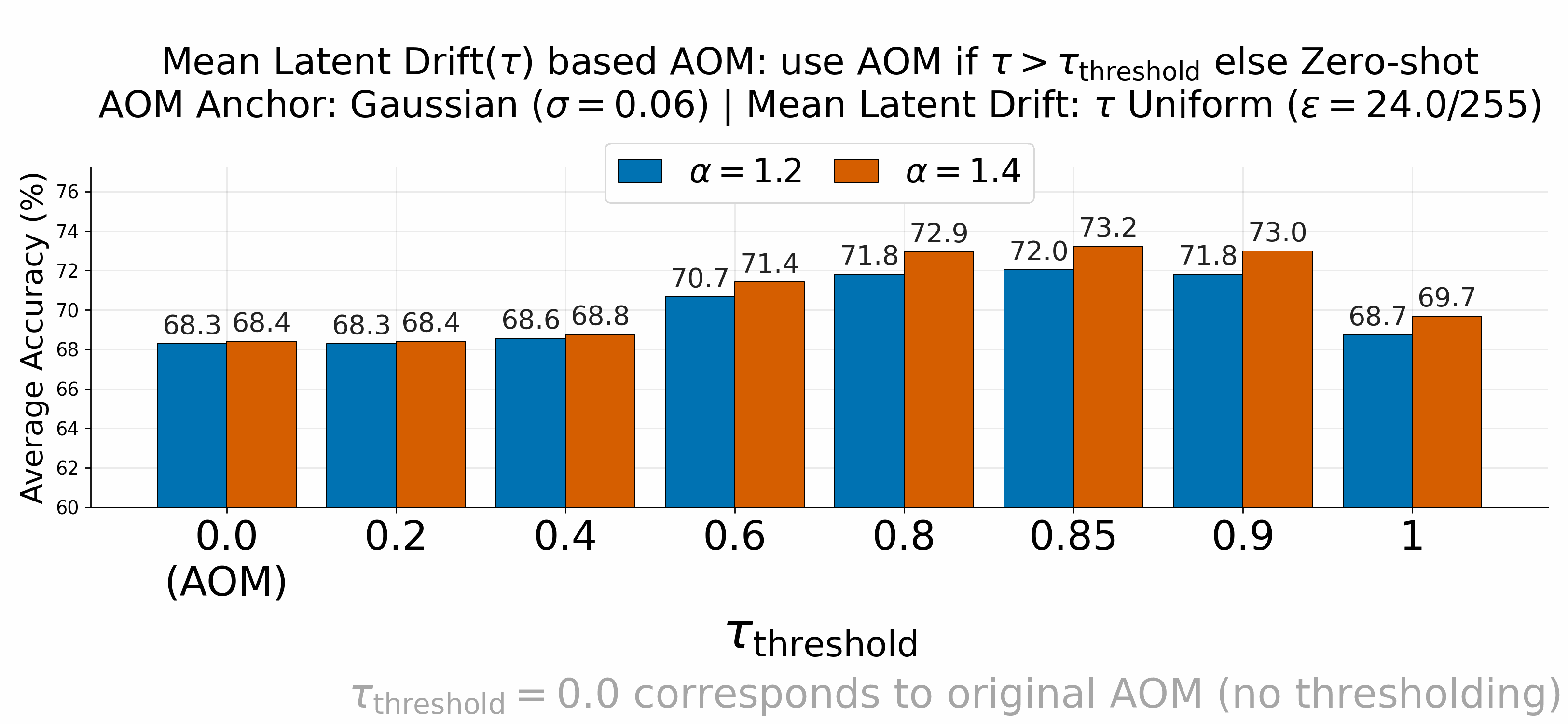}

\vspace{0.5em}

\includegraphics[width=\linewidth]{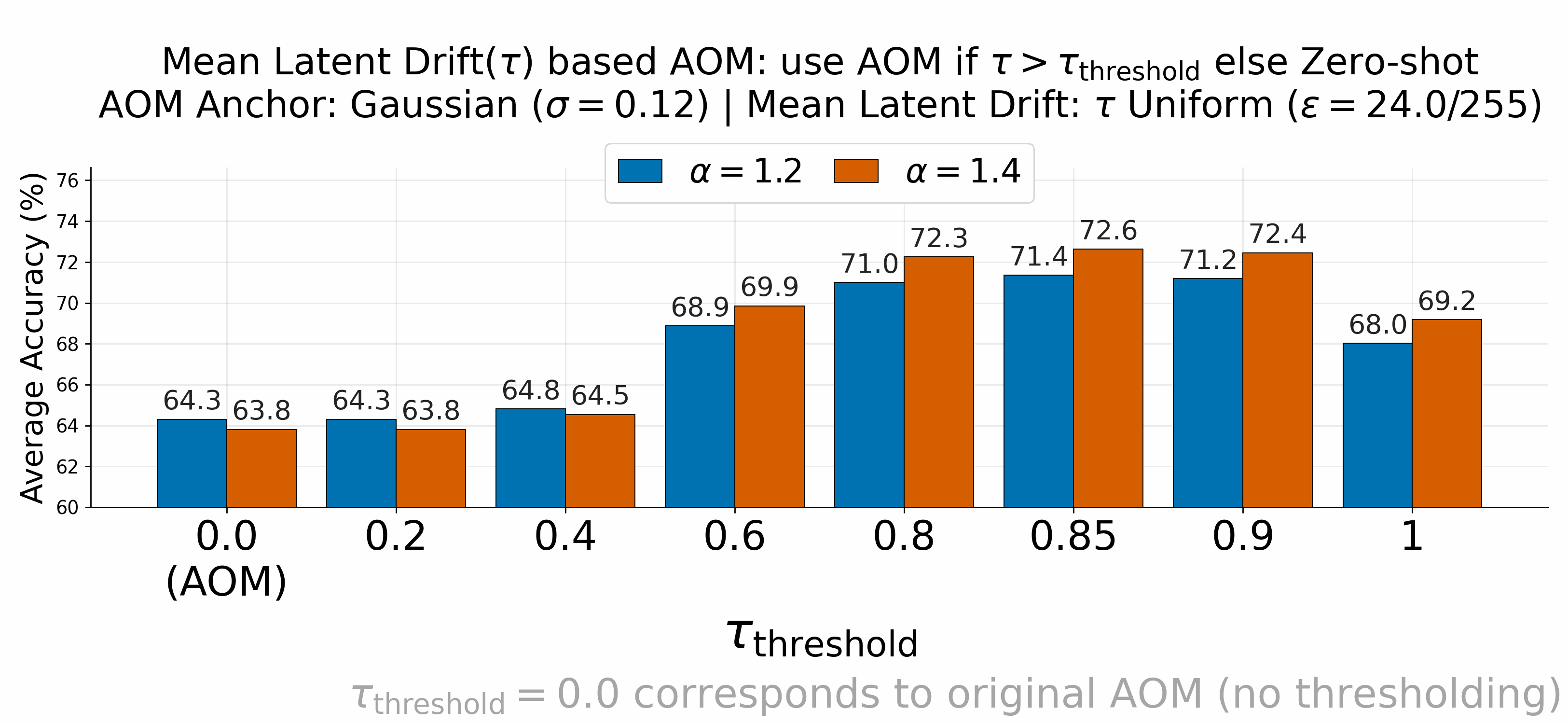}

\vspace{0.5em}

\includegraphics[width=\linewidth]{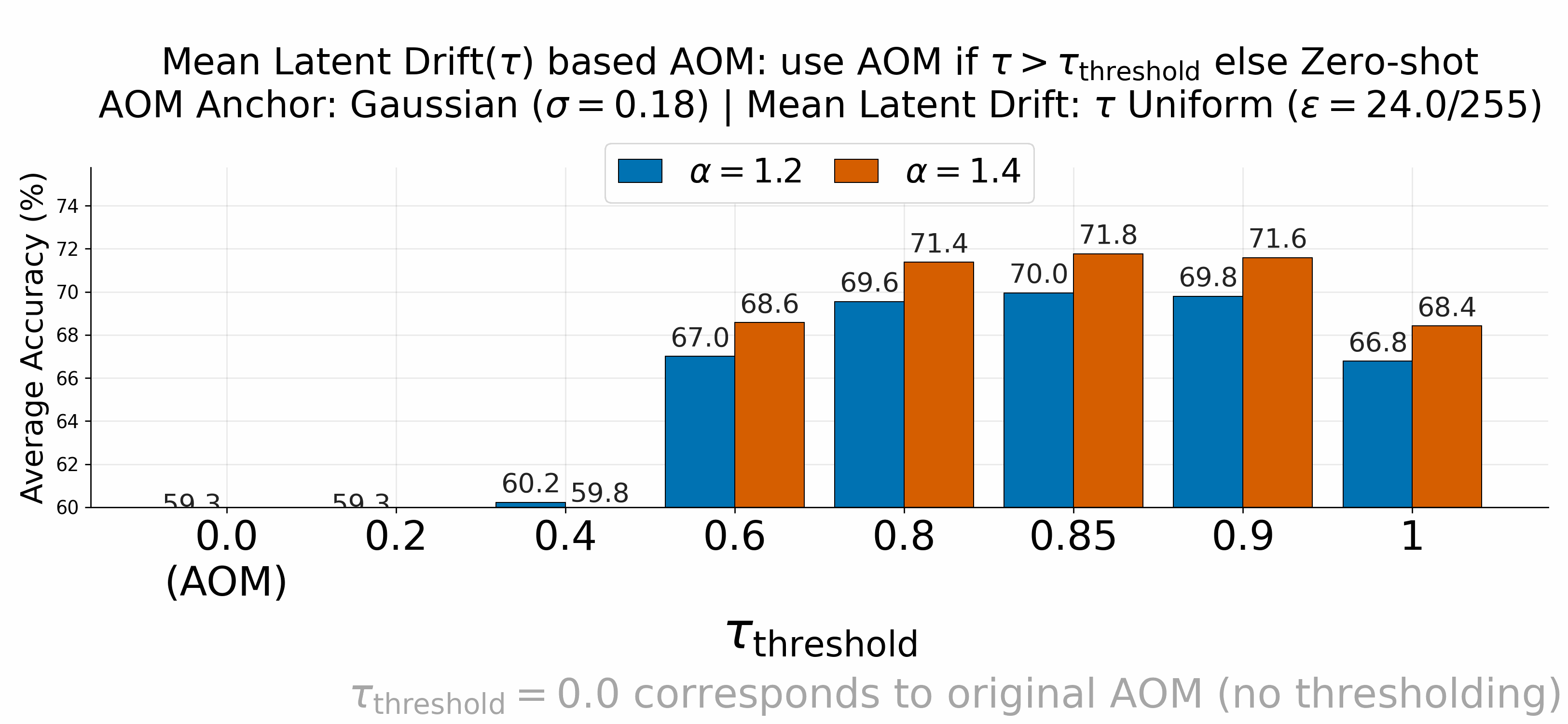}

\caption{\scriptsize Evaluation averaged across eight fine-grained datasets. Performance of our mean latent-drift--gated AOM~\cite{tong2025zero}. Average accuracy is reported as a function of the latent drift threshold $\tau$. Gaussian noisy anchors with different perturbation strengths are used for feature interpolation, while the type and strength of noise used to compute mean latent drift are specified in each figure. A threshold $\tau=0$ corresponds to the original AOM formulation, where no distinction is made between clean and adversarial samples and all representations are interpolated toward the noisy anchor. The interpolation is controlled by the factor $\alpha$. In this figure we report the resulting average accuracy for $\alpha \in \{1.2, 1.4\}$. As $\tau$ increases, the interpolation is applied more selectively based on mean latent drift, improving clean accuracy while largely preserving adversarial robustness.}

\label{fig:results_eval_gaussian_anchor_aom_vit_l_14_datacomp}

\vspace{-1em}
\end{figure}

Figures~\ref{fig:results_eval_uniform_anchor_aom_vit_l_14_datacomp} and~\ref{fig:results_eval_gaussian_anchor_aom_vit_l_14_datacomp} report results averaged across the eight fine-grained datasets using the ViT-L/14 (DataComp-1B) model. These experiments evaluate both uniform and Gaussian noisy anchors used for feature interpolation. The type and strength of noise used to compute mean latent drift are specified in each figure. A threshold $\tau=0$ corresponds to the original AOM formulation, where interpolation is applied to all inputs. As the threshold increases, the interpolation is applied more selectively, reducing unnecessary modifications to clean inputs while maintaining robustness.

\begin{figure}[t]
\centering
\small

\includegraphics[width=\linewidth]{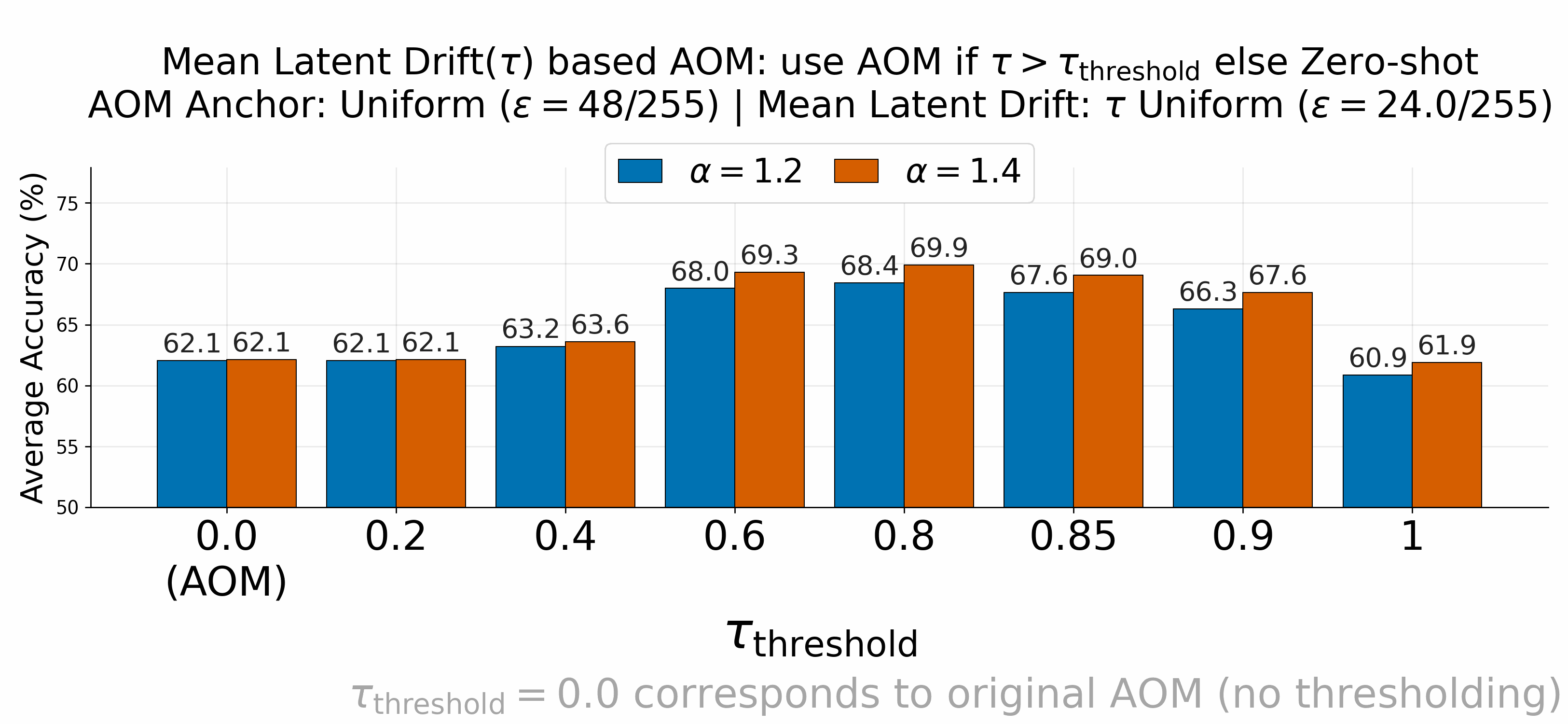}

\vspace{0.5em}

\includegraphics[width=\linewidth]{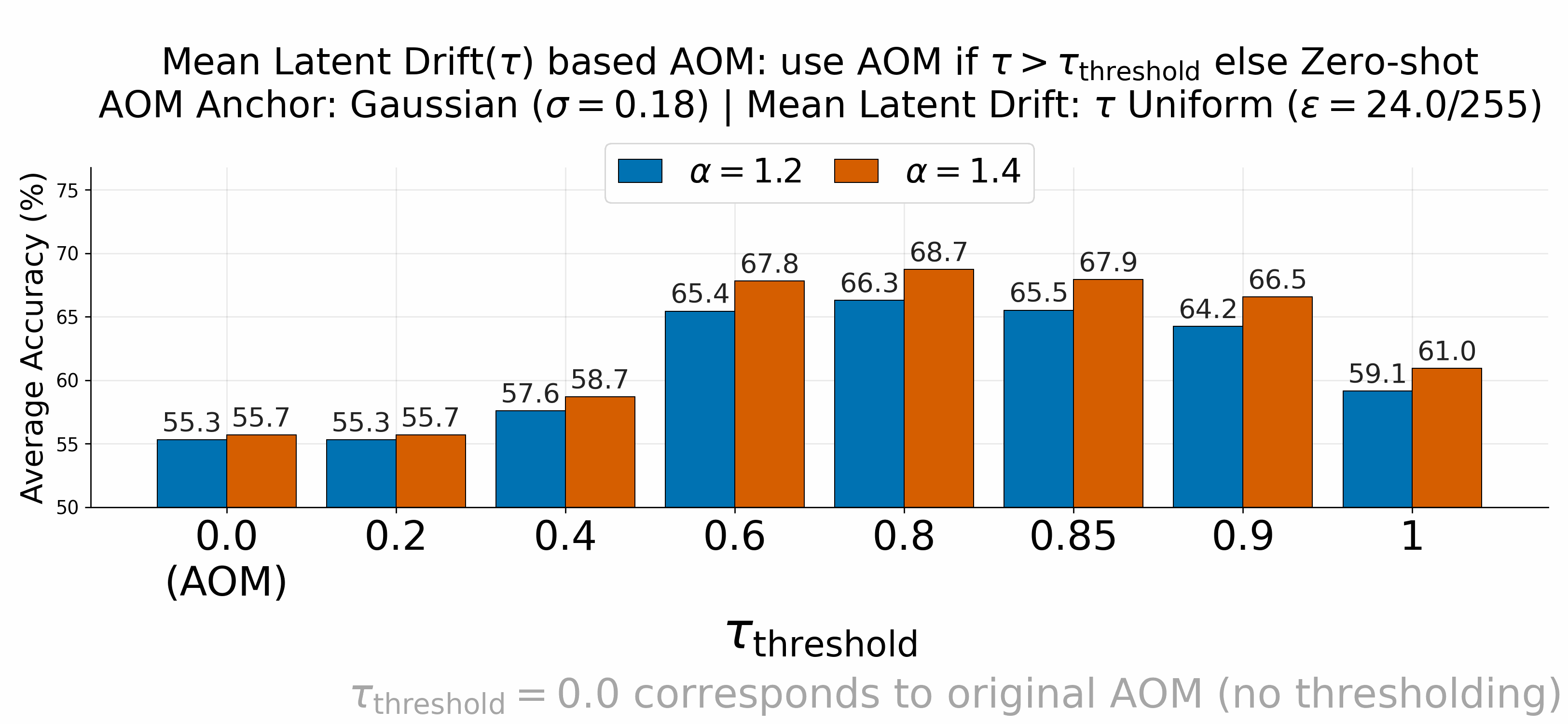}

\caption{\small Evaluation averaged across ImageNet and its four out-of-distribution variants. Performance of our mean latent-drift--gated AOM~\cite{tong2025zero}. Average accuracy is reported as a function of the latent drift threshold $\tau$. Uniform  and gaussian noisy anchors are used for feature interpolation, while the type and strength of noise used to compute mean latent drift are specified in each figure. A threshold $\tau=0$ corresponds to the original AOM formulation, where no distinction is made between clean and adversarial samples and all representations are interpolated toward the noisy anchor. The interpolation is controlled by the factor $\alpha$. In this figure we report the resulting average accuracy for $\alpha \in \{1.2, 1.4\}$. As $\tau$ increases, the interpolation is applied more selectively based on mean latent drift, improving clean accuracy while largely preserving adversarial robustness.}

\label{fig:results_eval_uniform_gaussian_anchor_aom_vit_l_14_datacomp_imagenet}

\vspace{-1em}
\end{figure}

We additionally evaluate the same approach on ImageNet and its four out-of-distribution variants, as shown in Figure~\ref{fig:results_eval_uniform_gaussian_anchor_aom_vit_l_14_datacomp_imagenet}. Similar trends are observed: selective interpolation based on mean latent drift improves the clean–robust trade-off compared to the original AOM formulation.

\subsection{Evaluating Performance with R-TPT}
\label{sec:app_eval_rtpt}

Robust Test-Time Prompt Tuning (R-TPT)~\cite{sheng2025r} improves adversarial robustness mostly by aggregating predictions across multiple stochastic views of the input during inference. While this strategy preserves strong clean performance, its robustness gains primarily arise from view aggregation and remain limited compared to stronger interventions such as counterattacks or feature interpolation.

To combine the strengths of both approaches, we retain the original R-TPT pipeline and augment it with our mean latent-drift gating strategy. Specifically, we use mean latent drift computed under stochastic perturbations to identify inputs exhibiting adversarial-like instability. TTC-style counterattacks are then selectively triggered only for these inputs before applying R-TPT, while only standard R-TPT inference is applied to the remaining samples.

Table~\ref{tab:rtpt_ttc_dataset_results} reports dataset-wise clean and adversarial accuracy across the eight fine-grained benchmarks. The original R-TPT achieves an average accuracy of $68.83\%$(clean and adversarial combined) across these datasets. By incorporating our drift-gated TTC intervention, the average accuracy improves to $73.16\%$ when latent drift is computed using uniform noise and to $72.86\%$ when computed using Gaussian noise.

These results demonstrate that selectively combining counterattack-based defenses with stochastic view aggregation substantially improves adversarial robustness while largely preserving the strong clean performance of the original R-TPT framework.

\clearpage

\begin{table}[h]
\centering
\small
\setlength{\tabcolsep}{4pt}
\renewcommand{\arraystretch}{1.12}

\begin{adjustbox}{width=\linewidth}
\begin{tabular}{lccccccccc}
\toprule
& Aircraft & Caltech101 & Cars & DTD & EuroSAT & Flowers & Pets & UCF101 & Avg. \\
\midrule

\rowcolor{gray!15}
\multicolumn{10}{l}{\textbf{RTPT}} \\
\rowcolor{gray!5}
\quad Clean & 46.59 & 97.85 & 94.01 & 62.94 & 54.96 & 82.38 & 95.07 & 72.11 & 75.74 \\
\rowcolor{gray!5}
\quad Adv   & 31.41 & 94.04 & 81.61 & 51.48 & 26.07 & 72.72 & 82.69 & 55.38 & 61.92 \\
\midrule

\rowcolor{blue!12}
\multicolumn{10}{l}{\textbf{RTPT + TTC (Ours, Uniform $\epsilon=24/255$)}} \\
\rowcolor{blue!4}
\quad Clean & 45.48 & 97.69 & 93.82 & 62.88 & 43.74 & 82.38 & 94.99 & 71.66 & 74.08 \\
\rowcolor{blue!4}
\quad Adv   & 42.39 & 96.47 & 91.77 & 58.39 & 45.31 & 81.16 & 94.06 & 68.28 & 72.23 \\
\midrule

\rowcolor{green!12}
\multicolumn{10}{l}{\textbf{RTPT + TTC (Ours, Gaussian $\sigma=0.12$)}} \\
\rowcolor{green!4}
\quad Clean & 45.39 & 97.63 & 93.74 & 62.35 & 39.57 & 82.34 & 94.99 & 70.95 & 73.37 \\
\rowcolor{green!4}
\quad Adv   & 42.42 & 96.51 & 91.88 & 58.63 & 45.31 & 81.32 & 94.36 & 68.33 & 72.35 \\
\bottomrule
\end{tabular}
\end{adjustbox}

\caption{Dataset-wise clean and adversarial accuracy comparison between the original R-TPT~\cite{sheng2025r} and our drift-gated R-TPT + TTC variants across eight fine-grained datasets. Our method augments the R-TPT pipeline with a mean latent-drift gate that selectively triggers TTC-style counterattacks on inputs exhibiting adversarial-like instability. Results are reported for gating signals computed using uniform noise ($\epsilon=24/255$) and Gaussian noise ($\sigma=0.12$). The proposed hybrid strategy substantially improves adversarial robustness across datasets while largely preserving the strong clean performance of R-TPT.}
\label{tab:rtpt_ttc_dataset_results}
\end{table}

\subsection{Attack Scope}
\label{sec:app_attack_scope}

In this work we consider the standard white-box adversarial evaluation setting commonly used for evaluating test-time defenses in vision--language models~\cite{xing2025clip,tong2025zero,sheng2025r}. Specifically, adversarial examples are generated using Projected Gradient Descent (PGD) with $L_{\infty}$ perturbation budget $\epsilon$. The attack is applied to the underlying CLIP model without any test-time defense, and the resulting adversarial examples are then evaluated under different inference-time defenses.

This evaluation protocol reflects a realistic deployment scenario in which attackers have access to the publicly available pretrained model but are unaware of the defense applied at inference time. Under this assumption, adversarial examples are generated against the base CLIP model prior to applying any test-time intervention.

Our experiments focus on defenses that operate purely at inference time without modifying the pretrained model weights. This includes all baselines considered in this work, namely Test-Time Counterattack (TTC)~\cite{xing2025clip}, Anchor-based Interpolation Method (AOM)~\cite{tong2025zero}, and Robust Test-Time Prompt Tuning (R-TPT)~\cite{sheng2025r}. These methods share a common assumption that adversarial inputs are generated against the original model and that the defense is applied only during inference.

Under this attack scope, our drift-gated framework remains fully compatible with existing test-time defenses, as it only modifies the decision of when the intervention should be applied, rather than altering the attack model or requiring access to adversarial training data.

\end{document}